%% file: main.tex
\documentclass{article}
\usepackage{iclr2027_arxiv,times}
\input{math_commands.tex}

\usepackage{amsmath,amssymb}
\usepackage{booktabs}
\usepackage{array,colortbl}
\usepackage{tabularx}
\usepackage{graphicx}
\usepackage{wrapfig}
\usepackage{float}
\usepackage{placeins}
\usepackage[ruled,linesnumbered,longend,noline]{algorithm2e}
\usepackage{tcolorbox}
\usepackage{hyperref}
\usepackage{url}
\usepackage{microtype}
\input{generated/results_macros.tex}
\input{generated/online_macros.tex}

\newcommand{\method}{\textsc{CASE}}

\definecolor{tableheader}{RGB}{238,238,238}
\definecolor{caseblue}{RGB}{221,235,247}
\definecolor{promptback}{RGB}{246,249,252}
\definecolor{promptframe}{RGB}{172,188,204}
\definecolor{prompttitle}{RGB}{234,241,248}
\newcommand{\tablefont}{\fontsize{7.5pt}{9.0pt}\selectfont}

\newcommand{\tablestandard}{\tablefont\renewcommand{\arraystretch}{1.04}\setlength{\tabcolsep}{3pt}}
\newcolumntype{Y}{>{\centering\arraybackslash}X}
\newcolumntype{Z}{>{\hsize=1.35\hsize\centering\arraybackslash}X}
\newcolumntype{Q}{>{\hsize=0.65\hsize\centering\arraybackslash}X}
\newcommand{\tableheadtoprule}{\specialrule{\heavyrulewidth}{\abovetopsep}{0pt}}
\newcommand{\tableheadmidrule}{\specialrule{\lightrulewidth}{0pt}{\belowrulesep}}
\makeatletter
\newcommand{\tableheadtoppad}{%
  \rule{0pt}{\dimexpr\ht\@arstrutbox+\belowrulesep\relax}}
\newcommand{\tableheadbottompad}{%
  \rule[-\dimexpr\dp\@arstrutbox+\aboverulesep\relax]{0pt}{%
    \dimexpr\ht\@arstrutbox+\dp\@arstrutbox+\aboverulesep\relax}}
\makeatother
\newcommand{\tableheadrow}{\rowcolor{tableheader}}
\newcommand{\tableheadrowspan}{\rowcolor{tableheader}}
\newcommand{\oursrow}{\rowcolor{caseblue}}
\SetAlgoNlRelativeSize{-1}
\SetAlFnt{\small}
\SetAlCapFnt{\small\bfseries}
\SetAlCapNameFnt{\small\bfseries}
\SetAlgoCaptionSeparator{\quad}
\DontPrintSemicolon
\newtcolorbox{promptbox}[1]{%
  colback=promptback,
  colframe=promptframe,
  colbacktitle=prompttitle,
  coltitle=black,
  boxrule=0.45pt,
  arc=1.5pt,
  titlerule=0.35pt,
  left=7pt,right=7pt,top=4pt,bottom=4pt,
  before skip=4pt,after skip=5pt,
  title=#1,
  fonttitle=\small\bfseries,
  fontupper=\footnotesize\ttfamily\raggedright}
\newcommand{\accpair}[2]{%
  \ifdim #1pt>#2pt \textbf{#1} $\rightarrow$ #2%
  \else\ifdim #2pt>#1pt #1 $\rightarrow$ \textbf{#2}%
  \else #1 $\rightarrow$ #2\fi\fi}
\newcommand{\tokpair}[2]{%
  \mbox{\ifdim #1pt<#2pt \textbf{#1} $\rightarrow$ #2%
  \else\ifdim #2pt<#1pt #1 $\rightarrow$ \textbf{#2}%
  \else #1 $\rightarrow$ #2\fi\fi}}

\title{CASE: Cost-Aware Stopping for Efficient Long-Video Agents}
\author{\textbf{Yiming Du\textsuperscript{*}, Chenghao Liu\textsuperscript{*$\dagger$}, Zhiyuan Liu, Fangxing Zheng,}\\
\textbf{Zhao Wang, Junnan Nie, Songfang Huang\textsuperscript{$\ddagger$}}\\
{\normalfont Peking University}}
\iclrfinalcopy
\begin{document}
\maketitle
\begingroup
\renewcommand{\thefootnote}{\fnsymbol{footnote}}
\footnotetext[1]{Equal contribution.\quad\textsuperscript{$\dagger$} Project leader.\quad\textsuperscript{$\ddagger$} Corresponding author: \href{mailto:hsf@pku.edu.cn}{\texttt{hsf@pku.edu.cn}}.}
\endgroup

\begin{abstract}
Long-video agents can actively gather question-relevant evidence, but they typically leave a central decision implicit: when has the agent seen enough to answer? We propose CASE, a plug-in termination framework that frames this decision as policy-conditioned sequential stopping. At each causal checkpoint, CASE combines an auxiliary multiple-choice assessment of accumulated evidence with the host agent's execution state. From complete native trajectories, we construct a cost-aware target that compares answering now with stopping later along the same search path, accounting jointly for answer correctness and the full cost of continued reasoning. A lightweight Ridge regressor learns this decision gap and produces STOP/CONTINUE decisions. We evaluate three vision-language models with VideoSeek and AVP. On Video-MME, end-to-end accuracy changes by $+0.67$ percentage points on average while CASE reduces model-token use by 53.63\%. The same frozen policies then transfer zero-shot to LongVideoBench and MLVU, with end-to-end accuracy changes of $+3.38$ and $+4.58$ points while saving 58.78\% and 51.28\% of model tokens, respectively. Across all agent--model--benchmark combinations, CASE attains the highest accuracy--efficiency Pareto-frontier coverage among the compared stopping methods (83.3\%) at the selected operating points. Online execution preserves this favorable accuracy--efficiency trade-off and additionally reduces measured runtime by 54.1\% on average. CASE provides a plug-in termination framework for long-video reasoning agents, enabling them to decide when further evidence acquisition is no longer worthwhile.
\end{abstract}

\section{Introduction}
Long-video question answering often depends on locating sparse question-relevant evidence across a long temporal span. As videos grow longer, dense analysis increases the costs of visual processing, context construction, and model inference.

Existing work has substantially improved how long-video agents find evidence. VideoAgent methods use iterative visual queries and structured memory to locate and organize question-relevant information, while VideoSeek combines video logic flows with multi-granularity tools to narrow search from a global overview to relevant temporal regions in a think--act--observe loop \citep{wang2024videoagent,fan2024videoagentmemory,lin2026videoseek}. These advances make active evidence acquisition effective, yet leave a key decision unresolved: after receiving new evidence, has the agent seen enough to answer, or is another search step still worthwhile?

Existing agents typically leave termination implicit in the generative model or impose a fixed step or budget limit. Some systems add explicit sufficiency checks; AVP, for example, uses a reflector to assess whether accumulated evidence is sufficient to answer \cite{wang2026activevideoperception}. Fixed limits apply the same stopping pressure across questions, while confidence, entropy, or agreement signals summarize current certainty rather than the value of another search step \citep{zhou2020bertpatience,schuster2022calm,kieback2026tasr}. Useful evidence can still follow a high-confidence state, and an unstable answer can coincide with an expensive low-value tool call. The relevant comparison is therefore between the risk of answering now and the expected benefit and cost of continuing.

Sequential hypothesis testing provides a natural framework for this decision. Unlike fixed-sample inference, it allows a decision maker to stop as evidence arrives: each observation updates support for competing hypotheses, after which the process either terminates or continues sampling, trading lower decision risk against additional observation cost \citep{wald1945sequential}. Active multiple-hypothesis testing and controlled sensing extend this setting to chosen observations and nonuniform costs \citep{chernoff1959sequential,naghshvar2013active,nitinawarat2015controlled,vershinin2025nonhomogeneous}. Once the host agent's evidence-acquisition policy is fixed, the remaining problem is optimal stopping---comparing the risk of answering now with the value of acquiring more information along that policy \citep{shiryaev2008optimal,novikov2010optimal}. This perspective maps directly to long-video agents: answer options are competing hypotheses, video-tool results arrive as sequential evidence, and each causal checkpoint poses a STOP/CONTINUE decision. Because future tool use and model computation depend on the evolving trajectory, continuation cost is state-dependent and heterogeneous.

\begin{wrapfigure}[20]{r}{0.45\textwidth}
\vspace{-0.35\baselineskip}
\centering
\includegraphics[width=\linewidth]{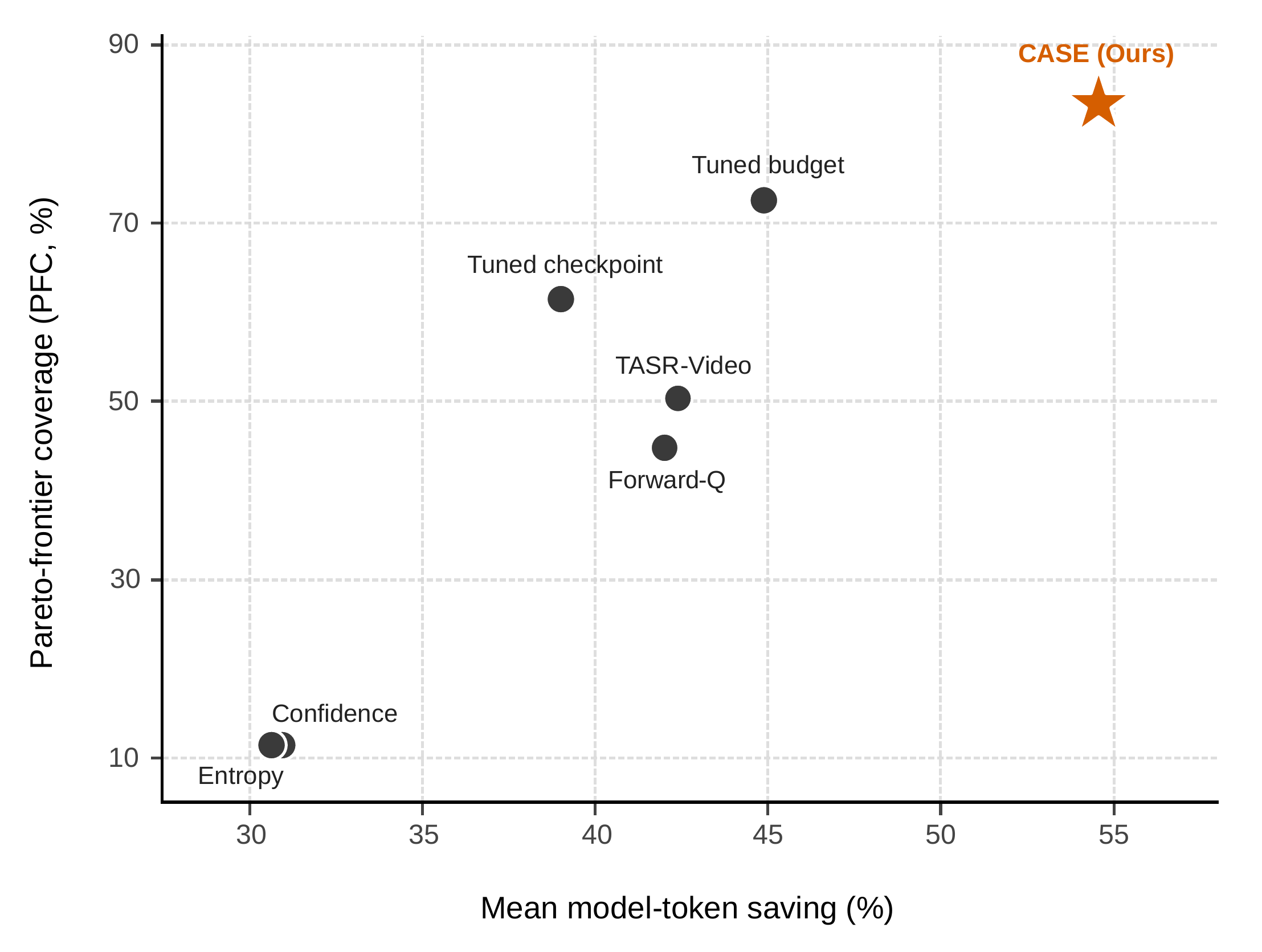}
\caption{\textbf{Cross-configuration accuracy--efficiency frontier.} PFC is the share of agent--model--benchmark configurations on the Pareto frontier.}
\label{fig:intro_tradeoff}
\end{wrapfigure}

\begin{figure}[!t]
\centering
\includegraphics[width=\textwidth]{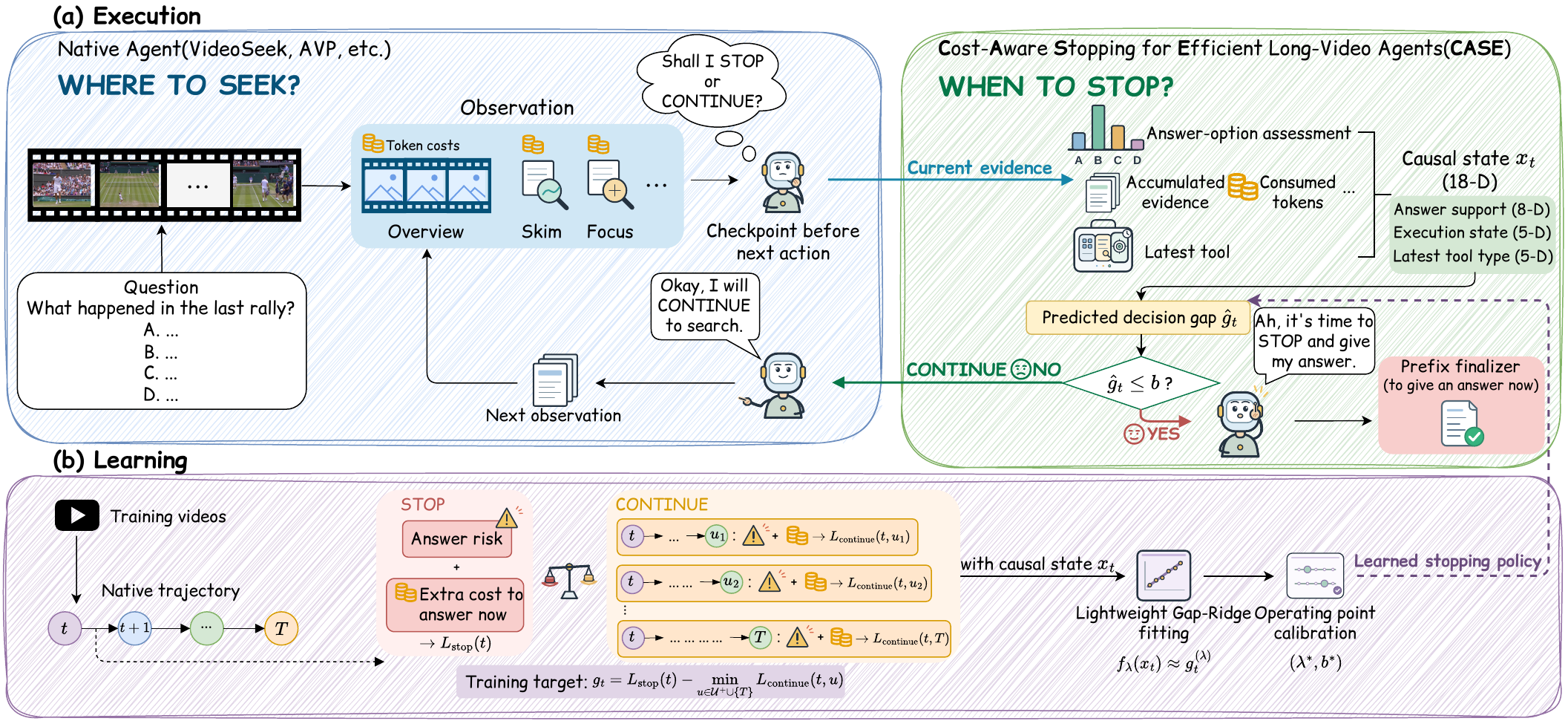}
\caption{\textbf{Overview of \method{}.} The host chooses where to acquire evidence; \method{} wraps the native search loop with STOP/CONTINUE control and constrained prefix finalization. Complete native trajectories supervise answer quality and heterogeneous continuation cost.}
\label{fig:case_overview}
\end{figure}

Motivated by this perspective, we propose \method{} (\textbf{C}ost-\textbf{A}ware \textbf{S}topping for \textbf{E}fficient Long-Video Agents), a plug-in termination framework for long-video agents. The host agent retains control over where to seek evidence, while its learned stopping controller decides whether to execute the next proposed visual action at each causal checkpoint. At each checkpoint, it combines answer support with the agent's execution state. Complete native trajectories expose the future evidence and realized costs reachable under the host search policy, yielding the stopping target
\[
g_t =
L_{\mathrm{stop}}(t)
-
L_{\mathrm{continue}}^{*}(t),
\]
where $L_{\mathrm{continue}}^{*}(t)$ is the lowest answer-risk-plus-cost loss among later checkpoints and the native terminal point on the same trajectory. We factorize this gap into a future answer-quality advantage and a heterogeneous-cost correction and fit both with lightweight Ridge models. Figure~\ref{fig:case_overview} summarizes the execution and learning pipeline. This checkpoint-level stopping decision differs from Stop-RAG's iterative retrieval stopping and CAM-DF's pre-execution tool configuration \citep{park2025stoprag,feng2026camdf}.

We evaluate \method{} on VideoSeek and Active Video Perception (AVP) \citep{lin2026videoseek,wang2026activevideoperception} with three open-source vision-language models over Video-MME, LongVideoBench, and MLVU. Figure~\ref{fig:intro_tradeoff} summarizes the selected accuracy--efficiency operating points across the 18 agent--model--benchmark configurations. Policies selected on Video-MME transfer zero-shot to the other benchmarks and are also evaluated online, where the favorable accuracy--efficiency trade-off persists and measured runtime falls by 54.1\% on average.

To summarize, our contributions are three-fold:
\vspace{0.6\baselineskip}
\begin{itemize}
\item We formulate tool-guided long-video question answering as a multi-hypothesis sequential stopping problem under an existing evidence-acquisition policy and model the state-dependent heterogeneous costs induced by generative video search.
\item We propose \method{}, a plug-in termination framework for long-video reasoning agents. Its learned stopping controller factorizes trajectory-level stopping supervision into future answer-quality advantage and a heterogeneous continuation-cost correction, yielding a lightweight causal policy while keeping the host vision-language model fixed.
\item Across the evaluated hosts, termination exposes substantial removable model-token use: without changing the host search policy or VLM, \method{} reduces model-token use by 54.6\% on average, transfers zero-shot across benchmarks, and achieves the highest Pareto-frontier coverage (83.3\%) among the compared stopping methods. Online execution preserves this trade-off while reducing measured runtime by 54.1\%.
\end{itemize}
\section{Related Work}
\paragraph{Long-video evidence acquisition and efficient processing.} Long-video agents localize question-relevant evidence through visual queries, temporal search, and memory \citep{wang2024videoagent,fan2024videoagentmemory,zhi2025videoagent2}, with newer systems adding specialized search and reasoning tools \citep{zhang2025deepvideodiscovery,pan2026timesearchr,lin2026videoseek,li2026videothinker,liu2026longvideoagent,wang2026activevideoperception}. Other work improves reliability or efficiency through discriminative acquisition, adaptive frame/token allocation, and representation compression \citep{qiu2026videoseal,xu2026pace,wang2025videotree,tang2025aks,zhu2026focus,qi2026adapttoken,jiang2025storm,wang2025adaretake}. These methods primarily decide where or how much to inspect or compress; \method{} leaves acquisition unchanged and asks whether the next proposed search action is worth its state-dependent continuation cost.

\paragraph{Sequential decisions with observation costs.} Sequential hypothesis testing and controlled sensing trade decision error against observation cost, including active and nonuniform observations \citep{wald1945sequential,chernoff1959sequential,naghshvar2013active,nitinawarat2015controlled,vershinin2025nonhomogeneous}. Optimal stopping characterizes when further information acquisition is no longer worthwhile under a given observation process \citep{shiryaev2008optimal,snell1952applications,novikov2010optimal}. \method{} instantiates this perspective for generative video search with state-dependent continuation cost.

\paragraph{Adaptive stopping and decision-focused learning.} Adaptive Computation Time and PonderNet learn input-dependent halting \citep{graves2016act,banino2021pondernet}; BERT Loses Patience, CALM, Fast yet Safe, and Learning To Stop While Learning To Predict use agreement, confidence, calibrated risk, or jointly learned stopping \citep{zhou2020bertpatience,schuster2022calm,jazbec2024fastsafe,chen2020learningstop}. For iterative acquisition, Probing-RAG predicts retrieval need, TASR uses answer agreement and a calibrated margin, and Stop-RAG learns continuation value \citep{baek2025probingrag,kieback2026tasr,park2025stoprag}. CAM-DF selects tool configurations before execution, while LongVideo-R1 jointly trains navigation and halting \citep{feng2026camdf,qiu2026longvideor1}. In contrast, \method{} controls termination of an existing host at a causal checkpoint, using current evidence/execution state and realized downstream quality and cost without modifying the host VLM or search policy.

\section{Method}
\label{sec:method}
\subsection{Sequential stopping along the native search}
\method{} augments an existing tool-guided video agent without changing its evidence-acquisition policy. For question $i$, let $Y_i\in\{1,\ldots,K_i\}$ be the correct option and $\mathcal{F}_{i,t}$ the accumulated evidence, history, and computation available at checkpoint $t$. After the native policy $\pi$ proposes the next visual action but before its execution, \method{} chooses STOP, invoking a constrained prefix finalizer, or CONTINUE, executing that proposal. Complete native trajectories provide training supervision; deployment uses only the current causal state.

This setting follows the fixed-policy stopping principle \citep{shiryaev2008optimal,novikov2010optimal}. Let $R_{\mathrm{stop}}(\mathcal{F}_{i,t})$ be the expected loss of answering now and $Q_{\mathrm{cont}}^{\pi}(\mathcal{F}_{i,t})$ the minimum expected loss achievable by continuing along $\pi$ and stopping later. The ideal value is
\begin{equation}
V^{\pi}(\mathcal{F}_{i,t})=
\min\!\left\{
R_{\mathrm{stop}}(\mathcal{F}_{i,t}),
Q_{\mathrm{cont}}^{\pi}(\mathcal{F}_{i,t})
\right\}.
\label{eq:ideal_stopping}
\end{equation}
The optimal comparison stops when the first term is no larger than the second. Here the continuation cost is state-dependent: the next tool, temporal window, decoded frames, context, and model output are generated as the trajectory unfolds. The stopping decision must therefore value the expected evidence gain together with the full computation needed to obtain it.

\subsection{Targets from complete native trajectories}
The conditional continuation value in Equation~\ref{eq:ideal_stopping} averages over future trajectories; we use the best realized loss on each recorded suffix as a hindsight surrogate (Appendix~\ref{app:theory_hindsight}). Let $a_{i,t}$ and $C_f(i,t)$ be the prefix-finalizer answer and cost at checkpoint $t$, $S$ the mean native training-trajectory cost, and $\lambda$ the computation weight. Stopping now has loss
\begin{equation}
L_{\mathrm{stop}}(i,t)=
\mathbb{1}[a_{i,t}\ne Y_i]
+\lambda\frac{C_f(i,t)}{S}.
\label{eq:stop_loss}
\end{equation}

For a later checkpoint $u$, let $C(i,t\!\rightarrow\!u)$ include native model-token cost and intervening option assessments. Stopping at $u$ has loss
\begin{equation}
L_{\mathrm{future}}(i,t,u)=
\mathbb{1}[a_{i,u}\ne Y_i]
+\lambda\frac{C(i,t\!\rightarrow\!u)+C_f(i,u)}{S}.
\label{eq:future_loss}
\end{equation}
The native terminal point is also a candidate, with loss
\begin{equation}
L_{\mathrm{native}}(i,t)=
\mathbb{1}[a_i^{\mathrm{native}}\ne Y_i]
+\lambda\frac{C(i,t\!\rightarrow\!T_i)}{S}.
\label{eq:native_loss}
\end{equation}
The assessment at checkpoint $t$ is sunk under either decision; subsequent assessments and answer-generation costs remain in the objective.

Let $\mathcal{U}_{i,t}$ contain the later intermediate checkpoints on the same trajectory. Making the cost weight explicit, we define the trajectory-level decision gap as
\begin{equation}
g_{i,t}^{(\lambda)}=L_{\mathrm{stop}}(i,t)
-\min\!\left(
\{L_{\mathrm{future}}(i,t,u):u\in\mathcal{U}_{i,t}\}
\cup\{L_{\mathrm{native}}(i,t)\}
\right),
\label{eq:gap_target}
\end{equation}
A positive gap indicates a better realized later stopping point; a nonpositive gap favors answering now. Future labels and states construct this supervision only, while deployment uses the causal state defined next.

\subsection{Causal stopping state}
At each checkpoint, an external multiple-choice assessment reads the accumulated evidence and returns scores restricted to the question's original options. The assessment is generated at temperature zero, then temperature-scaled by checkpoint group using the training trajectories \cite{guo2017calibration}. The assessment remains external to the native agent context, preserving the native search decisions.

We represent each checkpoint with an 18-dimensional causal vector $x_{i,t}$: eight answer-support statistics (including top probabilities, entropy, margin, and changes in option support), five progress/cost features (checkpoint, consumed model tokens, evidence length, video duration, and subtitles), and five tool-type indicators. Sorting the top probabilities removes option-label dependence, while temporal changes remain aligned by option identity. VideoSeek supplies tool types directly; AVP uses the shared none state for host-specific actions. Every feature is available at or before the current checkpoint.

\subsection{Learning and execution}
The target admits an exact decomposition that makes the role of continuation cost explicit. Let
\begin{equation}
q_{i,t}=g_{i,t}^{(0)},\qquad
c_{i,t}^{(\lambda)}
=\frac{g_{i,t}^{(\lambda)}-g_{i,t}^{(0)}}{\lambda},
\quad \lambda>0,
\label{eq:gap_factorization}
\end{equation}
so that $g_{i,t}^{(\lambda)}=q_{i,t}+\lambda c_{i,t}^{(\lambda)}$. The quality term $q_{i,t}$ compares current answer error with the best later answer error without charging computation. The second term is a cost-induced correction: because the best later point is chosen after answer risk and cost are combined, it captures both heterogeneous continuation cost and any cost-induced change in which later stopping point is preferred.

We fit both terms with the same weighted Ridge operator. For a supervision target $y_{i,t}$, where $y$ is either $q$ or $c^{(\lambda)}$, and trajectory $i$ contains $m_i$ learnable checkpoints, the model solves
\begin{equation}
(\hat w_y,\hat\beta_y)=
\arg\min_{w,\beta}
\sum_i\frac{1}{m_i}\sum_{t=0}^{m_i-1}
\left(y_{i,t}-w^{\top}\widetilde{x}_{i,t}-\beta\right)^2
+\alpha\lVert w\rVert_2^2,
\label{eq:ridge}
\end{equation}
where $\widetilde{x}_{i,t}$ uses weighted training-set standardization and $\alpha=10$. The $1/m_i$ weights equalize total influence across trajectories of different lengths. Regressing the decision gap aligns the learned quantity with the downstream stopping comparison \cite{elmachtoub2022smart}; the vision-language model, host agent, and tools remain fixed.

Writing the two fitted components as $\hat q(x)$ and $\hat c_{\lambda}(x)$, the deployed score is
\begin{equation}
\hat g_{i,t}^{(\lambda)}
=\hat q(x_{i,t})+\lambda\hat c_{\lambda}(x_{i,t}).
\label{eq:factorized_score}
\end{equation}
By linearity of weighted Ridge in its target, this factorized fit is exactly equivalent to fitting $g^{(\lambda)}$ directly, so the two components collapse to one affine deployment score. Appendix~\ref{app:ablations_cost} verifies the equivalence and isolates the cost correction.

At an eligible checkpoint, the decision is
\begin{equation}
d_{i,t}=
\begin{cases}
\textsc{STOP}, & \hat g_{i,t}^{(\lambda)}\le b,\\
\textsc{CONTINUE}, & \hat g_{i,t}^{(\lambda)}>b.
\end{cases}
\label{eq:deployment}
\end{equation}
The cost weight $\lambda$ and boundary $b$ are selected on calibration videos. A STOP decision invokes a constrained prefix finalizer using the same underlying VLM and the evidence accumulated up to the current checkpoint, with decoding restricted to the original answer options. A CONTINUE decision executes the proposed visual action unchanged.

\section{Experiments}
\label{sec:experiments}
\subsection{Setup}
\paragraph{Data and hosts.} We evaluate Video-MME, LongVideoBench, and multiple-choice MLVU \citep{fu2025videomme,wu2024longvideobench,zhou2025mlvu}. Video-MME is split by video into 50/100/750 training/calibration/evaluation videos (150/300/2,250 questions); each agent--model pair fits on the training trajectories, selects its operating point on calibration, and transfers the frozen policy zero-shot to LongVideoBench and MLVU. We pair VideoSeek and AVP with Qwen3.5-35B-A3B-FP8, GLM-4.6V-Flash, and InternVL3.5-30B-A3B-HF. For each host, Native, \method{}, and all stopping baselines share the same host configuration; \method{} intervenes only in termination, leaving native evidence acquisition unchanged. \method{} acts on Medium/Long videos, with Short videos left native, and each agent--model pair has its own Ridge regressor and calibrated operating point.

\paragraph{Selection and accounting.} For \method{}, we search $\lambda\in\{0.05,0.1,0.2,0.4,0.8\}$ and 81 boundaries. The shared selector first requires at least 55\% model-token saving and an accuracy lower confidence bound no worse than $-1$ pp, then maximizes saving among candidates within 2 pp of the best feasible accuracy; if none is feasible, it selects closest to 40\% saving. The same rule applies to all tunable methods (Appendix~\ref{app:baselines}). We report end-to-end accuracy and accuracy change, model-token saving, and visual-token saving relative to Native. Model tokens include text input/output and visual tokens; option assessments, prefix finalizers, and native computation for continuing episodes are fully charged.

\paragraph{Baselines.} We compare with six stopping strategies. \textit{Tuned checkpoint} uses one global checkpoint, while \textit{tuned budget} stops at a cumulative model-token budget. \textit{Top-option confidence} and \textit{option entropy} threshold the external multiple-choice assessment \citep{geifman2019selectivenet,schuster2022calm}. \textit{TASR-Video} adapts TASR to the option-level probe, combining consecutive-answer agreement with a calibrated top-two margin \cite{kieback2026tasr}. \textit{Forward-Q} implements a Monte Carlo continuation baseline following Stop-RAG: separate Ridge predictors estimate current-prefix correctness and the best later correctness, and their difference is thresholded \cite{park2025stoprag}. Each checkpoint, budget, margin, or threshold is selected on Video-MME with the same calibration rule as \method{}. All methods use the same activation schedule, prefix finalizer, and cost ledger.

\subsection{Main results}
Table~\ref{tab:main} reports end-to-end results relative to native search. Under a common-terminal-finalizer control, \method{} saves 55.24\% of model tokens with a 1.92-point accuracy decrease relative to Native with the same constrained terminal finalizer applied at the terminal state (Native+F); Appendix~\ref{app:stopping_attribution} compares the seven stopping methods under this protocol.

\begin{table}[H]
\centering
\tablestandard
\caption{\textbf{Main results.} Model/visual tokens per question (k) and end-to-end accuracy are Native $\rightarrow$ \method{}; savings are relative to Native. Bold marks lower token use or higher accuracy.}
\label{tab:main}
\par\noindent\hspace*{-\tabcolsep}\begin{tabularx}{\dimexpr\textwidth+2\tabcolsep\relax}{lllZQZQZQ}
\tableheadtoprule
\tableheadrow
\tableheadtoppad\tableheadbottompad\textbf{Benchmark} & \textbf{Agent} & \textbf{Model} & \textbf{M tok. (k)} & \textbf{M save} & \textbf{V tok. (k)} & \textbf{V save} & \textbf{Acc. (\%)} & \textbf{$\Delta$ (pp)} \\
\tableheadmidrule
Video-MME & VideoSeek & Qwen3.5 & \tokpair{\VSVQwenNativeTokQ}{\VSVQwenOursTokQ} & \VSVQwenModelSave & \tokpair{\VSVQwenNativeVisualTokQ}{\VSVQwenOursVisualTokQ} & \VSVQwenVisualSave & \mbox{\accpair{\VSVQwenNativeAcc}{\VSVQwenOursAcc}} & \VSVQwenDelta \\
Video-MME & VideoSeek & GLM-4.6V & \tokpair{\VSVGLMNativeTokQ}{\VSVGLMOursTokQ} & \VSVGLMModelSave & \tokpair{\VSVGLMNativeVisualTokQ}{\VSVGLMOursVisualTokQ} & \VSVGLMVisualSave & \mbox{\accpair{\VSVGLMNativeAcc}{\VSVGLMOursAcc}} & \VSVGLMDelta \\
Video-MME & VideoSeek & InternVL3.5 & \tokpair{\VSVInternVLNativeTokQ}{\VSVInternVLOursTokQ} & \VSVInternVLModelSave & \tokpair{\VSVInternVLNativeVisualTokQ}{\VSVInternVLOursVisualTokQ} & \VSVInternVLVisualSave & \mbox{\accpair{\VSVInternVLNativeAcc}{\VSVInternVLOursAcc}} & \VSVInternVLDelta \\
Video-MME & AVP & Qwen3.5 & \tokpair{\AVPVQwenNativeTokQ}{\AVPVQwenOursTokQ} & \AVPVQwenModelSave & \tokpair{\AVPVQwenNativeVisualTokQ}{\AVPVQwenOursVisualTokQ} & \AVPVQwenVisualSave & \mbox{\accpair{\AVPVQwenNativeAcc}{\AVPVQwenOursAcc}} & \AVPVQwenDelta \\
Video-MME & AVP & GLM-4.6V & \tokpair{\AVPVGLMNativeTokQ}{\AVPVGLMOursTokQ} & \AVPVGLMModelSave & \tokpair{\AVPVGLMNativeVisualTokQ}{\AVPVGLMOursVisualTokQ} & \AVPVGLMVisualSave & \mbox{\accpair{\AVPVGLMNativeAcc}{\AVPVGLMOursAcc}} & \AVPVGLMDelta \\
Video-MME & AVP & InternVL3.5 & \tokpair{\AVPVInternVLNativeTokQ}{\AVPVInternVLOursTokQ} & \AVPVInternVLModelSave & \tokpair{\AVPVInternVLNativeVisualTokQ}{\AVPVInternVLOursVisualTokQ} & \AVPVInternVLVisualSave & \mbox{\accpair{\AVPVInternVLNativeAcc}{\AVPVInternVLOursAcc}} & \AVPVInternVLDelta \\
\midrule
LongVideoBench & VideoSeek & Qwen3.5 & \tokpair{\LSVQwenNativeTokQ}{\LSVQwenOursTokQ} & \LSVQwenModelSave & \tokpair{\LSVQwenNativeVisualTokQ}{\LSVQwenOursVisualTokQ} & \LSVQwenVisualSave & \mbox{\accpair{\LSVQwenNativeAcc}{\LSVQwenOursAcc}} & \LSVQwenDelta \\
LongVideoBench & VideoSeek & GLM-4.6V & \tokpair{\LSVGLMNativeTokQ}{\LSVGLMOursTokQ} & \LSVGLMModelSave & \tokpair{\LSVGLMNativeVisualTokQ}{\LSVGLMOursVisualTokQ} & \LSVGLMVisualSave & \mbox{\accpair{\LSVGLMNativeAcc}{\LSVGLMOursAcc}} & \LSVGLMDelta \\
LongVideoBench & VideoSeek & InternVL3.5 & \tokpair{\LSVInternVLNativeTokQ}{\LSVInternVLOursTokQ} & \LSVInternVLModelSave & \tokpair{\LSVInternVLNativeVisualTokQ}{\LSVInternVLOursVisualTokQ} & \LSVInternVLVisualSave & \mbox{\accpair{\LSVInternVLNativeAcc}{\LSVInternVLOursAcc}} & \LSVInternVLDelta \\
LongVideoBench & AVP & Qwen3.5 & \tokpair{\AVPLVQwenNativeTokQ}{\AVPLVQwenOursTokQ} & \AVPLVQwenModelSave & \tokpair{\AVPLVQwenNativeVisualTokQ}{\AVPLVQwenOursVisualTokQ} & \AVPLVQwenVisualSave & \mbox{\accpair{\AVPLVQwenNativeAcc}{\AVPLVQwenOursAcc}} & \AVPLVQwenDelta \\
LongVideoBench & AVP & GLM-4.6V & \tokpair{\AVPLVGLMNativeTokQ}{\AVPLVGLMOursTokQ} & \AVPLVGLMModelSave & \tokpair{\AVPLVGLMNativeVisualTokQ}{\AVPLVGLMOursVisualTokQ} & \AVPLVGLMVisualSave & \mbox{\accpair{\AVPLVGLMNativeAcc}{\AVPLVGLMOursAcc}} & \AVPLVGLMDelta \\
LongVideoBench & AVP & InternVL3.5 & \tokpair{\AVPLVInternVLNativeTokQ}{\AVPLVInternVLOursTokQ} & \AVPLVInternVLModelSave & \tokpair{\AVPLVInternVLNativeVisualTokQ}{\AVPLVInternVLOursVisualTokQ} & \AVPLVInternVLVisualSave & \mbox{\accpair{\AVPLVInternVLNativeAcc}{\AVPLVInternVLOursAcc}} & \AVPLVInternVLDelta \\
\midrule
MLVU & VideoSeek & Qwen3.5 & \tokpair{\MSVQwenNativeTokQ}{\MSVQwenOursTokQ} & \MSVQwenModelSave & \tokpair{\MSVQwenNativeVisualTokQ}{\MSVQwenOursVisualTokQ} & \MSVQwenVisualSave & \mbox{\accpair{\MSVQwenNativeAcc}{\MSVQwenOursAcc}} & \MSVQwenDelta \\
MLVU & VideoSeek & GLM-4.6V & \tokpair{\MSVGLMNativeTokQ}{\MSVGLMOursTokQ} & \MSVGLMModelSave & \tokpair{\MSVGLMNativeVisualTokQ}{\MSVGLMOursVisualTokQ} & \MSVGLMVisualSave & \mbox{\accpair{\MSVGLMNativeAcc}{\MSVGLMOursAcc}} & \MSVGLMDelta \\
MLVU & VideoSeek & InternVL3.5 & \tokpair{\MSVInternVLNativeTokQ}{\MSVInternVLOursTokQ} & \MSVInternVLModelSave & \tokpair{\MSVInternVLNativeVisualTokQ}{\MSVInternVLOursVisualTokQ} & \MSVInternVLVisualSave & \mbox{\accpair{\MSVInternVLNativeAcc}{\MSVInternVLOursAcc}} & \MSVInternVLDelta \\
MLVU & AVP & Qwen3.5 & \tokpair{\MAVPQwenNativeTokQ}{\MAVPQwenOursTokQ} & \MAVPQwenModelSave & \tokpair{\MAVPQwenNativeVisualTokQ}{\MAVPQwenOursVisualTokQ} & \MAVPQwenVisualSave & \mbox{\accpair{\MAVPQwenNativeAcc}{\MAVPQwenOursAcc}} & \MAVPQwenDelta \\
MLVU & AVP & GLM-4.6V & \tokpair{\MAVPGLMNativeTokQ}{\MAVPGLMOursTokQ} & \MAVPGLMModelSave & \tokpair{\MAVPGLMNativeVisualTokQ}{\MAVPGLMOursVisualTokQ} & \MAVPGLMVisualSave & \mbox{\accpair{\MAVPGLMNativeAcc}{\MAVPGLMOursAcc}} & \MAVPGLMDelta \\
MLVU & AVP & InternVL3.5 & \tokpair{\MAVPInternVLNativeTokQ}{\MAVPInternVLOursTokQ} & \MAVPInternVLModelSave & \tokpair{\MAVPInternVLNativeVisualTokQ}{\MAVPInternVLOursVisualTokQ} & \MAVPInternVLVisualSave & \mbox{\accpair{\MAVPInternVLNativeAcc}{\MAVPInternVLOursAcc}} & \MAVPInternVLDelta \\
\bottomrule
\end{tabularx}\hspace*{-\tabcolsep}\par
\end{table}

Across Video-MME, LongVideoBench, and MLVU, \method{} changes end-to-end accuracy by $+0.67$, $+3.38$, and $+4.58$ points while saving 53.63\%, 58.78\%, and 51.28\% of model tokens, respectively; LongVideoBench and MLVU use the same frozen Video-MME policies. Visual-token savings are 60.01\%, 66.66\%, and 59.48\%.

\subsection{Comparison with stopping baselines}
Table~\ref{tab:baselines} compares the six stopping baselines with \method{} under the same Video-MME calibration rule. The two blocks report end-to-end accuracy change and model-token saving for VideoSeek and AVP.

The baselines occupy different parts of the accuracy--saving plane: some favor accuracy, while others stop more aggressively. The next analysis summarizes these two objectives jointly using Pareto dominance.

\input{generated/baselines_table.tex}

\subsection{Accuracy--Efficiency Frontier Analysis}
\label{sec:pareto_frontier}
Inference-efficient systems form quality--compute Pareto frontiers \citep{bian2025inferenceefficient,saadfalcon2025archon}. For evaluation configuration $j$, method $m$ gives $z_{jm}=(\Delta A_{jm},S_{jm})$, where $\Delta A$ is end-to-end accuracy change and $S$ is Model Token saving. A point is Pareto-optimal if no compared method is at least as good in both objectives and strictly better in one.

We summarize how consistently a method reaches this empirical nondominated frontier with \emph{Pareto-frontier coverage} (PFC):
\begin{equation}
\operatorname{PFC}(m)=\frac{1}{J}\sum_{j=1}^{J}\mathbb{1}[m\in\mathcal{P}_j],\qquad J=18,
\label{eq:pfc}
\end{equation}
where $\mathcal{P}_j$ is the nondominated set. We also report mean rank from successive nondominated sorting; rank 1 is best. Table~\ref{tab:pareto_frontier} gives benchmark-wise coverage and ranks.

\input{generated/pareto_frontier_table.tex}
\vspace{-0.55\baselineskip}

\method{} has the highest PFC at 83.3\% (15/18; next-best Tuned budget: 72.2\%), the largest mean model-token saving (54.56\%), and the lowest mean Pareto rank (1.22). It is Pareto-optimal for 6/6 Video-MME, 5/6 LongVideoBench, and 4/6 MLVU agent--model pairs; only five of 108 pairwise CASE--baseline comparisons dominate it on both objectives. Across 10,000 paired video-cluster bootstrap draws with frozen operating points, \method{} has the highest or tied-highest PFC in 97.5\% of draws and is strictly highest in 90.0\% (Appendix~\ref{app:pfc_bootstrap}).

\FloatBarrier
\vspace{-0.35\baselineskip}

\subsection{Long-Trajectory Analysis}
{\setlength{\intextsep}{6pt}%
\begin{figure}[H]
\centering
\includegraphics[width=0.52\textwidth]{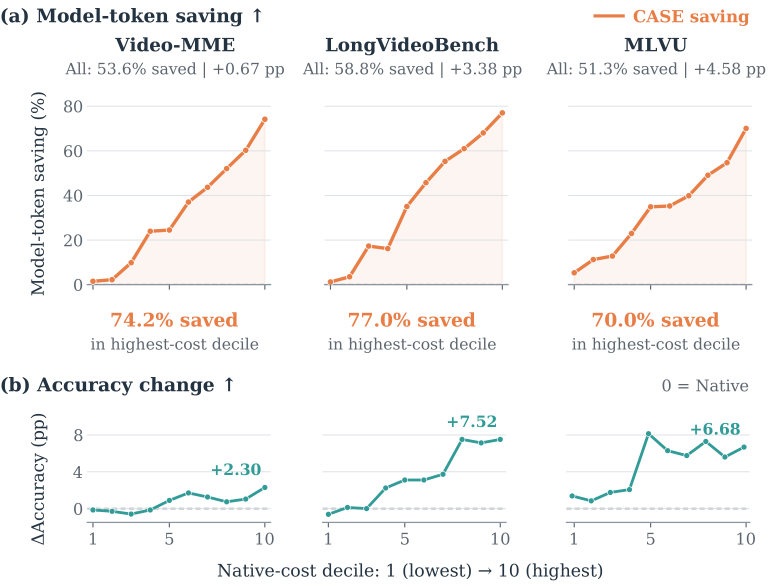}
\caption{\textbf{Savings concentrate on expensive Native trajectories.} Native-cost deciles run from 1 (lowest) to 10 (highest); curves macro-average the six agent--model cells per benchmark.}
\label{fig:long_trajectory_deciles}
\end{figure}
}
The native computation tail is concentrated in a small set of expensive trajectories. Figure~\ref{fig:long_trajectory_deciles} ranks questions by Native model-token cost within each agent--model--benchmark cell and divides them into ten approximately equal deciles. The saving rises sharply toward the expensive end: in the highest-cost decile, \method{} saves 74.2\%, 77.0\%, and 70.0\% of model tokens on Video-MME, LongVideoBench, and MLVU, with end-to-end accuracy changes of $+2.30$, $+7.52$, and $+6.68$ points. The complementary post-hoc Top20\% and Top10\% slices, including all agent--model cells, are reported in Appendix~\ref{app:complete_results_tail}.

\subsection{Ablations}
Table~\ref{tab:ablation} treats each ablation as a complete stopping method, selected on the same Video-MME training/calibration roles and transferred frozen to LongVideoBench and MLVU. Variants remove answer-support features, execution-state features, continuation-aware supervision, or the heterogeneous-cost correction.

{\setlength{\intextsep}{6pt}%
\input{generated/ablation_table.tex}
}

Full gives $+2.88$ points in end-to-end accuracy / 54.56\% model-token saving; without the cost correction it gives $+3.00$/46.00\%, showing that the cost term mainly shifts selection toward greater saving. Appendix~\ref{app:ablations} gives detailed controls.

\vspace{-0.10\baselineskip}
\subsection{Online execution}
In online execution on 900 questions per benchmark and agent--model pair, every STOP changes the subsequent trajectory. The favorable accuracy--efficiency trade-off persists: across all 12 cells, \method{} averages \OnlineOverallMeanDelta{} points with \OnlineOverallMeanModelSave\% model-token saving, while measured runtime additionally falls by \OnlineOverallMeanRuntimeSave\% (Figure~\ref{fig:online_execution}). The close agreement between token and runtime reductions provides an execution-time check on the token-based efficiency metric: Video-MME averages 52.9\% model-token versus 52.5\% runtime saving, and LongVideoBench 56.1\% versus 55.7\%.

{\setlength{\intextsep}{6pt}%
\begin{figure}[H]
\centering
\includegraphics[width=0.81\textwidth]{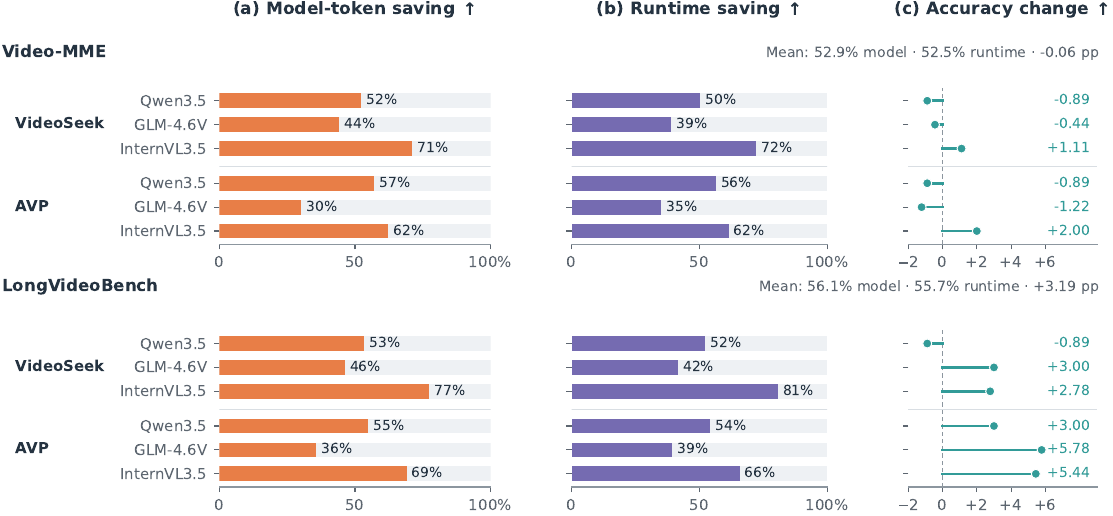}
\caption{\textbf{Online execution.} End-to-end accuracy change and savings in model tokens and measured runtime; exact values are in Appendix Table~\ref{tab:online_execution}.}
\label{fig:online_execution}
\end{figure}
}

\vspace{-0.55\baselineskip}
\section{Conclusion}
Long-video agents must decide both where to seek evidence and when further search is no longer worth its cost. \method{} learns this termination decision from complete native trajectories while leaving the host search policy and VLM unchanged. Across two agents, three VLMs, and three benchmarks, it reduces model-token use by 54.6\% on average and attains the highest PFC (83.3\%) among the compared stopping methods; frozen policies transfer zero-shot across benchmarks. Online execution preserves the same favorable accuracy--efficiency trade-off and additionally reduces measured runtime by 54.1\% on average. Together, these results show that explicit stopping is a practical complement to better evidence acquisition: long-video agents can retain their native search policy while learning when further search is no longer worth its cost.

\section*{AI Use Statement}

We used generative AI tools to assist with literature discovery, experimental design, implementation and debugging, and manuscript polishing.

We did not use generative AI to generate synthetic training or evaluation datasets. The experimental results reported in this paper were obtained from executed experiments and analyses of their outputs.

AI-assisted implementations were tested, numerical claims were checked against experimental artifacts, and references were verified against primary sources.

\bibliographystyle{iclr2027_conference}
\bibliography{references}
\appendix
\section*{Appendix}
\input{appendix/appendix.tex}

\end{document}

%% file: math_commands.tex
\usepackage{amsmath,amsfonts,bm}

\def\eqref#1{equation~\ref{#1}}

\def\1{\bm{1}}

\DeclareMathAlphabet{\mathsfit}{\encodingdefault}{\sfdefault}{m}{sl}
\SetMathAlphabet{\mathsfit}{bold}{\encodingdefault}{\sfdefault}{bx}{n}



%% file: generated/results_macros.tex
\newcommand{\VSVQwenNativeAcc}{51.82}
\newcommand{\VSVQwenOursAcc}{53.73}
\newcommand{\VSVQwenDelta}{+1.91}
\newcommand{\VSVQwenModelSave}{52.43}
\newcommand{\VSVQwenVisualSave}{57.64}
\newcommand{\VSVQwenNativeTokQ}{57.2}
\newcommand{\VSVQwenOursTokQ}{27.2}
\newcommand{\VSVQwenNativeVisualTokQ}{12.0}
\newcommand{\VSVQwenOursVisualTokQ}{5.1}
\newcommand{\LSVQwenNativeAcc}{40.91}
\newcommand{\LSVQwenOursAcc}{44.13}
\newcommand{\LSVQwenDelta}{+3.22}
\newcommand{\LSVQwenModelSave}{64.24}
\newcommand{\LSVQwenVisualSave}{64.03}
\newcommand{\LSVQwenNativeTokQ}{79.5}
\newcommand{\LSVQwenOursTokQ}{28.4}
\newcommand{\LSVQwenNativeVisualTokQ}{16.9}
\newcommand{\LSVQwenOursVisualTokQ}{6.1}
\newcommand{\MSVQwenNativeAcc}{45.58}
\newcommand{\MSVQwenOursAcc}{47.88}
\newcommand{\MSVQwenDelta}{+2.30}
\newcommand{\MSVQwenModelSave}{59.46}
\newcommand{\MSVQwenVisualSave}{69.53}
\newcommand{\MSVQwenNativeTokQ}{47.7}
\newcommand{\MSVQwenOursTokQ}{19.3}
\newcommand{\MSVQwenNativeVisualTokQ}{15.0}
\newcommand{\MSVQwenOursVisualTokQ}{4.6}
\newcommand{\VSVGLMNativeAcc}{56.40}
\newcommand{\VSVGLMOursAcc}{55.69}
\newcommand{\VSVGLMDelta}{-0.71}
\newcommand{\VSVGLMModelSave}{49.77}
\newcommand{\VSVGLMVisualSave}{58.07}
\newcommand{\VSVGLMNativeTokQ}{64.3}
\newcommand{\VSVGLMOursTokQ}{32.3}
\newcommand{\VSVGLMNativeVisualTokQ}{22.4}
\newcommand{\VSVGLMOursVisualTokQ}{9.4}
\newcommand{\LSVGLMNativeAcc}{47.49}
\newcommand{\LSVGLMOursAcc}{47.34}
\newcommand{\LSVGLMDelta}{-0.15}
\newcommand{\LSVGLMModelSave}{54.87}
\newcommand{\LSVGLMVisualSave}{61.39}
\newcommand{\LSVGLMNativeTokQ}{98.3}
\newcommand{\LSVGLMOursTokQ}{44.4}
\newcommand{\LSVGLMNativeVisualTokQ}{35.8}
\newcommand{\LSVGLMOursVisualTokQ}{13.8}
\newcommand{\MSVGLMNativeAcc}{50.23}
\newcommand{\MSVGLMOursAcc}{49.95}
\newcommand{\MSVGLMDelta}{-0.28}
\newcommand{\MSVGLMModelSave}{45.52}
\newcommand{\MSVGLMVisualSave}{53.51}
\newcommand{\MSVGLMNativeTokQ}{65.9}
\newcommand{\MSVGLMOursTokQ}{35.9}
\newcommand{\MSVGLMNativeVisualTokQ}{28.4}
\newcommand{\MSVGLMOursVisualTokQ}{13.2}
\newcommand{\VSVInternVLNativeAcc}{52.84}
\newcommand{\VSVInternVLOursAcc}{52.67}
\newcommand{\VSVInternVLDelta}{-0.18}
\newcommand{\VSVInternVLModelSave}{69.85}
\newcommand{\VSVInternVLVisualSave}{77.06}
\newcommand{\VSVInternVLNativeTokQ}{49.5}
\newcommand{\VSVInternVLOursTokQ}{14.9}
\newcommand{\VSVInternVLNativeVisualTokQ}{19.0}
\newcommand{\VSVInternVLOursVisualTokQ}{4.4}
\newcommand{\LSVInternVLNativeAcc}{42.11}
\newcommand{\LSVInternVLOursAcc}{46.00}
\newcommand{\LSVInternVLDelta}{+3.89}
\newcommand{\LSVInternVLModelSave}{74.08}
\newcommand{\LSVInternVLVisualSave}{85.39}
\newcommand{\LSVInternVLNativeTokQ}{48.1}
\newcommand{\LSVInternVLOursTokQ}{12.5}
\newcommand{\LSVInternVLNativeVisualTokQ}{21.2}
\newcommand{\LSVInternVLOursVisualTokQ}{3.1}
\newcommand{\MSVInternVLNativeAcc}{44.99}
\newcommand{\MSVInternVLOursAcc}{50.87}
\newcommand{\MSVInternVLDelta}{+5.89}
\newcommand{\MSVInternVLModelSave}{54.20}
\newcommand{\MSVInternVLVisualSave}{59.96}
\newcommand{\MSVInternVLNativeTokQ}{68.5}
\newcommand{\MSVInternVLOursTokQ}{31.4}
\newcommand{\MSVInternVLNativeVisualTokQ}{48.2}
\newcommand{\MSVInternVLOursVisualTokQ}{19.3}
\newcommand{\AVPVQwenNativeAcc}{65.02}
\newcommand{\AVPVQwenOursAcc}{65.29}
\newcommand{\AVPVQwenDelta}{+0.27}
\newcommand{\AVPVQwenModelSave}{56.67}
\newcommand{\AVPVQwenVisualSave}{69.67}
\newcommand{\AVPVQwenNativeTokQ}{51.9}
\newcommand{\AVPVQwenOursTokQ}{22.5}
\newcommand{\AVPVQwenNativeVisualTokQ}{14.9}
\newcommand{\AVPVQwenOursVisualTokQ}{4.5}
\newcommand{\AVPLVQwenNativeAcc}{59.01}
\newcommand{\AVPLVQwenOursAcc}{61.71}
\newcommand{\AVPLVQwenDelta}{+2.69}
\newcommand{\AVPLVQwenModelSave}{55.83}
\newcommand{\AVPLVQwenVisualSave}{71.40}
\newcommand{\AVPLVQwenNativeTokQ}{55.3}
\newcommand{\AVPLVQwenOursTokQ}{24.4}
\newcommand{\AVPLVQwenNativeVisualTokQ}{11.7}
\newcommand{\AVPLVQwenOursVisualTokQ}{3.4}
\newcommand{\MAVPQwenNativeAcc}{55.43}
\newcommand{\MAVPQwenOursAcc}{59.89}
\newcommand{\MAVPQwenDelta}{+4.46}
\newcommand{\MAVPQwenModelSave}{65.70}
\newcommand{\MAVPQwenVisualSave}{81.88}
\newcommand{\MAVPQwenNativeTokQ}{36.4}
\newcommand{\MAVPQwenOursTokQ}{12.5}
\newcommand{\MAVPQwenNativeVisualTokQ}{12.6}
\newcommand{\MAVPQwenOursVisualTokQ}{2.3}
\newcommand{\AVPVGLMNativeAcc}{57.91}
\newcommand{\AVPVGLMOursAcc}{57.51}
\newcommand{\AVPVGLMDelta}{-0.40}
\newcommand{\AVPVGLMModelSave}{30.72}
\newcommand{\AVPVGLMVisualSave}{32.36}
\newcommand{\AVPVGLMNativeTokQ}{55.5}
\newcommand{\AVPVGLMOursTokQ}{38.5}
\newcommand{\AVPVGLMNativeVisualTokQ}{14.6}
\newcommand{\AVPVGLMOursVisualTokQ}{9.9}
\newcommand{\AVPLVGLMNativeAcc}{51.31}
\newcommand{\AVPLVGLMOursAcc}{54.38}
\newcommand{\AVPLVGLMDelta}{+3.07}
\newcommand{\AVPLVGLMModelSave}{35.43}
\newcommand{\AVPLVGLMVisualSave}{42.33}
\newcommand{\AVPLVGLMNativeTokQ}{62.2}
\newcommand{\AVPLVGLMOursTokQ}{40.2}
\newcommand{\AVPLVGLMNativeVisualTokQ}{10.8}
\newcommand{\AVPLVGLMOursVisualTokQ}{6.2}
\newcommand{\MAVPGLMNativeAcc}{53.13}
\newcommand{\MAVPGLMOursAcc}{53.54}
\newcommand{\MAVPGLMDelta}{+0.41}
\newcommand{\MAVPGLMModelSave}{31.19}
\newcommand{\MAVPGLMVisualSave}{38.73}
\newcommand{\MAVPGLMNativeTokQ}{43.5}
\newcommand{\MAVPGLMOursTokQ}{29.9}
\newcommand{\MAVPGLMNativeVisualTokQ}{18.7}
\newcommand{\MAVPGLMOursVisualTokQ}{11.5}
\newcommand{\AVPVInternVLNativeAcc}{51.11}
\newcommand{\AVPVInternVLOursAcc}{54.27}
\newcommand{\AVPVInternVLDelta}{+3.16}
\newcommand{\AVPVInternVLModelSave}{62.35}
\newcommand{\AVPVInternVLVisualSave}{65.26}
\newcommand{\AVPVInternVLNativeTokQ}{81.5}
\newcommand{\AVPVInternVLOursTokQ}{30.7}
\newcommand{\AVPVInternVLNativeVisualTokQ}{49.7}
\newcommand{\AVPVInternVLOursVisualTokQ}{17.3}
\newcommand{\AVPLVInternVLNativeAcc}{38.22}
\newcommand{\AVPLVInternVLOursAcc}{45.77}
\newcommand{\AVPLVInternVLDelta}{+7.55}
\newcommand{\AVPLVInternVLModelSave}{68.23}
\newcommand{\AVPLVInternVLVisualSave}{75.43}
\newcommand{\AVPLVInternVLNativeTokQ}{79.4}
\newcommand{\AVPLVInternVLOursTokQ}{25.2}
\newcommand{\AVPLVInternVLNativeVisualTokQ}{39.7}
\newcommand{\AVPLVInternVLOursVisualTokQ}{9.7}
\newcommand{\MAVPInternVLNativeAcc}{34.08}
\newcommand{\MAVPInternVLOursAcc}{48.76}
\newcommand{\MAVPInternVLDelta}{+14.67}
\newcommand{\MAVPInternVLModelSave}{51.58}
\newcommand{\MAVPInternVLVisualSave}{53.26}
\newcommand{\MAVPInternVLNativeTokQ}{49.0}
\newcommand{\MAVPInternVLOursTokQ}{23.7}
\newcommand{\MAVPInternVLNativeVisualTokQ}{33.8}
\newcommand{\MAVPInternVLOursVisualTokQ}{15.8}

%% file: generated/online_macros.tex
\newcommand{\OnlineOverallMeanDelta}{+1.56}
\newcommand{\OnlineOverallMeanModelSave}{54.50}

\newcommand{\OnlineOverallMeanRuntimeSave}{54.10}

%% file: generated/baselines_table.tex
\begin{table}[!t]
\centering
\tablestandard
\caption{\textbf{Stopping baselines.} All methods use the same Video-MME calibration rule. Entries are end-to-end accuracy change (\(\Delta\), pp), model-token saving (M, \%), and visual-token saving (V, \%). LVB and MLVU use the frozen Video-MME operating point; bold marks the best value, underlining the second-best, and blue shading marks \method{}.}
\label{tab:baselines}
\par\noindent\hspace*{-\tabcolsep}\begin{tabularx}{\dimexpr\textwidth+2\tabcolsep\relax}{llYYYYYYYYY}
\multicolumn{11}{l}{\textbf{(a) VideoSeek}} \\
\tableheadtoprule
\tableheadrowspan
\tableheadtoppad \textbf{Model} & \textbf{Method} & \multicolumn{3}{c}{\textbf{Video-MME}} & \multicolumn{3}{c}{\textbf{LongVideoBench}} & \multicolumn{3}{c}{\textbf{MLVU}} \\
\tableheadrow
\tableheadbottompad  &  & \textbf{M} & \textbf{V} & \textbf{$\Delta$} & \textbf{M} & \textbf{V} & \textbf{$\Delta$} & \textbf{M} & \textbf{V} & \textbf{$\Delta$} \\
\tableheadmidrule
Qwen3.5 & Tuned checkpoint & \underline{42.57} & 42.17 & \underline{+2.00} & \underline{57.15} & 52.11 & \underline{+3.89} & \underline{59.14} & 62.99 & +2.39 \\
Qwen3.5 & Tuned budget & 36.73 & 34.58 & \textbf{+2.36} & 51.99 & 45.10 & \textbf{+4.19} & 42.34 & 48.70 & +2.16 \\
Qwen3.5 & Confidence & 36.13 & 42.66 & +1.69 & 37.57 & 39.57 & +2.69 & 48.31 & 57.58 & +2.21 \\
Qwen3.5 & Entropy & 33.72 & 40.13 & +1.51 & 42.52 & 44.58 & +2.09 & 43.84 & 53.71 & \underline{+2.58} \\
Qwen3.5 & TASR-Video & 37.65 & \underline{46.10} & +1.56 & 52.77 & \underline{54.10} & +3.29 & 52.79 & \underline{65.26} & \textbf{+3.36} \\
Qwen3.5 & Forward-Q & 38.64 & 43.25 & +0.93 & 50.47 & 49.59 & +3.74 & 32.91 & 43.63 & +2.44 \\
\oursrow Qwen3.5 & CASE & \textbf{52.43} & \textbf{57.64} & +1.91 & \textbf{64.24} & \textbf{64.03} & +3.22 & \textbf{59.46} & \textbf{69.53} & +2.30 \\
\addlinespace[0.4pt]
GLM-4.6V & Tuned checkpoint & 30.51 & 29.92 & \underline{+0.22} & 32.82 & 31.95 & +0.15 & 31.77 & 30.95 & \textbf{+0.55} \\
GLM-4.6V & Tuned budget & 32.36 & 33.06 & \textbf{+0.36} & 35.14 & 34.54 & \textbf{+0.45} & 31.48 & 36.97 & +0.41 \\
GLM-4.6V & Confidence & 34.64 & 43.10 & $-$2.09 & 32.51 & 36.21 & $-$2.09 & \textbf{56.83} & \textbf{62.83} & $-$5.66 \\
GLM-4.6V & Entropy & 32.53 & 42.61 & $-$1.82 & 36.92 & 40.14 & $-$2.99 & \underline{51.06} & 56.90 & $-$5.24 \\
GLM-4.6V & TASR-Video & 26.86 & 42.77 & $-$0.36 & 37.75 & 50.59 & $-$0.75 & 44.08 & \underline{58.41} & $-$0.18 \\
GLM-4.6V & Forward-Q & \underline{43.49} & \underline{50.08} & $-$0.13 & \underline{50.19} & \underline{56.58} & \underline{+0.22} & 32.00 & 39.67 & \underline{+0.46} \\
\oursrow GLM-4.6V & CASE & \textbf{49.77} & \textbf{58.07} & $-$0.71 & \textbf{54.87} & \textbf{61.39} & $-$0.15 & 45.52 & 53.51 & $-$0.28 \\
\addlinespace[0.4pt]
InternVL3.5 & Tuned checkpoint & 44.72 & 41.88 & +0.80 & 36.23 & 34.31 & +3.59 & 38.84 & 40.72 & +3.08 \\
InternVL3.5 & Tuned budget & 28.01 & 20.86 & +0.13 & 18.70 & 14.37 & +0.22 & 21.74 & 21.94 & +1.10 \\
InternVL3.5 & Confidence & 33.90 & 38.85 & +0.18 & 10.71 & 21.03 & +2.47 & 35.87 & 40.06 & +2.71 \\
InternVL3.5 & Entropy & 30.91 & 35.50 & +0.22 & 12.13 & 21.64 & +2.69 & 30.29 & 34.14 & +2.02 \\
InternVL3.5 & TASR-Video & 48.30 & 55.65 & \textbf{+1.60} & 38.36 & 50.93 & \textbf{+4.71} & \underline{52.17} & \textbf{60.69} & \underline{+5.47} \\
InternVL3.5 & Forward-Q & \underline{54.53} & \underline{59.96} & \underline{+1.51} & \underline{46.82} & \underline{54.88} & \underline{+4.04} & 35.38 & 40.29 & +3.40 \\
\oursrow InternVL3.5 & CASE & \textbf{69.85} & \textbf{77.06} & $-$0.18 & \textbf{74.08} & \textbf{85.39} & +3.89 & \textbf{54.20} & \underline{59.96} & \textbf{+5.89} \\
\bottomrule
\end{tabularx}
\vspace{2pt}
\par\noindent\hspace*{-\tabcolsep}\begin{tabularx}{\dimexpr\textwidth+2\tabcolsep\relax}{llYYYYYYYYY}
\multicolumn{11}{l}{\textbf{(b) AVP}} \\
\tableheadtoprule
\tableheadrowspan
\tableheadtoppad \textbf{Model} & \textbf{Method} & \multicolumn{3}{c}{\textbf{Video-MME}} & \multicolumn{3}{c}{\textbf{LongVideoBench}} & \multicolumn{3}{c}{\textbf{MLVU}} \\
\tableheadrow
\tableheadbottompad  &  & \textbf{M} & \textbf{V} & \textbf{$\Delta$} & \textbf{M} & \textbf{V} & \textbf{$\Delta$} & \textbf{M} & \textbf{V} & \textbf{$\Delta$} \\
\tableheadmidrule
Qwen3.5 & Tuned checkpoint & 25.27 & 32.53 & \textbf{+1.96} & 23.75 & 32.47 & \textbf{+4.41} & 31.23 & 40.49 & \textbf{+5.75} \\
Qwen3.5 & Tuned budget & \textbf{64.11} & \textbf{73.24} & $-$0.49 & \textbf{72.06} & \textbf{80.59} & $-$2.54 & \textbf{71.89} & \textbf{87.01} & +4.78 \\
Qwen3.5 & Confidence & 28.04 & 38.97 & $-$0.76 & 9.49 & 24.66 & +0.07 & 28.99 & 39.89 & $-$0.64 \\
Qwen3.5 & Entropy & 29.59 & 40.23 & $-$1.11 & 13.30 & 28.26 & +0.00 & 29.57 & 40.45 & $-$0.51 \\
Qwen3.5 & TASR-Video & 45.72 & 61.45 & \underline{+0.76} & 39.85 & 60.66 & \underline{+3.66} & 57.53 & 74.50 & \underline{+5.47} \\
Qwen3.5 & Forward-Q & 54.11 & 67.36 & +0.49 & 53.15 & 68.72 & +2.99 & 61.00 & 77.04 & +4.65 \\
\oursrow Qwen3.5 & CASE & \underline{56.67} & \underline{69.67} & +0.27 & \underline{55.83} & \underline{71.40} & +2.69 & \underline{65.70} & \underline{81.88} & +4.46 \\
\addlinespace[0.4pt]
GLM-4.6V & Tuned checkpoint & 23.01 & 26.06 & \underline{$-$0.13} & 24.13 & 27.77 & \textbf{+3.96} & 33.01 & 38.07 & \underline{+1.70} \\
GLM-4.6V & Tuned budget & 30.00 & 29.02 & \textbf{+0.09} & \textbf{39.05} & 40.06 & \underline{+3.66} & 26.57 & 32.41 & \textbf{+1.89} \\
GLM-4.6V & Confidence & 27.24 & 32.71 & $-$0.71 & 21.04 & 29.52 & +1.65 & \underline{44.83} & 52.11 & +0.28 \\
GLM-4.6V & Entropy & 27.79 & \underline{33.22} & $-$0.80 & 25.15 & 32.98 & +1.65 & \textbf{45.72} & \underline{53.13} & $-$0.23 \\
GLM-4.6V & TASR-Video & 24.41 & \textbf{34.00} & $-$0.40 & 20.43 & 34.52 & +1.80 & 43.48 & \textbf{53.89} & +1.15 \\
GLM-4.6V & Forward-Q & \textbf{30.76} & 32.60 & $-$0.44 & 35.36 & \underline{42.33} & +3.07 & 32.92 & 40.56 & +0.55 \\
\oursrow GLM-4.6V & CASE & \underline{30.72} & 32.36 & $-$0.40 & \underline{35.43} & \textbf{42.33} & +3.07 & 31.19 & 38.73 & +0.41 \\
\addlinespace[0.4pt]
InternVL3.5 & Tuned checkpoint & 54.30 & 56.13 & \textbf{+4.04} & 58.66 & 62.90 & \underline{+8.90} & \underline{56.56} & \underline{56.75} & +14.03 \\
InternVL3.5 & Tuned budget & \textbf{67.70} & \textbf{69.63} & +1.69 & \textbf{78.13} & \textbf{84.14} & +6.58 & \textbf{61.49} & \textbf{62.76} & +14.12 \\
InternVL3.5 & Confidence & 28.35 & 32.40 & +0.58 & 14.36 & 22.33 & +3.89 & 29.88 & 33.64 & +7.27 \\
InternVL3.5 & Entropy & 25.68 & 29.70 & +0.98 & 16.34 & 24.45 & +3.89 & 26.08 & 29.73 & +5.75 \\
InternVL3.5 & TASR-Video & 46.07 & 51.06 & \underline{+3.87} & 48.28 & 58.47 & \textbf{+9.65} & 48.30 & 50.36 & \underline{+14.54} \\
InternVL3.5 & Forward-Q & 59.85 & 62.91 & +3.73 & 64.44 & 71.82 & +7.93 & 51.90 & 53.81 & +14.49 \\
\oursrow InternVL3.5 & CASE & \underline{62.35} & \underline{65.26} & +3.16 & \underline{68.23} & \underline{75.43} & +7.55 & 51.58 & 53.26 & \textbf{+14.67} \\
\bottomrule
\end{tabularx}
\end{table}

%% file: generated/pareto_frontier_table.tex
\begin{table}[!t]
\centering
\tablestandard
\caption{\textbf{Accuracy--efficiency frontier summary.} Each benchmark contains six agent--model configurations. A method is Pareto-optimal when no compared method jointly improves accuracy change and Model Token saving. PFC is the percentage of all 18 configurations in which the method is Pareto-optimal; Pareto rank is obtained by successive nondominated sorting (lower is better). Bold marks the best value and underlining the second-best.}
\label{tab:pareto_frontier}
\par\noindent\hspace*{-\tabcolsep}\begin{tabularx}{\dimexpr\textwidth+2\tabcolsep\relax}{lYYYYYYYY}
\tableheadtoprule
\tableheadrowspan
\tableheadtoppad \textbf{Method} & \textbf{Mean M} & \textbf{Mean $\Delta$} & \multicolumn{3}{c}{\textbf{Pareto-optimal agent--model pairs}} & \textbf{Overall} & \textbf{PFC} & \textbf{Mean rank} \\
\tableheadrow
\tableheadbottompad  & \textbf{(\%)} & \textbf{(pp)} & \textbf{VMM} & \textbf{LVB} & \textbf{MLVU} & \textbf{/18} & \textbf{(\%)} & \textbf{$\downarrow$} \\
\tableheadmidrule
Tuned checkpoint & 39.09 & \textbf{+3.41} & 3/6 & \underline{4/6} & \textbf{4/6} & 11/18 & 61.1 & \underline{1.44} \\
Tuned budget & 44.97 & +2.29 & \underline{5/6} & \textbf{5/6} & \underline{3/6} & \underline{13/18} & \underline{72.2} & 1.72 \\
Confidence & 31.04 & +0.76 & 0/6 & 0/6 & 2/6 & 2/18 & 11.1 & 2.72 \\
Entropy & 30.73 & +0.59 & 0/6 & 0/6 & 2/6 & 2/18 & 11.1 & 2.67 \\
TASR-Video & 42.49 & \underline{+3.29} & 2/6 & 3/6 & \textbf{4/6} & 9/18 & 50.0 & 1.67 \\
Forward-Q & \underline{46.00} & +3.00 & \underline{5/6} & \underline{4/6} & 2/6 & 11/18 & 61.1 & 1.56 \\
\oursrow CASE & \textbf{54.56} & +2.88 & \textbf{6/6} & \textbf{5/6} & \textbf{4/6} & \textbf{15/18} & \textbf{83.3} & \textbf{1.22} \\
\bottomrule
\end{tabularx}\hspace*{-\tabcolsep}\par
\end{table}

%% file: generated/ablation_table.tex
\begin{table}[H]
\centering
\tablestandard
\caption{\textbf{End-to-end component and target ablations.} Variants are independently selected on Video-MME calibration and transferred zero-shot to LongVideoBench and MLVU. Entries are mean accuracy change ($\Delta$, pp) and model-token saving (M, \%) over six agent--model cells; Overall averages all 18 configurations and Strict counts primary-branch selections (/6). Bold/underline mark best/second-best.}
\label{tab:ablation}
\par\noindent\hspace*{-\tabcolsep}\begin{tabularx}{\dimexpr\textwidth+2\tabcolsep\relax}{lYYYYYYYYY}
\tableheadtoprule
\tableheadrowspan
\tableheadtoppad \textbf{Variant} & \multicolumn{2}{c}{\textbf{Video-MME}} & \multicolumn{2}{c}{\textbf{LongVideoBench}} & \multicolumn{2}{c}{\textbf{MLVU}} & \multicolumn{2}{c}{\textbf{Overall}} & \textbf{Strict} \\
\tableheadrow
\tableheadbottompad  & \textbf{M} & \textbf{$\Delta$} & \textbf{M} & \textbf{$\Delta$} & \textbf{M} & \textbf{$\Delta$} & \textbf{M} & \textbf{$\Delta$} & \textbf{/6} \\
\tableheadmidrule
\oursrow Full & \underline{53.63} & \underline{+0.67} & \underline{58.78} & \underline{+3.38} & \textbf{51.28} & \textbf{+4.58} & \underline{54.56} & \underline{+2.88} & \textbf{5/6} \\
Belief-only & 36.90 & +0.38 & 33.46 & +2.49 & \underline{49.28} & +2.70 & 39.88 & +1.86 & 2/6 \\
Exec-state-only & \textbf{55.21} & +0.09 & \textbf{64.48} & +2.26 & 46.20 & +4.27 & \textbf{55.30} & +2.21 & \underline{4/6} \\
Current-corr. & 35.99 & $-$0.19 & 31.55 & +2.02 & 40.57 & +1.34 & 36.04 & +1.06 & 1/6 \\
w/o cost correction & 46.90 & \textbf{+1.01} & 50.07 & \textbf{+3.66} & 41.02 & \underline{+4.33} & 46.00 & \textbf{+3.00} & \underline{4/6} \\
\bottomrule
\end{tabularx}
\end{table}

%% file: appendix/appendix.tex
\raggedbottom
\section{Formal Formulation, Properties, and Proofs}
\label{app:theory}
We expand the stopping formulation in Section~\ref{sec:method} and connect the trajectory-level target used by \method{} to the corresponding policy-conditioned stopping problem. Throughout this section, the native evidence-acquisition policy is fixed and denoted by $\pi$; the stopping policy only chooses when to terminate the resulting evidence-acquisition process.

\subsection{Policy-conditioned stopping problem}
\label{app:theory_problem}
For question $i$, let $Y_i\in\{1,\ldots,K_i\}$ be the correct option and let $\{\mathcal F_{i,t}\}_{t=0}^{T_i}$ be the information available along the native trajectory. The filtration grows as video-tool observations, interaction history, and consumed computation accumulate. A causal stopping time $\tau$ satisfies $\{\tau=t\}\in\mathcal F_{i,t}$ and may terminate at an intermediate checkpoint or at the native terminal state.

For an intermediate stop, $A_{i,t}$ denotes the prefix-finalizer answer. At the native terminal state it denotes the native answer. Let $C_i(t,\tau)$ be the incremental model-token cost from checkpoint $t$ until answering at $\tau$: it contains all native computation after $t$, every future option assessment, and the answer-generation cost at the stopping point. The conditional loss of a stopping time is
\begin{equation}
J_{i,t}^{\pi}(\tau)
=
\mathbb E\!\left[
\mathbb 1[A_{i,\tau}\ne Y_i]
+\lambda\frac{C_i(t,\tau)}{S}
\,\middle|\,\mathcal F_{i,t}
\right],
\label{eq:app_stopping_objective}
\end{equation}
where the expectation follows future trajectories generated by $\pi$. If $\mathcal T_{i,t}^{\pi}$ is the set of causal stopping times no earlier than $t$, the policy-conditioned value is
\begin{equation}
V_{i,t}^{\pi}
=
\inf_{\tau\in\mathcal T_{i,t}^{\pi}} J_{i,t}^{\pi}(\tau).
\label{eq:app_value}
\end{equation}
This is the fixed-policy optimal-stopping problem underlying Equation~\ref{eq:ideal_stopping} \citep{shiryaev2008optimal,novikov2010optimal}. The multiple-choice structure determines the answer error, while the native agent induces the sequence of future evidence and computation on which the stopping decision operates.

\subsection{Finite-horizon Bellman relation}
\label{app:theory_bellman}
Let $K_{i,t+1}$ denote the complete incremental cost incurred by choosing CONTINUE at checkpoint $t$ and reaching the next decision state. It includes the native model/tool computation and the next option assessment. Because the next tool, temporal window, decoded frames, and generated context depend on the evolving agent state, $K_{i,t+1}$ is generally state dependent and random conditional on $\mathcal F_{i,t}$.

\paragraph{Proposition 1 (policy-conditioned Bellman recursion).}
For a finite native trajectory, the value in Equation~\ref{eq:app_value} satisfies
\begin{equation}
V_{i,T_i}^{\pi}=R_{\mathrm{stop}}(\mathcal F_{i,T_i}),
\qquad
V_{i,t}^{\pi}
=
\min\!\left\{
R_{\mathrm{stop}}(\mathcal F_{i,t}),
\mathbb E\!\left[
\lambda\frac{K_{i,t+1}}{S}+V_{i,t+1}^{\pi}
\,\middle|\,\mathcal F_{i,t}
\right]
\right\}.
\label{eq:app_bellman}
\end{equation}
Consequently, if
\begin{equation}
G_{i,t}^{\pi}
=
R_{\mathrm{stop}}(\mathcal F_{i,t})
-
\mathbb E\!\left[
\lambda\frac{K_{i,t+1}}{S}+V_{i,t+1}^{\pi}
\,\middle|\,\mathcal F_{i,t}
\right],
\label{eq:app_ideal_gap}
\end{equation}
then the ideal decision stops when $G_{i,t}^{\pi}\le 0$ and continues when $G_{i,t}^{\pi}>0$.

\paragraph{Proof.}
At the terminal state there is no continuation action, giving the first equality. At any earlier checkpoint, every admissible stopping time either stops immediately or chooses CONTINUE and subsequently follows an admissible stopping rule from the next state. The immediate branch has conditional risk $R_{\mathrm{stop}}(\mathcal F_{i,t})$; the continuation branch first pays $K_{i,t+1}$ and then incurs the optimal value from $t+1$. Taking the smaller conditional risk gives Equation~\ref{eq:app_bellman}. Equation~\ref{eq:app_ideal_gap} is the difference between these two branches, so its sign determines the minimizing action. \hfill$\square$

The relation shows that stopping depends jointly on answer support, continuation cost, and the future evidence reachable from the current state. Two states with similar answer support can therefore have different continuation values when the next evidence-acquisition step has different expected cost or reaches different future evidence.

\subsection{Reachable-suffix hindsight target}
\label{app:theory_hindsight}
The conditional continuation value in Equation~\ref{eq:app_bellman} averages over future trajectories, whereas training provides complete realized trajectories. For a recorded trajectory, define the best later realized loss
\begin{equation}
H_{i,t}
=
\min\!\left(
\{L_{\mathrm{future}}(i,t,u):u\in\mathcal U_{i,t}\}
\cup
\{L_{\mathrm{native}}(i,t)\}
\right).
\label{eq:app_hindsight}
\end{equation}
The training target in Equation~\ref{eq:gap_target} is therefore $g_{i,t}^{(\lambda)}=L_{\mathrm{stop}}(i,t)-H_{i,t}$. A positive target means that the observed suffix contains a later answer point with lower realized error-plus-cost loss.

\paragraph{Proposition 2 (hindsight lower bound).}
Let $\tau>t$ be any causal stopping rule restricted to the later checkpoints and native terminal state reachable on the recorded suffix. Then
\begin{equation}
\mathbb E[H_{i,t}\mid\mathcal F_{i,t}]
\le
\mathbb E[L_{i,t}(\tau)\mid\mathcal F_{i,t}],
\label{eq:app_hindsight_bound_one}
\end{equation}
and hence
\begin{equation}
\mathbb E[H_{i,t}\mid\mathcal F_{i,t}]
\le
\inf_{\tau\in\mathcal T_{i,t+1}^{\pi}}
\mathbb E[L_{i,t}(\tau)\mid\mathcal F_{i,t}].
\label{eq:app_hindsight_bound}
\end{equation}

\paragraph{Proof.}
For every realized suffix, $H_{i,t}$ is the minimum loss among all later candidate stopping locations on that suffix. The location selected by any causal $\tau$ is one member of the same candidate set, so $H_{i,t}\le L_{i,t}(\tau)$ pointwise. Conditional expectation preserves the inequality, and taking the infimum over causal $\tau$ gives Equation~\ref{eq:app_hindsight_bound}. \hfill$\square$

This property explains the role of the complete trajectory in supervision. A recorded suffix reveals the best answer point reachable on that realization, while Equation~\ref{eq:app_bellman} describes the conditional expectation over future trajectories. The resulting $g_{i,t}^{(\lambda)}$ is a cost-complete reachable-suffix surrogate for the policy-conditioned decision gap. Because its minimizing checkpoint is chosen after the suffix is observed, its conditional continuation loss can be lower than that of a causal stopping rule, which is the standard hindsight advantage captured by Equation~\ref{eq:app_hindsight_bound}.

\subsection{Cost cancellation at a checkpoint}
\label{app:theory_cost}
The stopping decision is made after the current option assessment has been computed. Let $C^{\mathrm{sunk}}_{i,t}$ collect all native computation and assessments already paid before the STOP/CONTINUE comparison. If the agent stops at $t$, the total episode cost can be written as
\begin{equation}
C^{\mathrm{stop}}_{i,t}
=
C^{\mathrm{sunk}}_{i,t}+C_f(i,t).
\label{eq:app_cost_stop}
\end{equation}
If it continues to a later checkpoint $u$, the corresponding cost is
\begin{equation}
C^{\mathrm{cont}}_{i,t\rightarrow u}
=
C^{\mathrm{sunk}}_{i,t}
+C(i,t\!\rightarrow\!u)
+C_f(i,u),
\label{eq:app_cost_continue}
\end{equation}
where $C(i,t\!\rightarrow\!u)$ contains future native computation and future option assessments. Subtracting the two losses removes $C^{\mathrm{sunk}}_{i,t}$ exactly. The current assessment therefore cancels from the decision gap, while every future assessment, native step, and answer-generation cost remains part of the continuation comparison in Equations~\ref{eq:stop_loss}--\ref{eq:future_loss}.

\subsection{Invariance to option relabeling}
\label{app:theory_permutation}
The probability-derived part of the 18-dimensional state is invariant to the literal option labels. Let $\sigma$ be any permutation of the $K_i$ answer options and let $p_t^{\sigma}(k)=p_t(\sigma^{-1}(k))$ be the consistently relabeled option distribution. The option count, sorted top-three probabilities, normalized entropy, and top-two margin are unchanged by $\sigma$. The aligned total variation also satisfies
\begin{equation}
\frac12\sum_{k=1}^{K_i}
\left|p_t^{\sigma}(k)-p_{t-1}^{\sigma}(k)\right|
=
\frac12\sum_{k=1}^{K_i}
\left|p_t(k)-p_{t-1}(k)\right|,
\label{eq:app_tv_invariance}
\end{equation}
and the indicator of whether the leading-option set changes between checkpoints is preserved under the same relabeling. The remaining progress, cost, video, subtitle, and tool-state features are label-independent. Hence the full feature vector $x_{i,t}$ is unchanged by a consistent permutation of the answer labels and supports three-, four-, and five-option questions with the same representation.

\section{\method{} Algorithms and Causal Implementation}
\label{app:algorithms}
We now describe the complete data flow from native trajectories to a deployed stopping policy. The implementation follows the quantities defined in Section~\ref{sec:method} and Appendix~\ref{app:theory}: complete native trajectories provide training supervision, the learned score is calibrated on a disjoint development role, and deployment evaluates checkpoints in causal order using only information available at that point.

\subsection{Target construction from a native trajectory}
\label{app:algorithm_target}
A native trajectory contains a sequence of nonterminal decision checkpoints followed by its native terminal state. For every nonterminal checkpoint, the training cache contains the constrained prefix-finalizer answer and its cost, while the native trajectory records the terminal answer and the cumulative native cost. Let $m_i$ be the number of nonterminal decision checkpoints for trajectory $i$. The cost scale $S$ is the mean full native model-token cost over training trajectories with $m_i>0$.

\begin{algorithm}[H]
\caption{Decision-gap target construction for one cost weight $\lambda$}
\label{alg:target_construction}
\KwIn{Complete native trajectories; gold answers; prefix-finalizer answers and costs; cost scale $S$; cost weight $\lambda$}
\KwOut{Decision-gap targets $\{g_{i,t}^{(\lambda)}\}$}
\ForEach{training trajectory $i$}{
  $m_i\leftarrow$ number of nonterminal decision checkpoints\;
  \For{$t\leftarrow 0$ \KwTo $m_i-1$}{
    $L_{\mathrm{stop}}\leftarrow L_{\mathrm{stop}}(i,t)$ using Eq.~\ref{eq:stop_loss}\;
    $L_{\mathrm{best}}\leftarrow L_{\mathrm{native}}(i,t)$ using Eq.~\ref{eq:native_loss}\;
    \For{$u\leftarrow t+1$ \KwTo $m_i-1$}{
      $L_{\mathrm{best}}\leftarrow\min\{L_{\mathrm{best}},L_{\mathrm{future}}(i,t,u)\}$ using Eq.~\ref{eq:future_loss}\;
    }
    $g_{i,t}^{(\lambda)}\leftarrow L_{\mathrm{stop}}-L_{\mathrm{best}}$\;
  }
}
\end{algorithm}
The option-probe cost at checkpoint $t$ is shared by both branches and cancels inside $g_{i,t}^{(\lambda)}$; Appendix~\ref{app:theory_cost} gives the corresponding algebra explicitly. Repeating the procedure for each $\lambda\in\{0.05,0.1,0.2,0.4,0.8\}$ produces one target set for each cost weight.

\subsection{Causal state and policy fitting}
\label{app:algorithm_training}
The external option probe is evaluated at temperature zero and returns a log-probability for every valid answer option. On the 50 training videos, we fit one scalar temperature for each checkpoint-depth group $0$, $1$, $2$, and $3+$ by minimizing mean negative log-likelihood. The calibrated option distribution is then combined with execution information into the 18-dimensional state used by every agent--model stopping policy; Table~\ref{tab:appendix_features} lists the complete state.

\begin{table}[!t]
\centering
\tablestandard
\caption{\textbf{Causal stopping state.} All quantities are available at or before the current checkpoint. Probability ranks are computed after temperature scaling; total variation is aligned by option identity.}
\label{tab:appendix_features}
\par\noindent\hspace*{-\tabcolsep}\begin{tabularx}{\dimexpr\textwidth+2\tabcolsep\relax}{YlX}
\tableheadtoprule
\tableheadrow
\tableheadtoppad\tableheadbottompad\textbf{Block} & \textbf{Dimensions} & \textbf{Definition} \\
\tableheadmidrule
Answer support & 1 & Number of answer options $K_i$ \\
& 2--4 & Three largest calibrated option probabilities, sorted descending \\
& 5 & Entropy of the calibrated distribution, normalized by $\log K_i$ \\
& 6 & Difference between the largest and second-largest probabilities \\
& 7 & Total variation from the previous calibrated distribution, aligned by option identity \\
& 8 & Indicator that the set of probability-maximizing options changed from the previous checkpoint \\
\midrule
Progress and cost & 9 & Checkpoint index \\
& 10 & Log of one plus model tokens consumed before the current planner call \\
& 11 & Log of one plus characters in accumulated visible evidence \\
& 12 & Log of one plus video duration in seconds \\
& 13 & Subtitle-availability indicator \\
\midrule
Latest tool & 14--18 & One-hot state: overview, skim, focus, answer, or none \\
\bottomrule
\end{tabularx}\hspace*{-\tabcolsep}\par
\end{table}

\begin{algorithm}[H]
\caption{Training the family of Gap-Ridge stopping scores}
\label{alg:gap_ridge_training}
\KwIn{Training trajectories and prefix finalizers; raw option-probe log-probabilities; $\Lambda=\{0.05,0.1,0.2,0.4,0.8\}$; $\alpha=10$}
\KwOut{Probe temperatures and fitted scores $\{f_{\lambda}\}_{\lambda\in\Lambda}$}
Fit one temperature for checkpoint-depth groups $0,1,2,3+$ by training-set negative log-likelihood\;
Apply the fitted temperatures and construct every causal feature vector $x_{i,t}$\;
Compute the model-token scale $S$ from learnable training trajectories\;
\ForEach{$\lambda\in\Lambda$}{
  Construct $g_{i,t}^{(\lambda)}$ for all training checkpoints using Algorithm~\ref{alg:target_construction}\;
  Assign checkpoint weight $1/m_i$ within trajectory $i$\;
  Compute weighted feature means and scales and standardize $x_{i,t}$\;
  Fit weighted Ridge $f_{\lambda}$ using Eq.~\ref{eq:ridge}\;
}
\end{algorithm}
The intercept is represented by the weighted target mean, while the coefficient penalty applies to the standardized feature weights. Probe calibration, feature construction, and Ridge fitting use only the training role.

The feature definition is shared across VideoSeek and AVP. The first eight dimensions summarize answer support with label-invariant statistics; the next five encode trajectory progress and accumulated computation; the final five expose the latest VideoSeek tool state when that taxonomy is defined. AVP uses the shared \emph{none} state for this last block, as described in Section~\ref{app:agent_adapters}.

\subsection{Operating-point calibration}
\label{app:algorithm_calibration}
Calibration converts the five fitted score functions into one frozen operating point. For each $\lambda$, we evaluate 81 boundaries
\[
 b\in\{-0.200,-0.195,\ldots,0.195,0.200\},
\]
with Medium+Long activation, giving $5\times81=405$ candidates per agent--model pair. Each candidate is replayed on the 100 Video-MME calibration videos using the same prefix finalizer and full cost accounting as deployment.

\begin{algorithm}[H]
\caption{Unified operating-point calibration}
\label{alg:operating_point}
\KwIn{Calibration trajectories; fitted scores $\{f_{\lambda}\}$; boundary grid $\mathcal B=\{-0.200,-0.195,\ldots,0.200\}$}
\KwOut{Frozen cost weight $\lambda^{*}$, boundary $b^{*}$, and score $f_{\lambda^{*}}$}
$\mathcal C\leftarrow\emptyset$\;
\ForEach{$\lambda\in\Lambda$}{
  \ForEach{$b\in\mathcal B$}{
    Replay first-threshold stopping with $(f_{\lambda},b)$ on calibration videos\;
    Record accuracy change $\Delta_{\lambda,b}$, model-token saving $s_{\lambda,b}$, and video-cluster lower-bound diagnostic $\ell_{\lambda,b}$\;
    Add $(\lambda,b,\Delta_{\lambda,b},s_{\lambda,b},\ell_{\lambda,b})$ to $\mathcal C$\;
  }
}
$\mathcal C_{\mathrm{strict}}\leftarrow\{c\in\mathcal C:s_c\ge0.55,\ \ell_c\ge-0.01\}$\;
\eIf{$\mathcal C_{\mathrm{strict}}\neq\emptyset$}{
  $\Delta^{*}\leftarrow\max_{c\in\mathcal C_{\mathrm{strict}}}\Delta_c$\;
  $\mathcal N\leftarrow\{c\in\mathcal C_{\mathrm{strict}}:\Delta_c\ge\Delta^{*}-0.02\}$\;
  $(\lambda^{*},b^{*})\leftarrow\arg\max_{c\in\mathcal N}(s_c,\Delta_c,\ell_c,\mathrm{index}_c)$ lexicographically\;
}{
  $(\lambda^{*},b^{*})\leftarrow\arg\min_{c\in\mathcal C}(|s_c-0.40|,-\ell_c,-\Delta_c,\mathrm{index}_c)$ lexicographically\;
}
\end{algorithm}
The selected $(\lambda,b)$ and the corresponding Ridge model are then frozen. The same frozen policy is applied to the 750 Video-MME evaluation videos and transferred zero-shot to LongVideoBench and MLVU. Section~\ref{app:baselines} reports calibration sensitivity to the 55\% floor, 2-pp near-optimality band, and 40\% fallback target.

\subsection{Online STOP/CONTINUE wrapper}
\label{app:algorithm_online}
A deployment checkpoint is attached to a native visual proposal immediately before the associated visual observation is executed. The planning computation that produced the proposal is already part of the sunk native cost, while the stopping state contains only evidence available before that future visual observation. The wrapper processes eligible checkpoints in temporal order.

\begin{algorithm}[H]
\caption{Causal \method{} execution}
\label{alg:online_wrapper}
\KwIn{Native long-video agent; frozen $f_{\lambda^{*}}$ and boundary $b^{*}$; training-fitted probe temperatures}
\KwOut{Final multiple-choice answer}
\If{video is Short}{run native execution and \Return native terminal answer}
\While{the native agent exposes a nonterminal decision checkpoint $t$}{
  Run the constrained option probe on currently visible evidence\;
  Apply the fitted depth-group temperature and construct $x_{i,t}$\;
  $\hat g_{i,t}\leftarrow f_{\lambda^{*}}(x_{i,t})$\;
  \If{$\hat g_{i,t}\le b^{*}$}{
    Invoke the constrained prefix finalizer on the exact current prefix\;
    \If{a valid option is returned}{\Return prefix-finalizer answer}
  }
  Execute the native visual action and resume the native agent\;
}
\Return native terminal answer\;
\end{algorithm}
Thus the deployed policy terminates at the first available threshold crossing. When it stops at checkpoint $t$, its episode cost is the native cost already sunk at $t$, plus every option probe visited through $t$, plus the stopping finalizer. When it reaches the native terminal, the cost is the full native episode plus all option probes visited by the stopping wrapper. These are the same costs used during calibration and offline replay.

\subsection{Agent-specific checkpoint adapters}
\label{app:agent_adapters}
The two host agents expose different internal trajectories, but both are converted to the same checkpoint interface: accumulated evidence, a native cost view before the next visual observation, a sequence position, and a terminal state.

\paragraph{VideoSeek.}
VideoSeek records causal planner turns and their completed tool observations. A nonterminal checkpoint corresponds to a planner turn that proposes a visual action. The visible state contains the question and options, subtitles available to the native agent, and the observations returned by all completed earlier tool calls. The feature cost uses the native computation consumed before the current planner call, while the decision ledger separately records the planner cost that has already been incurred before the proposed visual action is executed. The latest-tool block records the most recent VideoSeek visual tool among overview, skim, and focus (with the answer/none states retained in the common encoding). For a stopped prefix, the finalizer reconstructs the native conversation up to the accumulated evidence at that checkpoint and requests a constrained multiple-choice answer.

\paragraph{AVP.}
AVP alternates planning with query-conditioned visual observations. A nonterminal checkpoint is created after a valid AVP planning proposal and immediately before its associated visual inference. The visible state contains the question and options, video duration, available subtitles, and all completed AVP observation rounds, including their returned evidence summaries. Backend recovery calls belonging to one logical observation remain grouped under the same planner proposal and form one stopping state. The common tool block reserves overview/skim/focus for VideoSeek, while AVP uses the shared \emph{none} state. The terminal checkpoint records the complete accrued native cost and terminal answer status.

These adapters leave the learned interface unchanged: both produce the same 18-dimensional state, use the same target construction and Ridge learner, and apply the same first-threshold-crossing rule.

\subsection{Option probe, prefix finalizer, and parsing}
\label{app:prompts_parsers}
Both auxiliary calls use the same VLM backend as the corresponding host execution and are generated at temperature zero. They operate only on the question's original three to five answer choices.

\paragraph{Option probe.}
For VideoSeek, the probe receives the question, the subtitles visible to the native policy, and the accumulated completed tool evidence. For AVP, it receives the rendered visible state described above. Its system instruction is:
\begin{promptbox}{Option-probe system prompt}
Answer the multiple-choice question using only the supplied visible evidence. Return exactly one allowed uppercase option letter.
\end{promptbox}
The call is restricted to the valid option letters, generates one token, and requests token log-probabilities. A probe record is accepted only when a log-probability is available for every valid option; these values are subsequently temperature-scaled into the calibrated distribution used by the stopping state.

\paragraph{Prefix finalizer.}
The multiple-choice query lists the original options as \texttt{(A)}, \texttt{(B)}, and so on. Its final instruction is instantiated with the valid letter set for that question:
\begin{promptbox}{Prefix-finalizer query suffix}
Please directly answer with the best option letter from the valid choices: [valid letters]. Output exactly one uppercase option letter and no other text.
\end{promptbox}
VideoSeek keeps its native system prompt and reconstructed prefix conversation, then appends a final user request to provide the answer in this required format. AVP uses the same visible state supplied to its adapter together with the system instruction:
\begin{promptbox}{AVP prefix-finalizer system prompt}
You are a prefix answerer. Use only the supplied visible state. Return exactly one allowed uppercase option letter and no other text. Do not infer unseen video content.
\end{promptbox}
The finalizer is restricted by structured decoding to the valid option set. The accepted prefix-parser forms are a bare allowed letter (including the accepted \texttt{Reply-}\emph{n} suffix form) and the exact boxed wrapper used by the backend. The prediction and the complete finalizer token cost are stored together and enter the target and evaluation ledger at the checkpoint where the finalizer is used.

\section{Complete Main Results and Cost Accounting}
\label{app:complete_results}
We expand the main results with paired offline uncertainty intervals, online execution values, component-wise token accounting, prefix-finalization/stopping attribution, post-hoc native-cost slices, stopping locations, and the one-time cost of collecting the supervision used to fit the stopping policy.

\subsection{Complete offline results}
\label{app:complete_results_full}
Table~\ref{tab:appendix_complete_offline} reports all eighteen agent--model--benchmark cells together with paired video-cluster bootstrap intervals and the fraction of questions stopped before the native terminal state.

\input{generated/appendix_complete_offline_table.tex}

Mean Model Token saving is 53.63\% on Video-MME, 58.78\% on LongVideoBench, and 51.28\% on MLVU. The corresponding mean accuracy changes are $+0.67$, $+3.38$, and $+4.58$ points. The complete intervals show the variation across host agents and models around these aggregate results.

\subsection{Online execution results}
\label{app:online_execution}
Figure~\ref{fig:online_execution} summarizes the online accuracy--efficiency pattern visually. Table~\ref{tab:online_execution} gives the exact Native $\rightarrow$ \method{} accuracy, model-token saving, runtime saving, and speedup for every agent--model pair on the two online benchmarks. Runtime saving is $1-T_{\mathrm{CASE}}/T_{\mathrm{Native}}$ and speedup is $T_{\mathrm{Native}}/T_{\mathrm{CASE}}$, where $T$ is measured per-question agent-execution time within the host online loop; one-time service/process initialization is excluded, and CASE-specific probe/finalizer calls are included.

\input{generated/online_execution_table.tex}

\subsection{Complete cost accounting}
\label{app:complete_results_cost}
Our primary Model Token measure is the sum of text and visual tokens. For a stopped trajectory, the policy total includes the native prefix already executed, every option probe visited through the stopping checkpoint, and the finalizer actually used to answer at that prefix. Native-terminal episodes contain the full native trajectory together with the visited probes. Table~\ref{tab:appendix_cost_decomposition} gives the corresponding absolute totals and component-wise savings.

\input{generated/appendix_cost_decomposition_table.tex}

Across the three benchmarks, reductions in visual processing accompany reductions in the growing text context and decoded model output. The absolute Model Token totals in Table~\ref{tab:appendix_cost_decomposition} are the quantities from which the percentage savings in the main tables are computed.

\subsection{Prefix finalization and stopping attribution}
\label{app:stopping_attribution}
Early stopping changes two parts of native execution at once. It removes the remaining trajectory, reducing future computation but also discarding any evidence that would have been acquired later. A STOP also invokes the constrained prefix finalizer on the retained evidence; all compared stopping methods use this same finalization path. The end-to-end accuracy change therefore combines a stopping-location effect with a prefix-finalization effect.

For each method $m$, let $H_m$ be a matched terminal-finalizer hybrid. On the questions that $m$ stops, $H_m$ keeps the complete Native trajectory but replaces the Native terminal answer with the same constrained finalizer used by $m$; on the remaining questions it keeps the Native terminal answer. The finalizer only maps the available evidence to one of the original answer options and does not acquire additional video evidence. Comparing $H_m$ with Native isolates the prefix-finalization effect, while comparing $m$ with $H_m$ isolates the effect of stopping earlier under the same finalizer:
\[
\Delta A_m
=\underbrace{A(H_m)-A(\mathrm{Native})}_{\Delta A_m^{\mathrm{fin}}}
+\underbrace{A(m)-A(H_m)}_{\Delta A_m^{\mathrm{stop}}}.
\]
Table~\ref{tab:appendix_stopping_attribution} reports this decomposition for all seven methods. For \method{}, the observed $+2.88$-point end-to-end change decomposes into a $+4.80$-point prefix-finalization effect and a $-1.92$-point stopping-location effect, alongside 54.56\% Model Token saving. The negative stopping term is consistent with the role of early stopping: removing future computation can also remove useful later evidence. The same average pattern appears across the six baselines.

\input{generated/appendix_stopping_attribution_table.tex}

For a method-independent reference, we additionally apply the same constrained finalizer to every complete Native trajectory. At each frozen paper operating point, early-stopped questions retain their prefix finalizer while non-stopped questions use this common terminal finalizer, whose cost is charged explicitly. Table~\ref{tab:appendix_common_finalizer} therefore compares all methods against one shared finalization rule. \method{} remains nondominated in 16/18 configurations, with 55.24\% Model Token saving and a $-1.92$-point accuracy difference relative to Native+F.

\input{generated/appendix_common_finalizer_table.tex}

We also form a stopping-only frontier by retaining each method's Model Token saving and replacing end-to-end accuracy change with $\Delta A_m^{\mathrm{stop}}$. Under this matched-finalizer diagnostic, \method{} is nondominated in 16 of 18 agent--model--benchmark configurations (88.9\%) with mean Pareto rank 1.11.

We apply the same paired video-cluster bootstrap used for the end-to-end frontier, keeping all operating points frozen and recomputing the matched stopping effect, saving, frontier membership, and rank in every draw. Table~\ref{tab:appendix_stopping_only_bootstrap} reports the full seven-method result. Across 10,000 draws, \method{} has mean stopping-only PFC 84.5\% with a central 95\% bootstrap interval of [72.2\%, 100.0\%] and mean rank 1.17 with interval [1.00, 1.33]. Its stopping-only PFC is strictly highest in 72.2\% of draws and highest or tied-highest in 88.9\%.

\input{generated/appendix_stopping_only_bootstrap_table.tex}

\subsection{Native-cost tail}
\label{app:complete_results_tail}
The aggregate saving is concentrated in trajectories that are expensive under native execution. We rank questions independently within each agent--model--benchmark cell by native Model Token and report the post-hoc Top20\% and Top10\% slices in Table~\ref{tab:appendix_longtail}. These slices provide a descriptive view of where native computation and CASE savings are concentrated.

\input{generated/appendix_longtail_table.tex}

The Top20\% slices account for 48.08\%, 46.65\%, and 46.51\% of native Model Token on average for Video-MME, LongVideoBench, and MLVU. CASE removes 70.72\%, 74.40\%, and 65.33\% of the corresponding model-token cost. The Top10\% slices show the same concentration more strongly.

\subsection{Where stopping occurs}
\label{app:complete_results_stops}
Figure~\ref{fig:appendix_stop_distribution} shows the first valid stopping checkpoint for every question. Checkpoint 0 is the first eligible nonterminal state; the native-terminal segment also includes Short videos, for which the main protocol leaves the host agent unchanged.

\input{generated/appendix_stop_distribution_figure.tex}

The mean stop rates are 49.09\%, 56.53\%, and 60.40\% on Video-MME, LongVideoBench, and MLVU. In most Video-MME and LongVideoBench cells, more than half of the stopped questions terminate at checkpoint 0 or 1. Table~\ref{tab:appendix_native_length} places these early stops in the context of the native trajectory length.

\input{generated/appendix_native_length_table.tex}

Across the current cells, checkpoint-0/1 stops bypass several later native decision checkpoints on average. The effect is largest on the longer VideoSeek trajectories, where early stops remove a substantial native suffix.

\subsection{Supervision-acquisition cost}
\label{app:complete_results_supervision}
Learning the stopping rule requires complete native training trajectories together with option probes and reachable-prefix finalizers on the 50-video training role. Table~\ref{tab:appendix_supervision_cost} audits this one-time acquisition cost separately from deployment.

\input{generated/appendix_supervision_cost_table.tex}

Across the six agent--model cells, collecting the 150 training-question supervision requires 3.02--8.78 million Model Tokens in total. The number of cached probe/finalizer pairs ranges from 132 to 630, reflecting the different native trajectory lengths exposed by the two hosts and three VLMs. The cached supervision supports feature construction, target construction, and Ridge fitting, and the uniform cost ledger provides the call and token accounting in Table~\ref{tab:appendix_supervision_cost}.

\section{Baseline Comparison and Operating-Point Calibration}
\label{app:baselines}
This section gives the complete baseline results and the calibration analysis behind the operating-point rule used throughout the paper.

\subsection{Stopping baselines}
\label{app:baseline_definitions}
The six controls span position-only, cost-only, confidence-based, stability-based, and learned stopping rules. \textbf{Tuned checkpoint} stops at a globally calibrated checkpoint index $k$. \textbf{Tuned budget} stops at a calibrated cumulative Model Token budget $B$. \textbf{Top-option confidence} and \textbf{option entropy} threshold the calibrated option distribution \cite{schuster2022calm}. \textbf{TASR-Video} adapts TASR's answer-stability gate to our multiple-choice setting by replacing TASR's calibrated answer-token logit margin with the external option probe's top-two probability margin; its threshold is selected by the shared calibration rule \cite{kieback2026tasr}. \textbf{Forward-Q} implements the Monte Carlo continuation target from Stop-RAG \cite{park2025stoprag}. Two weighted Ridge predictors operate on the checkpoint state, estimating current-prefix correctness and the maximum correctness among later prefixes and the Native terminal answer; their difference is thresholded. With the shared features, trajectory weights, standardization, and Ridge regularization used here, this score is the sign-reversed quality-only score in Appendix~\ref{app:ablations_cost}, so the two controls use threshold sets paired by sign. This is a lightweight video adaptation of the Monte Carlo target, whereas Stop-RAG's main method trains a text encoder with annealed forward-view $Q(\lambda)$ targets. Every tunable method uses the same Video-MME calibration rule described below, and the selected point is frozen for zero-shot evaluation on LongVideoBench and MLVU. Tuned checkpoint and tuned budget do not invoke the option probe; all other auxiliary calls are charged only when used.

\subsection{Complete baseline results}
\label{app:baseline_full_results}
Tables~\ref{tab:appendix_baselines_videoseek} and~\ref{tab:appendix_baselines_avp} expand the main baseline table with visual-token savings.

\input{generated/appendix_baselines_videoseek.tex}
\FloatBarrier
\input{generated/appendix_baselines_avp.tex}

The baseline families occupy different parts of the accuracy--cost plane. Fixed position and budget controls are competitive in several cells, while confidence and entropy thresholds vary more strongly with the host and model.

\subsection{Common-budget comparison}
\label{app:baseline_common_budget}
Table~\ref{tab:appendix_common_budget_finalizer} compares the seven stopping methods after selecting operating points toward the same pre-specified calibration saving targets and then evaluating them under the common terminal-finalizer protocol from Appendix~\ref{app:stopping_attribution}. This controls both stopping aggressiveness at selection time and the finalization rule at evaluation time; test labels do not affect operating-point selection.

\input{generated/appendix_common_budget_finalizer_table.tex}

At comparable calibration budgets, \method{}, Forward-Q, and TASR-Video have very similar stopping accuracy, while tuned budget is modestly higher at some intermediate targets. The result complements the independently selected operating points: the heterogeneous-cost correction primarily shifts \method{} toward a more aggressive saving regime rather than creating a large fixed-budget accuracy advantage.

\subsection{Pareto stability under evaluation resampling}
\label{app:pfc_bootstrap}
PFC summarizes frontier membership over finite evaluation sets, so we additionally characterize how the frontier changes under video-level resampling. Within each of the 18 agent--model--benchmark configurations, we resample videos with replacement and use the same cluster draw for every method, then recompute accuracy change, Model Token saving, Pareto membership, and Pareto rank. The calibrated operating points remain frozen throughout this analysis.

\input{generated/appendix_pfc_bootstrap_table.tex}

Across 10,000 paired bootstrap draws, \method{} has mean PFC 78.2\% with a central 95\% bootstrap interval of [66.7\%, 88.9\%] and mean Pareto rank 1.28 with interval [1.17, 1.44]. Its PFC is strictly the highest among the compared methods in 90.0\% of draws and is highest or tied-highest in 97.5\%. Thus the resampled frontier shows the same broad pattern as the point-estimate analysis while making the evaluation-sample variation explicit.

\subsection{Unified calibration rule}
\label{app:baseline_selector}
For a method's Video-MME calibration candidates, let $s_c$, $\Delta_c$, and $\ell_c$ denote Model Token saving, accuracy change, and the one-sided video-cluster accuracy lower bound. We first form
\[
\mathcal C_{\mathrm{strict}}=\{c:s_c\ge0.55,\ \ell_c\ge-0.01\}.
\]
If this set is nonempty, let $\Delta^*=\max_{c\in\mathcal C_{\mathrm{strict}}}\Delta_c$ and retain candidates satisfying $\Delta_c\ge\Delta^*-0.02$. Among them we select the point with the greatest saving, breaking ties by accuracy, the lower-bound statistic, and candidate index. If the strict set is empty, we select the candidate whose calibration saving is closest to 40\%, with ties broken by the lower-bound statistic, accuracy, and candidate index.

\subsection{Calibration sensitivity}
\label{app:baseline_sensitivity}
Table~\ref{tab:appendix_selector_floor} compares nearby strict saving floors. Raising the floor from 50\% to 55\% leaves CASE feasible in the same five of six agent--model cells, whereas a 60\% floor reduces that count to four.

\input{generated/appendix_selector_floor_table.tex}

Figure~\ref{fig:appendix_selector_sensitivity} varies the other two calibration choices. For CASE, widening the near-optimality band from 0 to 2 pp raises mean calibration Model Token saving from 55.98\% to 65.77\%; further widening to 2.5 or 3 pp changes the mean saving only to 65.97\%. For methods that enter the fallback branch, calibration accuracy is similar across the 35--45\% target range, so we use its round-number midpoint, 40\%.

\input{generated/appendix_selector_sensitivity_figure.tex}

\subsection{Branch usage and frozen operating points}
\label{app:baseline_operating_points}
Table~\ref{tab:appendix_operating_points} gives every frozen parameter together with its calibration statistics; S/F superscripts identify the strict and 40\% fallback branches. CASE uses the strict branch in five of six cells, with AVP--GLM using the fallback.

\input{generated/appendix_operating_points_table.tex}

\subsection{CASE calibration candidates}
\label{app:baseline_frontiers}
For each agent--model pair, CASE evaluates five cost weights and 81 boundaries, yielding 405 calibration candidates. Figure~\ref{fig:appendix_calibration_frontiers} shows the strict-feasible and near-optimal regions together with the selected point.

\input{generated/appendix_frontier_figure.tex}

\section{Extended Ablations}
\label{app:ablations}
We extend the main component analysis along cost modeling, target horizon, state representation, fitting procedure, training-set size, and probe design. The main-paper ablation treats every variant as a complete stopping method: each model selects its own Video-MME operating point with the shared calibration rule before zero-shot transfer. In this appendix, saving-matched diagnostics serve a different purpose: they hold stopping aggressiveness approximately fixed to isolate accuracy differences at similar computation savings. Unless stated otherwise, all alternatives use the same training/calibration roles and the complete deployment cost ledger.

\subsection{Independent end-to-end ablations}
\label{app:ablations_independent}
Table~\ref{tab:appendix_end_to_end_ablation_metrics} expands the five independently selected variants from the main paper with visual-token saving, stop rate, and strict-calibration frequency.

\input{generated/appendix_end_to_end_ablation_metrics_table.tex}

Execution-state-only excludes the answer-support block and therefore does not invoke the option probe; its continuation-cost target uses the corresponding no-probe execution path. It reaches $+2.21$ points with 55.30\% Model Token saving overall, compared with $+2.88$/54.56\% for Full.

\subsection{Quality and continuation-cost correction}
\label{app:ablations_cost}
Equation~\ref{eq:gap_factorization} separates the quality advantage from the correction induced by heterogeneous continuation cost without changing the Full predictor. Across all six agent--model calibration sets and all five cost weights, the factorized score in Equation~\ref{eq:factorized_score} agrees with direct Gap-Ridge fitting to within floating-point precision (maximum absolute difference $5.6\times10^{-15}$).

\input{generated/appendix_cost_components_table.tex}

The cost component changes the selected efficiency regime rather than increasing accuracy. Quality-only selection gives $+3.00$ points at 46.00\% Model Token saving, whereas Full gives $+2.88$ at 54.56\%: an accuracy difference of about $-0.13$ points accompanies an $8.57$-point increase in saving. Under the paired thresholds above, the linear Forward-Q control selects the same policies and obtains the same aggregate result as Quality-only. Replacing heterogeneous native continuation cost with a constant per-step cost reaches $+2.82$/49.03\%, recovering part but not all of the efficiency shift. Thus the heterogeneous cost correction aligns the quality predictor with a more aggressive computation-saving operating point while leaving answer quality nearly unchanged.

\subsection{Saving-matched core diagnostics}
\label{app:ablations_core}
Table~\ref{tab:appendix_core_ablations} controls stopping aggressiveness by evaluating each core alternative at the calibration point whose Model Token saving is closest to the corresponding Full policy. This saving-matched view isolates answer-quality differences at comparable early-stopping intensity; deployed operating points are the independently selected variants in Table~\ref{tab:appendix_end_to_end_ablation_metrics}.

\input{generated/appendix_core_ablations_table.tex}

Belief-only and Current-correctness remain lower than Full across the benchmark averages at comparable saving. Execution-state-only and the One-step horizon control remain closer, showing that these restricted variants preserve much of the local stopping signal.

\subsection{Continuation horizon}
\label{app:ablations_targets}
Table~\ref{tab:appendix_target_family} varies how far the realized suffix is exposed to the target while retaining the full state and Ridge learner. These controls are saving-matched to Full because the question here is the effect of supervision horizon at comparable stopping aggressiveness.

\input{generated/appendix_target_family_table.tex}

Native-terminal supervision is lower on both Video-MME and LongVideoBench. Finite-horizon suffix targets are also competitive: horizons 2 and 4 closely match the full-suffix result. These results indicate that the main gain comes from learning a continuation-aware stopping target; the detailed continuation-cost parameterization has a smaller effect on the benchmark averages.

\subsection{State channels, trajectory weighting, and regularization}
\label{app:ablations_fine}
Table~\ref{tab:appendix_fine_controls} removes individual groups of causal state variables and varies the Ridge fit.

\input{generated/appendix_fine_controls_table.tex}

Individual state channels are partly redundant: removing belief change, consumed cost, tool type, or the static duration/subtitle attributes changes the aggregate result only slightly. The larger Belief-only and Execution-state-only variants in Table~\ref{tab:appendix_core_ablations} provide the stronger test of the two state blocks. Ridge regularization also changes the selected operating point; $\alpha=10$ gives the paper's balance of Video-MME accuracy and saving, while nearby and larger values remain competitive.

\subsection{Training-video scaling}
\label{app:ablations_scaling}
We refit both the checkpoint-depth probe temperatures and Gap-Ridge using 10, 20, 30, or all 50 training videos. For 10--30 videos, three fixed random orders of the canonical 50-video training role define nested subsets; the 100-video calibration role is unchanged.

\input{generated/appendix_training_scaling_figure.tex}

Useful stopping policies emerge from small subsets: the 10-video fits average $+0.60$ points / 47.54\% saving on Video-MME and $+3.04$ / 54.56\% on LongVideoBench. Using all 50 training videos gives $+0.67$ / 53.63\% and $+3.38$ / 58.78\%, respectively.

\subsection{Probe cadence and probability calibration}
\label{app:ablations_probe}
Finally, we vary how often the stopping policy is evaluated and how the auxiliary option probabilities are calibrated.

\input{generated/appendix_probe_controls_table.tex}

Checking every second or third native checkpoint reduces both decision opportunities and realized saving under the current selector; LongVideoBench accuracy also falls. Evaluating every checkpoint therefore retains the strongest overall accuracy--efficiency profile. Raw logprobs and a single global temperature remain viable, while the checkpoint-depth temperatures used by \method{} give the highest mean accuracy change on both benchmarks.

\input{generated/appendix_probe_calibration_quality_table.tex}

The depth-group temperatures also improve probability quality outside the fitting videos: held-out NLL and Brier score decrease on Video-MME and under zero-shot transfer to LongVideoBench and MLVU.

\section{Stopping-Behavior and Mechanism Analysis}
\label{app:behavior}
The preceding sections establish the aggregate accuracy--efficiency trade-off. We now examine how that trade-off is produced at the level of individual answers, stopping locations, and learned state signals. The analysis uses the frozen evaluation trajectories and operating points.

\subsection{Answer transitions}
\label{app:behavior_transitions}
A net accuracy change can hide substantially different behavior across questions. Table~\ref{tab:appendix_answer_transitions} therefore reports the two directional transitions that determine the change in accuracy: Native wrong $\rightarrow$ CASE correct and Native correct $\rightarrow$ CASE wrong. Their difference, divided by the full evaluation size, is exactly the reported CASE$-$Native accuracy change.

\input{generated/appendix_answer_transitions_table.tex}

Both directions occur across the three benchmarks. For example, VideoSeek--Qwen on Video-MME corrects 197 Native errors and changes 154 Native-correct answers to wrong, yielding the net $+1.91$-point gain. The same table shows larger favorable flows in several zero-shot settings, especially AVP--InternVL on MLVU.

\subsection{Risk--saving curves}
\label{app:vtr}
\begin{figure}[!t]
\centering
\includegraphics[width=0.80\textwidth]{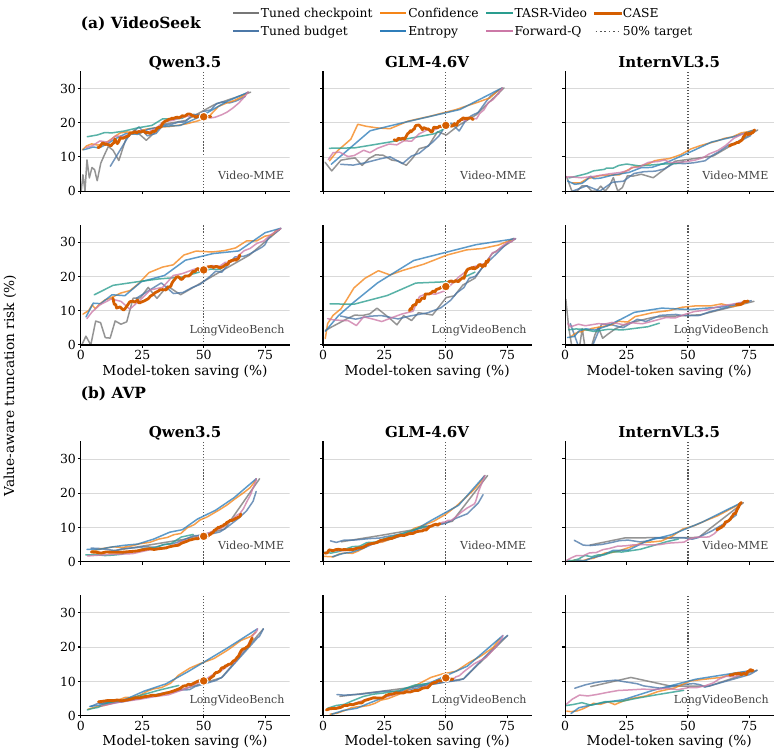}
\caption{\textbf{Value-aware truncation risk versus model-token saving.} Columns are Qwen3.5, GLM-4.6V, and InternVL3.5; rows cover Video-MME and LongVideoBench for VideoSeek and AVP. Lower VTR is better at a fixed saving level.}
\label{fig:vtr}
\end{figure}

As an auxiliary stopping-behavior diagnostic, we measure value-aware truncation risk (VTR). For questions stopped early by method $m$ at threshold $\theta$, VTR is the fraction whose recorded native continuation contains a later checkpoint with lower answer-risk--cost loss:
\begin{equation}
 \operatorname{VTR}_m(\theta)=\frac{1}{|\mathcal{S}_m(\theta)|}
 \sum_{i\in\mathcal{S}_m(\theta)}
 \mathbb{1}\!\left[g_{i,\tau_i^{m,\theta}}>0\right].
 \label{eq:vtr}
\end{equation}
Lower VTR is better at a fixed saving level. Figure~\ref{fig:vtr} reports the complete VTR--saving curves across the realized saving range of each method.

\subsection{Cost concentration and saved suffixes}
\label{app:behavior_cost}
CASE saving rises strongly with Native Model Token cost. We rank questions within each agent--model--benchmark cell by Native cost, partition them into deciles, and average the frozen CASE behavior within each decile. The bins provide a post-hoc diagnostic of cost concentration.

\input{generated/appendix_cost_decile_figure.tex}

CASE changes the cheapest trajectories little and concentrates saving on the expensive tail. Mean Model Token saving rises from 1.52\% to 74.16\% across the lowest-to-highest Video-MME cost deciles, from 1.21\% to 77.02\% on LongVideoBench, and from 5.32\% to 70.04\% on MLVU. Stop rates increase in the same direction. Together with Figure~\ref{fig:appendix_stop_distribution}, this shows that the efficiency gain comes mainly from removing long Native suffixes on expensive trajectories.

\subsection{Policy signals and stopping dynamics}
\label{app:behavior_policy}
Figure~\ref{fig:appendix_stop_distribution} shows substantial variation in the first STOP location across questions, models, and hosts. To inspect the fitted rule itself, Figure~\ref{fig:appendix_ridge_coefficients} visualizes the coefficients of the six selected Ridge policies after feature standardization.

\input{generated/appendix_ridge_coefficients_figure.tex}

The learned coefficients vary across hosts and models. Belief variables, checkpoint depth, accumulated cost/evidence, and host-specific execution channels contribute in different combinations, consistent with the feature-removal results in Section~\ref{app:ablations_fine}.

\section{Robustness}
\label{app:robustness}

\subsection{Sensitivity to the training/calibration assignment}
\label{app:dev_resampling}
We form nine balanced alternative assignments of the same 150 Video-MME development videos into 50 training and 100 calibration videos. Each assignment refits the probe temperatures and Gap-Ridge models and then applies the same calibration rule; the 750 Video-MME evaluation videos remain unchanged.

\input{generated/appendix_dev_resampling_table.tex}

Across the 54 alternative split--cell combinations, 28 use the primary branch and 26 use the 40\% fallback. Averaged across all six agent--model pairs and nine assignments, Video-MME gives $+0.09$ points with 54.27\% Model Token saving, while LongVideoBench gives $+2.11$ points with 58.50\% saving. The per-cell standard deviations in Table~\ref{tab:appendix_dev_resampling} reflect both refitting and the selected stopping aggressiveness.

\subsection{Trajectory-seed robustness}
\label{app:multiseed}
We isolate sensitivity to stochastic host trajectories by freezing the canonical seed-42 CASE controller and applying it unchanged to native trajectories generated with seeds 43 and 3407. Probe temperatures, feature standardization, Ridge coefficients, the selected cost weight $\lambda$, boundary $b$, and the Medium/Long activation rule are all fixed from seed 42; the target seeds are used only to generate the trajectories on which the controller is replayed and scored. No target-seed fitting, calibration, or operating-point selection is performed. The seed-42 member of each three-seed summary is the canonical paper run, so all three seeds are evaluated with the same controller parameters.

\input{generated/appendix_multiseed_table.tex}

Across the 18 agent--model--benchmark cells, the frozen controller averages $+2.37$ accuracy points and 52.33\% Model Token saving. Averaging the per-cell variability across the three trajectory seeds gives a standard deviation of 1.42 points for accuracy change and 3.59 points for Model Token saving. Thus the deployed controller retains a similar efficiency profile when the native agent follows different stochastic trajectories.

\section{Experimental Protocol and Reproducibility}
\label{app:protocol}
This section records the experimental roles and runtime configuration used throughout the paper. The protocol is fixed at the video level: all questions associated with one Video-MME video remain in the same role, and the learned stopping policy is selected on Video-MME before zero-shot evaluation on LongVideoBench and MLVU.

\subsection{Video-MME partitions and data roles}
\label{app:protocol_videomme}
Video-MME contains 900 videos with three questions per video. We use 50 videos (150 questions) to fit probe calibration and the stopping score, 100 disjoint videos (300 questions) to select the operating point, and the remaining 750 videos (2250 questions) for the reported Video-MME evaluation. Table~\ref{tab:appendix_data_roles} summarizes both the partition sizes and the stage at which each role is used.

\begin{table}[H]
\centering
\tablestandard
\caption{\textbf{Data roles in the experimental pipeline.} A check mark indicates that labels from that partition enter the corresponding stage. LongVideoBench and MLVU provide frozen-policy evaluation.}
\label{tab:appendix_data_roles}
\par\noindent\hspace*{-\tabcolsep}\begin{tabularx}{\dimexpr\textwidth+2\tabcolsep\relax}{lYYYYYY}
\tableheadtoprule
\tableheadrow
\tableheadtoppad\tableheadbottompad\textbf{Partition} & \textbf{Videos} & \textbf{Questions} & \textbf{Probe temp.} & \textbf{Ridge fit} & \textbf{$(\lambda,b)$ selection} & \textbf{Reported score} \\
\tableheadmidrule
Video-MME train & 50 & 150 & $\checkmark$ & $\checkmark$ & -- & -- \\
Video-MME calibration & 100 & 300 & -- & -- & $\checkmark$ & -- \\
Video-MME evaluation & 750 & 2250 & -- & -- & -- & $\checkmark$ \\
LongVideoBench validation & 753 & 1337 & -- & -- & -- & $\checkmark$ \\
MLVU multiple-choice & 1242 & 2174 & -- & -- & -- & $\checkmark$ \\
\bottomrule
\end{tabularx}\hspace*{-\tabcolsep}\par
\end{table}

The cost scale $S$, four checkpoint-depth probe temperatures, and the five Ridge models associated with the cost-weight grid are fit on the 50-video training role. The 100-video calibration role then evaluates the 405 candidate $(\lambda,b)$ pairs and applies Algorithm~\ref{alg:operating_point}. The resulting coefficients, probe temperatures, $\lambda$, and boundary remain frozen for Video-MME evaluation, LongVideoBench, and MLVU.

\subsection{Zero-shot benchmark transfer}
\label{app:protocol_lvb}
LongVideoBench and MLVU are evaluated after the complete Video-MME stopping policy is frozen. For each agent--model pair, the transferred object includes the probe calibration, 18-dimensional feature definition, Ridge coefficients, selected $\lambda$ and $b$, Medium+Long activation rule, prefix-finalizer interface, parser, and cost accounting. LongVideoBench contains 1337 scored questions over 753 videos. The MLVU multiple-choice set contains 2175 raw questions over 1242 videos; one objectively blank item is excluded for both Native and stopping methods, leaving 2174 scored questions. Target-benchmark labels enter at the final scoring stage.

\subsection{Models and host-agent configurations}
\label{app:protocol_models}
Table~\ref{tab:appendix_models} gives the exact local checkpoint identities used for the three VLM backends. Revision hashes come directly from the Hugging Face download metadata stored with each checkpoint.

\begin{table}[H]
\centering
\tablestandard
\caption{\textbf{Vision-language model checkpoints.} Service inference uses bfloat16 compute for all three backends; Qwen retains its FP8 checkpoint weights.}
\label{tab:appendix_models}
\par\noindent\hspace*{-\tabcolsep}\begin{tabularx}{\dimexpr\textwidth+2\tabcolsep\relax}{p{0.16\textwidth}p{0.25\textwidth}Xp{0.14\textwidth}}
\tableheadtoprule
\tableheadrow
\tableheadtoppad\tableheadbottompad\textbf{Paper name} & \textbf{Upstream checkpoint} & \textbf{Revision} & \textbf{Weights} \\
\tableheadmidrule
Qwen3.5-35B-A3B-FP8 & \texttt{Qwen/Qwen3.5-}\newline\texttt{35B-A3B-FP8} & \texttt{9d1823d2dee688a6b25e}\newline\texttt{77009dc727688c44936e} & FP8, $128\!\times\!128$ blocks \\
GLM-4.6V-Flash & \texttt{zai-org/}\newline\texttt{GLM-4.6V-Flash} & \texttt{411bb4d77144a3f03acc}\newline\texttt{bf4b780f5acb8b7cde4e} & bfloat16 \\
InternVL3.5-30B-A3B-HF & \texttt{OpenGVLab/InternVL3\_5-}\newline\texttt{30B-A3B-HF} & \texttt{647fbc96f5fc9798f9f5}\newline\texttt{811ddf622f07a09ccb9f} & bfloat16 \\
\bottomrule
\end{tabularx}\hspace*{-\tabcolsep}\par
\end{table}

For each host, the host-agent configuration is fixed across the three VLM backends and shared by Native, \method{}, and all stopping baselines. \method{} intervenes only in termination and otherwise leaves native evidence acquisition unchanged. Canonical reported results use trajectory seed 42. Appendix~\ref{app:multiseed} keeps the complete seed-42 \method{} controller fixed while changing only the native trajectory seed to 43 or 3407. In both hosts, \method{} is active for Medium and Long videos while Short videos retain native execution.

\subsection{Runtime and software environment}
\label{app:protocol_runtime}
The three model services use separate model-specific environments but the same core inference stack: vLLM 0.23.0, Transformers 5.12.1, and PyTorch 2.11.0+cu130. Qwen3.5 and GLM-4.6V-Flash run with tensor parallelism 1; InternVL3.5 uses tensor parallelism 2. The service context is 65,536 tokens for Qwen/GLM and 32,768 tokens for InternVL. For AVP--InternVL3.5, a context-length rejection triggers one retry after resampling the requested visual frames to 70\%; the context window is unchanged.

The experiments run on an NVIDIA server containing four RTX 5880 Ada Generation GPUs (49,140 MiB each) and two RTX PRO 6000 Blackwell Server Edition GPUs (97,887 MiB each). The installed driver is 580.65.06; \texttt{nvidia-smi} reports CUDA Version 13.0. The final full-service launchers use single-GPU services for Qwen/GLM and two GPUs for InternVL, consistent with the tensor-parallel settings above.

\subsection{Statistical protocol}
\label{app:protocol_statistics}
Accuracy uncertainty estimates respect the video grouping. For a fixed stopping policy, we form paired Native--\method{} per-question correctness differences, group all questions from the same video, and perform 10,000 video-clustered bootstrap draws with seed 42. Each draw resamples videos with replacement and retains every question belonging to each sampled video; the reported 95\% interval uses the 2.5th and 97.5th percentiles. Calibration also records a one-sided 95\% lower bound from the mean and standard error of the per-video paired accuracy differences. The strict branch of Algorithm~\ref{alg:operating_point} uses an LCB threshold of $-1$ pp, and an empty strict set activates the 40\% fallback.

The development-role experiment in Appendix~\ref{app:robustness} measures a different source of variation: it rebuilds the fitted policy after reassigning the same 150 development videos between the 50-video training and 100-video calibration roles. Thus the clustered bootstrap summarizes evaluation-sample uncertainty for a frozen policy, whereas the balanced resplits summarize how the fitted and selected stopping policy changes with the development partition.

%% file: generated/appendix_complete_offline_table.tex
\begin{table}[H]
\centering
\tablestandard
\caption{\textbf{Complete offline CASE results.} End-to-end accuracy is Native $\rightarrow$ CASE. The 95\% interval is a paired video-cluster bootstrap interval for the accuracy change (10,000 draws, seed 42).}
\label{tab:appendix_complete_offline}
\textbf{(a) Video-MME}\\[-2pt]
\par\noindent\hspace*{-\tabcolsep}\begin{tabularx}{\dimexpr\textwidth+2\tabcolsep\relax}{llYYYYYY}
\tableheadtoprule
\tableheadrow
\tableheadbottompad \tableheadtoppad \textbf{Agent} & \textbf{Model} & \textbf{Model save (\%)} & \textbf{Visual save (\%)} & \textbf{Accuracy (\%)} & \textbf{$\Delta$ (pp)} & \textbf{95\% CI (pp)} & \textbf{Stop (\%)} \\
\tableheadmidrule
VideoSeek & Qwen3.5 & 52.43 & 57.64 & 51.82 $\rightarrow$ 53.73 & +1.91 & [+0.22, +3.56] & 42.18 \\
VideoSeek & GLM-4.6V & 49.77 & 58.07 & 56.40 $\rightarrow$ 55.69 & $-$0.71 & [-1.78, +0.31] & 23.42 \\
VideoSeek & InternVL3.5 & 69.85 & 77.06 & 52.84 $\rightarrow$ 52.67 & $-$0.18 & [-1.78, +1.42] & 57.47 \\
AVP & Qwen3.5 & 56.67 & 69.67 & 65.02 $\rightarrow$ 65.29 & +0.27 & [-0.98, +1.56] & 56.80 \\
AVP & GLM-4.6V & 30.72 & 32.36 & 57.91 $\rightarrow$ 57.51 & $-$0.40 & [-1.51, +0.67] & 55.69 \\
AVP & InternVL3.5 & 62.35 & 65.26 & 51.11 $\rightarrow$ 54.27 & +3.16 & [+1.64, +4.71] & 58.98 \\
\bottomrule
\end{tabularx}
\vspace{3pt}
\textbf{(b) LongVideoBench}\\[-2pt]
\par\noindent\hspace*{-\tabcolsep}\begin{tabularx}{\dimexpr\textwidth+2\tabcolsep\relax}{llYYYYYY}
\tableheadtoprule
\tableheadrow
\tableheadbottompad \tableheadtoppad \textbf{Agent} & \textbf{Model} & \textbf{Model save (\%)} & \textbf{Visual save (\%)} & \textbf{Accuracy (\%)} & \textbf{$\Delta$ (pp)} & \textbf{95\% CI (pp)} & \textbf{Stop (\%)} \\
\tableheadmidrule
VideoSeek & Qwen3.5 & 64.24 & 64.03 & 40.91 $\rightarrow$ 44.13 & +3.22 & [+0.83, +5.56] & 48.02 \\
VideoSeek & GLM-4.6V & 54.87 & 61.39 & 47.49 $\rightarrow$ 47.34 & $-$0.15 & [-1.59, +1.26] & 29.54 \\
VideoSeek & InternVL3.5 & 74.08 & 85.39 & 42.11 $\rightarrow$ 46.00 & +3.89 & [+1.44, +6.36] & 66.57 \\
AVP & Qwen3.5 & 55.83 & 71.40 & 59.01 $\rightarrow$ 61.71 & +2.69 & [+0.68, +4.70] & 61.11 \\
AVP & GLM-4.6V & 35.43 & 42.33 & 51.31 $\rightarrow$ 54.38 & +3.07 & [+1.43, +4.68] & 65.52 \\
AVP & InternVL3.5 & 68.23 & 75.43 & 38.22 $\rightarrow$ 45.77 & +7.55 & [+4.79, +10.33] & 68.44 \\
\bottomrule
\end{tabularx}
\vspace{3pt}
\textbf{(c) MLVU}\\[-2pt]
\par\noindent\hspace*{-\tabcolsep}\begin{tabularx}{\dimexpr\textwidth+2\tabcolsep\relax}{llYYYYYY}
\tableheadtoprule
\tableheadrow
\tableheadbottompad \tableheadtoppad \textbf{Agent} & \textbf{Model} & \textbf{Model save (\%)} & \textbf{Visual save (\%)} & \textbf{Accuracy (\%)} & \textbf{$\Delta$ (pp)} & \textbf{95\% CI (pp)} & \textbf{Stop (\%)} \\
\tableheadmidrule
VideoSeek & Qwen3.5 & 59.46 & 69.53 & 45.58 $\rightarrow$ 47.88 & +2.30 & [+0.50, +4.07] & 47.84 \\
VideoSeek & GLM-4.6V & 45.52 & 53.51 & 50.23 $\rightarrow$ 49.95 & $-$0.28 & [-1.26, +0.73] & 26.03 \\
VideoSeek & InternVL3.5 & 54.20 & 59.96 & 44.99 $\rightarrow$ 50.87 & +5.89 & [+4.27, +7.52] & 63.57 \\
AVP & Qwen3.5 & 65.70 & 81.88 & 55.43 $\rightarrow$ 59.89 & +4.46 & [+2.85, +6.05] & 75.76 \\
AVP & GLM-4.6V & 31.19 & 38.73 & 53.13 $\rightarrow$ 53.54 & +0.41 & [-0.47, +1.30] & 66.24 \\
AVP & InternVL3.5 & 51.58 & 53.26 & 34.08 $\rightarrow$ 48.76 & +14.67 & [+12.73, +16.67] & 82.98 \\
\bottomrule
\end{tabularx}
\end{table}

%% file: generated/online_execution_table.tex
\begin{table}[H]
\centering
\tablestandard
\caption{\textbf{Online execution.} Results are reported on 900-question subsets of Video-MME and LongVideoBench. End-to-end accuracy is Native $\rightarrow$ \method{}; $\Delta$ is in percentage points. Model-token/runtime savings are relative to Native, and speedup is $T_{\mathrm{Native}}/T_{\mathrm{CASE}}$. Bold indicates the higher accuracy in each pair.}
\label{tab:online_execution}
\par\noindent\hspace*{-\tabcolsep}\begin{tabularx}{\dimexpr\textwidth+2\tabcolsep\relax}{llYYYYY}
\tableheadtoprule
\tableheadrow
\multicolumn{7}{c}{\tableheadtoppad\textbf{Video-MME}} \\
\tableheadrow
\tableheadtoppad \tableheadbottompad \textbf{Agent} & \textbf{Model} & \textbf{Model save} & \textbf{Runtime save} & \textbf{Speedup} & \textbf{Accuracy} & \textbf{$\Delta$} \\
\tableheadmidrule
VideoSeek & Qwen3.5 & 52.18\% & 50.40\% & 2.02$\times$ & \accpair{60.89}{60.00} & -0.89 \\
VideoSeek & GLM-4.6V & 44.24\% & 39.04\% & 1.64$\times$ & \accpair{56.33}{55.89} & -0.44 \\
VideoSeek & InternVL3.5 & 71.20\% & 72.27\% & 3.61$\times$ & \accpair{52.11}{53.22} & +1.11 \\
AVP & Qwen3.5 & 57.02\% & 56.47\% & 2.30$\times$ & \accpair{64.22}{63.33} & -0.89 \\
AVP & GLM-4.6V & 30.27\% & 35.16\% & 1.54$\times$ & \accpair{57.67}{56.44} & -1.22 \\
AVP & InternVL3.5 & 62.27\% & 61.51\% & 2.60$\times$ & \accpair{51.11}{53.11} & +2.00 \\
\bottomrule
\end{tabularx}\hspace*{-\tabcolsep}\par
\vspace{4pt}
\par\noindent\hspace*{-\tabcolsep}\begin{tabularx}{\dimexpr\textwidth+2\tabcolsep\relax}{llYYYYY}
\tableheadtoprule
\tableheadrow
\multicolumn{7}{c}{\tableheadtoppad\textbf{LongVideoBench}} \\
\tableheadrow
\tableheadtoppad \tableheadbottompad \textbf{Agent} & \textbf{Model} & \textbf{Model save} & \textbf{Runtime save} & \textbf{Speedup} & \textbf{Accuracy} & \textbf{$\Delta$} \\
\tableheadmidrule
VideoSeek & Qwen3.5 & 53.49\% & 52.35\% & 2.10$\times$ & \accpair{53.78}{52.89} & -0.89 \\
VideoSeek & GLM-4.6V & 46.20\% & 41.68\% & 1.71$\times$ & \accpair{49.22}{52.22} & +3.00 \\
VideoSeek & InternVL3.5 & 77.47\% & 80.73\% & 5.19$\times$ & \accpair{41.78}{44.56} & +2.78 \\
AVP & Qwen3.5 & 54.93\% & 54.31\% & 2.19$\times$ & \accpair{58.22}{61.22} & +3.00 \\
AVP & GLM-4.6V & 35.75\% & 39.41\% & 1.65$\times$ & \accpair{51.22}{57.00} & +5.78 \\
AVP & InternVL3.5 & 69.02\% & 65.83\% & 2.93$\times$ & \accpair{41.78}{47.22} & +5.44 \\
\bottomrule
\end{tabularx}\hspace*{-\tabcolsep}\par
\end{table}

%% file: generated/appendix_cost_decomposition_table.tex
\begin{table}[H]
\centering
\tablestandard
\caption{\textbf{Complete token-cost accounting.} Model Token is text plus visual tokens. Absolute Model Token totals are shown in millions as Native $\rightarrow$ CASE.}
\label{tab:appendix_cost_decomposition}
\textbf{(a) Video-MME}\\[-2pt]
\par\noindent\hspace*{-\tabcolsep}\begin{tabularx}{\dimexpr\textwidth+2\tabcolsep\relax}{llYYYYY}
\tableheadtoprule
\tableheadrow
\tableheadbottompad \tableheadtoppad \textbf{Agent} & \textbf{Model} & \textbf{Model Token (M)} & \textbf{Input save} & \textbf{Output save} & \textbf{Text save} & \textbf{Visual save} \\
\tableheadmidrule
VideoSeek & Qwen3.5 & 128.70 $\rightarrow$ 61.22 & 51.79 & 46.64 & 51.04 & 57.64 \\
VideoSeek & GLM-4.6V & 144.72 $\rightarrow$ 72.69 & 45.21 & 47.08 & 45.35 & 58.07 \\
VideoSeek & InternVL3.5 & 111.35 $\rightarrow$ 33.58 & 65.19 & 67.24 & 65.34 & 77.06 \\
AVP & Qwen3.5 & 116.67 $\rightarrow$ 50.55 & 50.94 & 55.54 & 51.44 & 69.67 \\
AVP & GLM-4.6V & 124.90 $\rightarrow$ 86.53 & 29.19 & 37.15 & 30.14 & 32.36 \\
AVP & InternVL3.5 & 183.31 $\rightarrow$ 69.02 & 57.65 & 59.47 & 57.79 & 65.26 \\
\bottomrule
\end{tabularx}
\vspace{3pt}
\textbf{(b) LongVideoBench}\\[-2pt]
\par\noindent\hspace*{-\tabcolsep}\begin{tabularx}{\dimexpr\textwidth+2\tabcolsep\relax}{llYYYYY}
\tableheadtoprule
\tableheadrow
\tableheadbottompad \tableheadtoppad \textbf{Agent} & \textbf{Model} & \textbf{Model Token (M)} & \textbf{Input save} & \textbf{Output save} & \textbf{Text save} & \textbf{Visual save} \\
\tableheadmidrule
VideoSeek & Qwen3.5 & 106.34 $\rightarrow$ 38.03 & 65.93 & 53.67 & 64.30 & 64.03 \\
VideoSeek & GLM-4.6V & 131.39 $\rightarrow$ 59.30 & 51.49 & 47.01 & 51.13 & 61.39 \\
VideoSeek & InternVL3.5 & 64.32 $\rightarrow$ 16.67 & 64.64 & 72.10 & 65.14 & 85.39 \\
AVP & Qwen3.5 & 73.92 $\rightarrow$ 32.65 & 51.22 & 55.42 & 51.63 & 71.40 \\
AVP & GLM-4.6V & 83.16 $\rightarrow$ 53.70 & 33.10 & 41.27 & 33.99 & 42.33 \\
AVP & InternVL3.5 & 106.21 $\rightarrow$ 33.74 & 61.02 & 61.57 & 61.05 & 75.43 \\
\bottomrule
\end{tabularx}
\vspace{3pt}
\textbf{(c) MLVU}\\[-2pt]
\par\noindent\hspace*{-\tabcolsep}\begin{tabularx}{\dimexpr\textwidth+2\tabcolsep\relax}{llYYYYY}
\tableheadtoprule
\tableheadrow
\tableheadbottompad \tableheadtoppad \textbf{Agent} & \textbf{Model} & \textbf{Model Token (M)} & \textbf{Input save} & \textbf{Output save} & \textbf{Text save} & \textbf{Visual save} \\
\tableheadmidrule
VideoSeek & Qwen3.5 & 103.73 $\rightarrow$ 42.05 & 55.67 & 51.24 & 54.86 & 69.53 \\
VideoSeek & GLM-4.6V & 143.18 $\rightarrow$ 78.01 & 39.84 & 35.11 & 39.44 & 53.51 \\
VideoSeek & InternVL3.5 & 149.01 $\rightarrow$ 68.25 & 41.36 & 34.96 & 40.60 & 59.96 \\
AVP & Qwen3.5 & 79.13 $\rightarrow$ 27.14 & 56.01 & 63.74 & 57.18 & 81.88 \\
AVP & GLM-4.6V & 94.57 $\rightarrow$ 65.07 & 24.21 & 34.42 & 25.50 & 38.73 \\
AVP & InternVL3.5 & 106.46 $\rightarrow$ 51.55 & 46.63 & 57.66 & 47.85 & 53.26 \\
\bottomrule
\end{tabularx}
\end{table}

%% file: generated/appendix_stopping_attribution_table.tex
\begin{table}[H]
\centering
\tablestandard
\caption{\textbf{Matched attribution of prefix finalization and stopping.} For each method $m$, the matched hybrid $H_m$ runs the complete Native trajectory but applies the same constrained finalizer on exactly the questions that $m$ would stop; other questions retain the Native terminal answer. Finalizer is $A(H_m)-A(\mathrm{Native})$ and Stopping is $A(m)-A(H_m)$, so the two terms sum to the end-to-end accuracy change. M is the formal Model Token saving. Stop-PFC and rank use Stopping and M as the two Pareto objectives. Bold marks the best value and underlining the second-best.}
\label{tab:appendix_stopping_attribution}
\par\noindent\hspace*{-\tabcolsep}\begin{tabularx}{\dimexpr\textwidth+2\tabcolsep\relax}{lYYYYYY}
\tableheadtoprule
\tableheadrow
\tableheadtoppad \textbf{Method} & \textbf{M} & \textbf{End-to-end $\Delta$} & \textbf{Finalizer} & \textbf{Stopping} & \textbf{Stop-PFC} & \textbf{Rank $\downarrow$} \\
\tableheadrow
\tableheadbottompad  & \textbf{(\%)} & \textbf{(pp)} & \textbf{(pp)} & \textbf{(pp)} & \textbf{/18 (\%)} &  \\
\tableheadmidrule
Tuned checkpoint & 39.09 & \textbf{+3.41} & +4.13 & \textbf{$-$0.72} & 12/18 (66.7) & 1.39 \\
Tuned budget & 44.97 & +2.29 & +3.84 & $-$1.55 & \underline{13/18 (72.2)} & \underline{1.33} \\
Confidence & 31.04 & +0.76 & +2.30 & $-$1.54 & 3/18 (16.7) & 2.39 \\
Entropy & 30.73 & +0.59 & +2.23 & $-$1.64 & 5/18 (27.8) & 2.11 \\
TASR-Video & 42.49 & \underline{+3.29} & \underline{+4.42} & \underline{$-$1.13} & 8/18 (44.4) & 1.67 \\
Forward-Q & \underline{46.00} & +3.00 & +4.25 & $-$1.25 & \underline{13/18 (72.2)} & \underline{1.33} \\
\oursrow CASE & \textbf{54.56} & +2.88 & \textbf{+4.80} & $-$1.92 & \textbf{16/18 (88.9)} & \textbf{1.11} \\
\bottomrule
\end{tabularx}\hspace*{-\tabcolsep}\par
\end{table}

%% file: generated/appendix_common_finalizer_table.tex
\begin{table}[H]
\centering
\tablestandard
\caption{\textbf{Common terminal-finalizer control.} Native+F applies the same constrained finalizer to every complete Native trajectory. Each stopping method keeps its frozen paper operating point; an early-stopped question uses its prefix finalizer, while a non-stopped question uses the same terminal finalizer as Native+F. Model-token saving is measured relative to Native+F and therefore charges the terminal-finalizer call on non-stopped questions. PFC and rank are computed only among the seven stopping methods.}
\label{tab:appendix_common_finalizer}
\par\noindent\hspace*{-\tabcolsep}\begin{tabularx}{\dimexpr\textwidth+2\tabcolsep\relax}{lYYYYY}
\tableheadtoprule
\tableheadrow
\tableheadbottompad \tableheadtoppad \textbf{Method} & \textbf{Acc. (\%)} & \textbf{$\Delta$ vs. Native+F} & \textbf{M save (\%)} & \textbf{PFC /18 (\%)} & \textbf{Rank $\downarrow$} \\
\tableheadmidrule
Native+F & 56.20 & 0.00 & 0.00 & -- & -- \\
\midrule
Tuned checkpoint & \textbf{55.48} & \textbf{$-$0.72} & 40.11 & 12/18 (66.7) & 1.39 \\
Tuned budget & 54.65 & $-$1.55 & 45.62 & \underline{13/18 (72.2)} & \underline{1.33} \\
Confidence & 54.67 & $-$1.54 & 31.68 & 3/18 (16.7) & 2.39 \\
Entropy & 54.56 & $-$1.64 & 31.38 & 4/18 (22.2) & 2.17 \\
TASR-Video & \underline{55.07} & \underline{$-$1.13} & 43.49 & 8/18 (44.4) & 1.67 \\
Forward-Q & 54.96 & $-$1.25 & \underline{46.81} & \underline{13/18 (72.2)} & \underline{1.33} \\
\oursrow CASE & 54.28 & $-$1.92 & \textbf{55.24} & \textbf{16/18 (88.9)} & \textbf{1.11} \\
\bottomrule
\end{tabularx}\hspace*{-\tabcolsep}\par
\end{table}

%% file: generated/appendix_stopping_only_bootstrap_table.tex
\begin{table}[H]
\centering
\tablestandard
\caption{\textbf{Stopping-only frontier stability under evaluation resampling.} We use 10,000 paired video-cluster bootstrap draws (seed 42) within each agent--model--benchmark configuration. Operating points remain frozen. Each draw recomputes the method-specific matched stopping effect, Model Token saving, frontier membership, and Pareto rank. Bold marks the best value and underlining the second-best.}
\label{tab:appendix_stopping_only_bootstrap}
\par\noindent\hspace*{-\tabcolsep}\begin{tabularx}{\dimexpr\textwidth+2\tabcolsep\relax}{lYYYYY}
\tableheadtoprule
\tableheadrow
\tableheadbottompad \tableheadtoppad \textbf{Method} & \textbf{Point PFC} & \textbf{Bootstrap PFC} & \textbf{Central 95\% interval} & \textbf{Point rank} & \textbf{Bootstrap rank (95\% interval)} \\
\tableheadmidrule
Tuned checkpoint & 66.7 & 62.1 & [44.4, 77.8] & 1.39 & 1.47 [1.28, 1.72] \\
Tuned budget & \underline{72.2} & \underline{76.1} & [66.7, 83.3] & \underline{1.33} & \underline{1.41} [1.22, 1.61] \\
Confidence & 16.7 & 23.6 & [11.1, 38.9] & 2.39 & 2.26 [1.94, 2.56] \\
Entropy & 27.8 & 25.8 & [11.1, 38.9] & 2.11 & 2.21 [1.89, 2.50] \\
TASR-Video & 44.4 & 46.9 & [33.3, 61.1] & 1.67 & 1.73 [1.50, 2.00] \\
Forward-Q & \underline{72.2} & 63.4 & [44.4, 77.8] & \underline{1.33} & 1.43 [1.22, 1.67] \\
\oursrow CASE & \textbf{88.9} & \textbf{84.5} & [72.2, 100.0] & \textbf{1.11} & \textbf{1.17} [1.00, 1.33] \\
\bottomrule
\end{tabularx}\hspace*{-\tabcolsep}\par
\end{table}

%% file: generated/appendix_longtail_table.tex
\begin{table}[H]
\centering
\tablestandard
\caption{\textbf{Post-hoc native-cost slices.} Top20 and Top10 contain the questions with the largest Native Model Token cost within each agent--model--benchmark cell.}
\label{tab:appendix_longtail}
\par\noindent\hspace*{-\tabcolsep}\begin{tabularx}{\dimexpr\textwidth+2\tabcolsep\relax}{lllYYYYYY}
\tableheadtoprule
\tableheadrowspan
\tableheadtoppad \textbf{Benchmark} & \textbf{Agent} & \textbf{Model} & \multicolumn{3}{c}{\textbf{Top20}} & \multicolumn{3}{c}{\textbf{Top10}} \\
\tableheadrow
\tableheadbottompad  &  &  & \textbf{Cost share} & \textbf{M save} & \textbf{$\Delta$} & \textbf{Cost share} & \textbf{M save} & \textbf{$\Delta$} \\
\tableheadmidrule
Video-MME & VideoSeek & Qwen3.5 & 58.49 & 69.92 & +6.44 & 42.85 & 73.82 & +11.11 \\
Video-MME & VideoSeek & GLM-4.6V & 73.81 & 65.55 & $-$1.11 & 60.66 & 72.28 & $-$2.22 \\
Video-MME & VideoSeek & InternVL3.5 & 58.67 & 86.19 & +0.67 & 45.57 & 90.32 & +0.00 \\
Video-MME & AVP & Qwen3.5 & 36.65 & 73.62 & $-$0.44 & 19.76 & 76.00 & +2.22 \\
Video-MME & AVP & GLM-4.6V & 27.87 & 45.07 & $-$2.44 & 14.89 & 48.43 & $-$1.33 \\
Video-MME & AVP & InternVL3.5 & 32.98 & 83.96 & +6.89 & 17.29 & 84.10 & +4.00 \\
\midrule
LongVideoBench & VideoSeek & Qwen3.5 & 61.70 & 83.04 & +8.58 & 47.36 & 87.49 & +8.21 \\
LongVideoBench & VideoSeek & GLM-4.6V & 74.52 & 68.06 & +1.87 & 50.42 & 74.08 & +1.49 \\
LongVideoBench & VideoSeek & InternVL3.5 & 50.09 & 90.34 & +10.82 & 35.92 & 94.19 & +5.22 \\
LongVideoBench & AVP & Qwen3.5 & 33.36 & 72.84 & +4.85 & 17.48 & 73.55 & +4.48 \\
LongVideoBench & AVP & GLM-4.6V & 26.21 & 47.11 & +0.00 & 13.73 & 47.32 & +0.75 \\
LongVideoBench & AVP & InternVL3.5 & 34.01 & 85.01 & +17.91 & 17.84 & 85.40 & +24.63 \\
\midrule
MLVU & VideoSeek & Qwen3.5 & 59.95 & 75.18 & +4.37 & 42.07 & 80.83 & +8.26 \\
MLVU & VideoSeek & GLM-4.6V & 69.62 & 62.03 & +1.15 & 52.51 & 72.44 & $-$2.75 \\
MLVU & VideoSeek & InternVL3.5 & 46.83 & 69.67 & +7.82 & 31.62 & 75.17 & +7.34 \\
MLVU & AVP & Qwen3.5 & 32.32 & 77.75 & +2.99 & 17.40 & 79.26 & +4.59 \\
MLVU & AVP & GLM-4.6V & 27.42 & 37.48 & +0.00 & 14.41 & 39.74 & +0.92 \\
MLVU & AVP & InternVL3.5 & 42.95 & 69.88 & +20.46 & 24.00 & 72.52 & +21.10 \\
\bottomrule
\end{tabularx}
\end{table}

%% file: generated/appendix_stop_distribution_figure.tex
\begin{figure}[H]
\centering
\includegraphics[width=0.98\textwidth]{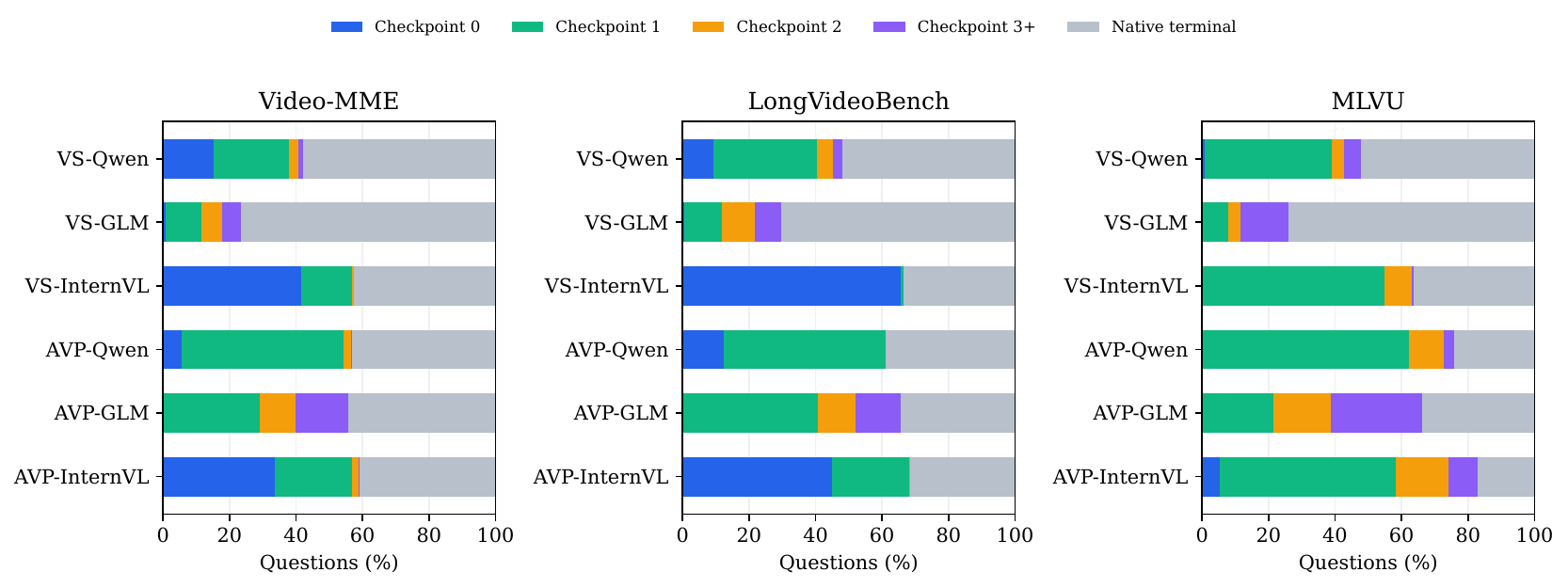}
\caption{\textbf{Where CASE terminates.} Each bar partitions questions by the checkpoint of the first valid STOP decision or the Native-terminal outcome.}
\label{fig:appendix_stop_distribution}
\end{figure}

%% file: generated/appendix_native_length_table.tex
\begin{table}[H]
\centering
\tablestandard
\caption{\textbf{Native trajectory length and early-stop suffix.} Native checkpoints are counted on Medium+Long questions where CASE is active. Early denotes stops at checkpoint 0 or 1.}
\label{tab:appendix_native_length}
\par\noindent\hspace*{-\tabcolsep}\begin{tabularx}{\dimexpr\textwidth+2\tabcolsep\relax}{lllYYY}
\tableheadtoprule
\tableheadrow
\tableheadbottompad \tableheadtoppad \textbf{Benchmark} & \textbf{Agent} & \textbf{Model} & \textbf{Native ckpts. mean / med.} & \textbf{Early among stops} & \textbf{Bypassed after early stop} \\
\tableheadmidrule
Video-MME & VideoSeek & Qwen3.5 & 2.92 / 2 & 89.9 & 3.28 \\
Video-MME & VideoSeek & GLM-4.6V & 3.78 / 2 & 49.0 & 5.32 \\
Video-MME & VideoSeek & InternVL3.5 & 3.14 / 2 & 99.0 & 3.19 \\
Video-MME & AVP & Qwen3.5 & 4.10 / 5 & 95.4 & 3.74 \\
Video-MME & AVP & GLM-4.6V & 4.79 / 5 & 52.4 & 3.89 \\
Video-MME & AVP & InternVL3.5 & 4.09 / 5 & 96.4 & 4.07 \\
\midrule
LongVideoBench & VideoSeek & Qwen3.5 & 3.92 / 3 & 84.3 & 4.21 \\
LongVideoBench & VideoSeek & GLM-4.6V & 4.95 / 2 & 40.3 & 6.61 \\
LongVideoBench & VideoSeek & InternVL3.5 & 2.69 / 2 & 100.0 & 2.93 \\
LongVideoBench & AVP & Qwen3.5 & 3.87 / 5 & 99.9 & 3.63 \\
LongVideoBench & AVP & GLM-4.6V & 4.93 / 5 & 62.1 & 3.96 \\
LongVideoBench & AVP & InternVL3.5 & 4.08 / 5 & 99.6 & 3.94 \\
\midrule
MLVU & VideoSeek & Qwen3.5 & 2.90 / 2 & 81.7 & 3.76 \\
MLVU & VideoSeek & GLM-4.6V & 4.16 / 2 & 30.0 & 4.96 \\
MLVU & VideoSeek & InternVL3.5 & 2.65 / 2 & 86.4 & 1.98 \\
MLVU & AVP & Qwen3.5 & 3.72 / 5 & 82.1 & 3.51 \\
MLVU & AVP & GLM-4.6V & 4.42 / 5 & 32.2 & 3.72 \\
MLVU & AVP & InternVL3.5 & 4.23 / 5 & 70.1 & 3.83 \\
\bottomrule
\end{tabularx}
\end{table}

%% file: generated/appendix_supervision_cost_table.tex
\begin{table}[H]
\centering
\tablestandard
\caption{\textbf{One-time supervision-acquisition cost on the 50-video training role.} Native trajectory, probe, and reachable-prefix finalizer costs are reported in millions of Model Tokens. Parentheses give the number of auxiliary model calls. Ridge fitting itself uses the cached features/targets and requires no VLM calls.}
\label{tab:appendix_supervision_cost}
\par\noindent\hspace*{-\tabcolsep}\begin{tabularx}{\dimexpr\textwidth+2\tabcolsep\relax}{llYYYY}
\tableheadtoprule
\tableheadrow
\tableheadbottompad \tableheadtoppad \textbf{Agent} & \textbf{Model} & \textbf{Native traj. (M)} & \textbf{Probes (M; calls)} & \textbf{Finalizers (M; calls)} & \textbf{Total (M)} \\
\tableheadmidrule
VideoSeek & Qwen3.5 & 5.50 & 0.57; 288 & 1.11; 288 & 7.19 \\
VideoSeek & GLM-4.6V & 5.11 & 0.45; 304 & 1.16; 304 & 6.71 \\
VideoSeek & InternVL3.5 & 2.67 & 0.09; 132 & 0.26; 132 & 3.02 \\
AVP & Qwen3.5 & 4.68 & 0.48; 406 & 0.53; 406 & 5.69 \\
AVP & GLM-4.6V & 6.82 & 0.71; 630 & 0.78; 630 & 8.32 \\
AVP & InternVL3.5 & 7.95 & 0.39; 405 & 0.44; 405 & 8.78 \\
\bottomrule
\end{tabularx}
\end{table}

%% file: generated/appendix_baselines_videoseek.tex
\begin{table}[H]
\centering
\tablestandard
\caption{\textbf{Complete stopping-baseline results for VideoSeek.} Entries report accuracy change (pp), Model Token saving (M), and visual-token saving (V). LongVideoBench and MLVU use the frozen Video-MME operating point.}
\label{tab:appendix_baselines_videoseek}
\par\noindent\hspace*{-\tabcolsep}\begin{tabularx}{\dimexpr\textwidth+2\tabcolsep\relax}{llYYYYYYYYY}
\tableheadtoprule
\tableheadrowspan
\tableheadtoppad \textbf{Model} & \textbf{Method} & \multicolumn{3}{c}{\textbf{Video-MME}} & \multicolumn{3}{c}{\textbf{LongVideoBench}} & \multicolumn{3}{c}{\textbf{MLVU}} \\
\tableheadrow
\tableheadbottompad  &  & \textbf{M} & \textbf{V} & \textbf{$\Delta$} & \textbf{M} & \textbf{V} & \textbf{$\Delta$} & \textbf{M} & \textbf{V} & \textbf{$\Delta$} \\
\tableheadmidrule
Qwen3.5 & Tuned checkpoint & \underline{42.57} & 42.17 & \underline{+2.00} & \underline{57.15} & 52.11 & \underline{+3.89} & \underline{59.14} & 62.99 & +2.39 \\
Qwen3.5 & Tuned budget & 36.73 & 34.58 & \textbf{+2.36} & 51.99 & 45.10 & \textbf{+4.19} & 42.34 & 48.70 & +2.16 \\
Qwen3.5 & Confidence & 36.13 & 42.66 & +1.69 & 37.57 & 39.57 & +2.69 & 48.31 & 57.58 & +2.21 \\
Qwen3.5 & Entropy & 33.72 & 40.13 & +1.51 & 42.52 & 44.58 & +2.09 & 43.84 & 53.71 & \underline{+2.58} \\
Qwen3.5 & TASR-Video & 37.65 & \underline{46.10} & +1.56 & 52.77 & \underline{54.10} & +3.29 & 52.79 & \underline{65.26} & \textbf{+3.36} \\
Qwen3.5 & Forward-Q & 38.64 & 43.25 & +0.93 & 50.47 & 49.59 & +3.74 & 32.91 & 43.63 & +2.44 \\
\oursrow Qwen3.5 & CASE & \textbf{52.43} & \textbf{57.64} & +1.91 & \textbf{64.24} & \textbf{64.03} & +3.22 & \textbf{59.46} & \textbf{69.53} & +2.30 \\
\addlinespace[1pt]
GLM-4.6V & Tuned checkpoint & 30.51 & 29.92 & \underline{+0.22} & 32.82 & 31.95 & +0.15 & 31.77 & 30.95 & \textbf{+0.55} \\
GLM-4.6V & Tuned budget & 32.36 & 33.06 & \textbf{+0.36} & 35.14 & 34.54 & \textbf{+0.45} & 31.48 & 36.97 & +0.41 \\
GLM-4.6V & Confidence & 34.64 & 43.10 & $-$2.09 & 32.51 & 36.21 & $-$2.09 & \textbf{56.83} & \textbf{62.83} & $-$5.66 \\
GLM-4.6V & Entropy & 32.53 & 42.61 & $-$1.82 & 36.92 & 40.14 & $-$2.99 & \underline{51.06} & 56.90 & $-$5.24 \\
GLM-4.6V & TASR-Video & 26.86 & 42.77 & $-$0.36 & 37.75 & 50.59 & $-$0.75 & 44.08 & \underline{58.41} & $-$0.18 \\
GLM-4.6V & Forward-Q & \underline{43.49} & \underline{50.08} & $-$0.13 & \underline{50.19} & \underline{56.58} & \underline{+0.22} & 32.00 & 39.67 & \underline{+0.46} \\
\oursrow GLM-4.6V & CASE & \textbf{49.77} & \textbf{58.07} & $-$0.71 & \textbf{54.87} & \textbf{61.39} & $-$0.15 & 45.52 & 53.51 & $-$0.28 \\
\addlinespace[1pt]
InternVL3.5 & Tuned checkpoint & 44.72 & 41.88 & +0.80 & 36.23 & 34.31 & +3.59 & 38.84 & 40.72 & +3.08 \\
InternVL3.5 & Tuned budget & 28.01 & 20.86 & +0.13 & 18.70 & 14.37 & +0.22 & 21.74 & 21.94 & +1.10 \\
InternVL3.5 & Confidence & 33.90 & 38.85 & +0.18 & 10.71 & 21.03 & +2.47 & 35.87 & 40.06 & +2.71 \\
InternVL3.5 & Entropy & 30.91 & 35.50 & +0.22 & 12.13 & 21.64 & +2.69 & 30.29 & 34.14 & +2.02 \\
InternVL3.5 & TASR-Video & 48.30 & 55.65 & \textbf{+1.60} & 38.36 & 50.93 & \textbf{+4.71} & \underline{52.17} & \textbf{60.69} & \underline{+5.47} \\
InternVL3.5 & Forward-Q & \underline{54.53} & \underline{59.96} & \underline{+1.51} & \underline{46.82} & \underline{54.88} & \underline{+4.04} & 35.38 & 40.29 & +3.40 \\
\oursrow InternVL3.5 & CASE & \textbf{69.85} & \textbf{77.06} & $-$0.18 & \textbf{74.08} & \textbf{85.39} & +3.89 & \textbf{54.20} & \underline{59.96} & \textbf{+5.89} \\
\bottomrule
\end{tabularx}\hspace*{-\tabcolsep}\par
\end{table}

%% file: generated/appendix_baselines_avp.tex
\begin{table}[H]
\centering
\tablestandard
\caption{\textbf{Complete stopping-baseline results for AVP.} Entries report accuracy change (pp), Model Token saving (M), and visual-token saving (V). LongVideoBench and MLVU use the frozen Video-MME operating point.}
\label{tab:appendix_baselines_avp}
\par\noindent\hspace*{-\tabcolsep}\begin{tabularx}{\dimexpr\textwidth+2\tabcolsep\relax}{llYYYYYYYYY}
\tableheadtoprule
\tableheadrowspan
\tableheadtoppad \textbf{Model} & \textbf{Method} & \multicolumn{3}{c}{\textbf{Video-MME}} & \multicolumn{3}{c}{\textbf{LongVideoBench}} & \multicolumn{3}{c}{\textbf{MLVU}} \\
\tableheadrow
\tableheadbottompad  &  & \textbf{M} & \textbf{V} & \textbf{$\Delta$} & \textbf{M} & \textbf{V} & \textbf{$\Delta$} & \textbf{M} & \textbf{V} & \textbf{$\Delta$} \\
\tableheadmidrule
Qwen3.5 & Tuned checkpoint & 25.27 & 32.53 & \textbf{+1.96} & 23.75 & 32.47 & \textbf{+4.41} & 31.23 & 40.49 & \textbf{+5.75} \\
Qwen3.5 & Tuned budget & \textbf{64.11} & \textbf{73.24} & $-$0.49 & \textbf{72.06} & \textbf{80.59} & $-$2.54 & \textbf{71.89} & \textbf{87.01} & +4.78 \\
Qwen3.5 & Confidence & 28.04 & 38.97 & $-$0.76 & 9.49 & 24.66 & +0.07 & 28.99 & 39.89 & $-$0.64 \\
Qwen3.5 & Entropy & 29.59 & 40.23 & $-$1.11 & 13.30 & 28.26 & +0.00 & 29.57 & 40.45 & $-$0.51 \\
Qwen3.5 & TASR-Video & 45.72 & 61.45 & \underline{+0.76} & 39.85 & 60.66 & \underline{+3.66} & 57.53 & 74.50 & \underline{+5.47} \\
Qwen3.5 & Forward-Q & 54.11 & 67.36 & +0.49 & 53.15 & 68.72 & +2.99 & 61.00 & 77.04 & +4.65 \\
\oursrow Qwen3.5 & CASE & \underline{56.67} & \underline{69.67} & +0.27 & \underline{55.83} & \underline{71.40} & +2.69 & \underline{65.70} & \underline{81.88} & +4.46 \\
\addlinespace[1pt]
GLM-4.6V & Tuned checkpoint & 23.01 & 26.06 & \underline{$-$0.13} & 24.13 & 27.77 & \textbf{+3.96} & 33.01 & 38.07 & \underline{+1.70} \\
GLM-4.6V & Tuned budget & 30.00 & 29.02 & \textbf{+0.09} & \textbf{39.05} & 40.06 & \underline{+3.66} & 26.57 & 32.41 & \textbf{+1.89} \\
GLM-4.6V & Confidence & 27.24 & 32.71 & $-$0.71 & 21.04 & 29.52 & +1.65 & \underline{44.83} & 52.11 & +0.28 \\
GLM-4.6V & Entropy & 27.79 & \underline{33.22} & $-$0.80 & 25.15 & 32.98 & +1.65 & \textbf{45.72} & \underline{53.13} & $-$0.23 \\
GLM-4.6V & TASR-Video & 24.41 & \textbf{34.00} & $-$0.40 & 20.43 & 34.52 & +1.80 & 43.48 & \textbf{53.89} & +1.15 \\
GLM-4.6V & Forward-Q & \textbf{30.76} & 32.60 & $-$0.44 & 35.36 & \underline{42.33} & +3.07 & 32.92 & 40.56 & +0.55 \\
\oursrow GLM-4.6V & CASE & \underline{30.72} & 32.36 & $-$0.40 & \underline{35.43} & \textbf{42.33} & +3.07 & 31.19 & 38.73 & +0.41 \\
\addlinespace[1pt]
InternVL3.5 & Tuned checkpoint & 54.30 & 56.13 & \textbf{+4.04} & 58.66 & 62.90 & \underline{+8.90} & \underline{56.56} & \underline{56.75} & +14.03 \\
InternVL3.5 & Tuned budget & \textbf{67.70} & \textbf{69.63} & +1.69 & \textbf{78.13} & \textbf{84.14} & +6.58 & \textbf{61.49} & \textbf{62.76} & +14.12 \\
InternVL3.5 & Confidence & 28.35 & 32.40 & +0.58 & 14.36 & 22.33 & +3.89 & 29.88 & 33.64 & +7.27 \\
InternVL3.5 & Entropy & 25.68 & 29.70 & +0.98 & 16.34 & 24.45 & +3.89 & 26.08 & 29.73 & +5.75 \\
InternVL3.5 & TASR-Video & 46.07 & 51.06 & \underline{+3.87} & 48.28 & 58.47 & \textbf{+9.65} & 48.30 & 50.36 & \underline{+14.54} \\
InternVL3.5 & Forward-Q & 59.85 & 62.91 & +3.73 & 64.44 & 71.82 & +7.93 & 51.90 & 53.81 & +14.49 \\
\oursrow InternVL3.5 & CASE & \underline{62.35} & \underline{65.26} & +3.16 & \underline{68.23} & \underline{75.43} & +7.55 & 51.58 & 53.26 & \textbf{+14.67} \\
\bottomrule
\end{tabularx}\hspace*{-\tabcolsep}\par
\end{table}

%% file: generated/appendix_common_budget_finalizer_table.tex
\begin{table}[H]
\centering
\tablestandard
\caption{\textbf{Common-budget comparison under a common terminal finalizer.} Each cell is test $\Delta$Acc (pp) / realized Model Token saving (\%), both relative to Native+F. For each requested saving target, method parameters are selected using Video-MME calibration only; evaluation labels never affect selection. CASE, Forward-Q, tuned budget, confidence, and entropy match all six agent--model calibration targets within the preset tolerance. Tuned checkpoint matches 5/6, 1/6, 5/6, and 2/6 cells at 30/40/50/60\%, and TASR-Video matches 6/6, 6/6, 5/6, and 4/6; unmatched entries use the nearest calibration point without extrapolation. Values macro-average the 18 agent--model--benchmark configurations.}
\label{tab:appendix_common_budget_finalizer}
\par\noindent\hspace*{-\tabcolsep}\begin{tabularx}{\dimexpr\textwidth+2\tabcolsep\relax}{lYYYY}
\tableheadtoprule
\tableheadrow
\tableheadbottompad \tableheadtoppad \textbf{Method} & \textbf{30\% target} & \textbf{40\% target} & \textbf{50\% target} & \textbf{60\% target} \\
\tableheadmidrule
Tuned checkpoint & $-0.36/27.55$ & $-0.59/31.39$ & $-0.84/42.93$ & $-1.40/49.16$ \\
Tuned budget & $-0.26/25.07$ & $-0.47/32.27$ & $-0.85/42.68$ & $-1.33/52.09$ \\
Confidence & $-1.17/23.76$ & $-1.58/31.75$ & $-2.26/40.80$ & $-3.05/50.73$ \\
Entropy & $-1.41/26.02$ & $-1.80/32.86$ & $-2.44/43.72$ & $-3.38/53.26$ \\
TASR-Video & $-0.72/26.15$ & $-0.99/34.05$ & $-1.20/42.08$ & $-1.39/49.39$ \\
Forward-Q & $-0.48/22.39$ & $-0.83/30.76$ & $-1.08/40.78$ & $-1.35/48.61$ \\
\oursrow CASE & $-0.39/21.43$ & $-0.73/29.95$ & $-1.05/40.38$ & $-1.40/47.97$ \\
\bottomrule
\end{tabularx}\hspace*{-\tabcolsep}\par
\end{table}

%% file: generated/appendix_pfc_bootstrap_table.tex
\begin{table}[H]
\centering
\tablestandard
\caption{\textbf{Pareto-frontier stability under evaluation resampling.} We use 10,000 paired video-cluster bootstrap draws (seed 42) within each agent--model--benchmark configuration. Operating points remain frozen; each draw recomputes accuracy change, Model Token saving, frontier membership, and Pareto rank. PFC follows Eq.~\ref{eq:pfc}. Bold marks the best bootstrap mean and underlining the second-best.}
\label{tab:appendix_pfc_bootstrap}
\par\noindent\hspace*{-\tabcolsep}\begin{tabularx}{\dimexpr\textwidth+2\tabcolsep\relax}{lYYYYY}
\tableheadtoprule
\tableheadrow
\tableheadbottompad \tableheadtoppad \textbf{Method} & \textbf{Point PFC} & \textbf{Bootstrap PFC} & \textbf{Central 95\% interval} & \textbf{Point rank} & \textbf{Bootstrap rank (95\% interval)} \\
\tableheadmidrule
Tuned checkpoint & 61.1 & 49.8 & [33.3, 66.7] & 1.44 & \underline{1.67} [1.44, 1.94] \\
Tuned budget & 72.2 & \underline{63.8} & [50.0, 77.8] & 1.72 & 1.85 [1.61, 2.11] \\
Confidence & 11.1 & 12.0 & [5.6, 16.7] & 2.72 & 2.93 [2.61, 3.22] \\
Entropy & 11.1 & 10.5 & [5.6, 16.7] & 2.67 & 2.92 [2.61, 3.22] \\
TASR-Video & 50.0 & 46.8 & [33.3, 61.1] & 1.67 & 1.82 [1.56, 2.06] \\
Forward-Q & 61.1 & 51.1 & [33.3, 66.7] & 1.56 & 1.72 [1.50, 2.00] \\
\oursrow CASE & 83.3 & \textbf{78.2} & [66.7, 88.9] & 1.22 & \textbf{1.28} [1.17, 1.44] \\
\bottomrule
\end{tabularx}\hspace*{-\tabcolsep}\par
\end{table}

%% file: generated/appendix_selector_floor_table.tex
\begin{table}[H]
\centering
\tablestandard
\caption{\textbf{Strict-set feasibility on Video-MME calibration.} Entries give the number of agent--model cells with at least one candidate satisfying the saving floor and the $-1$ pp accuracy-LCB constraint.}
\label{tab:appendix_selector_floor}
\begin{tabular}{lccc}
\tableheadtoprule
\tableheadrow
\tableheadbottompad \tableheadtoppad \textbf{Method} & \textbf{50\%} & \textbf{55\%} & \textbf{60\%} \\
\tableheadmidrule
Tuned checkpoint & \underline{4/6} & 2/6 & 1/6 \\
Tuned budget & 3/6 & 2/6 & 2/6 \\
Confidence & 0/6 & 0/6 & 0/6 \\
Entropy & 0/6 & 0/6 & 0/6 \\
TASR-Video & 3/6 & 3/6 & 1/6 \\
Forward-Q & \underline{4/6} & \underline{4/6} & \underline{3/6} \\
\oursrow CASE & \textbf{5/6} & \textbf{5/6} & \textbf{4/6} \\
\bottomrule
\end{tabular}
\end{table}

%% file: generated/appendix_selector_sensitivity_figure.tex
\begin{figure}[H]
\centering
\includegraphics[width=0.94\textwidth]{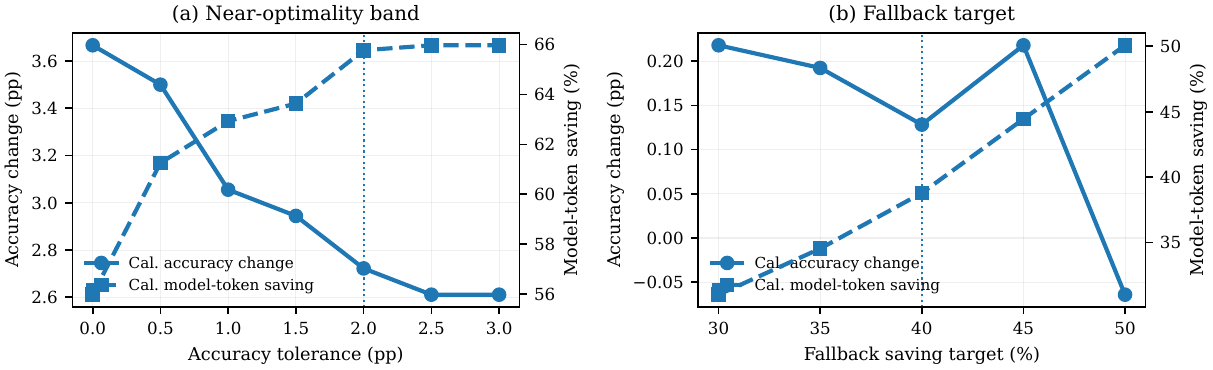}
\caption{\textbf{Calibration sensitivity of the operating-point rule.} (a) CASE mean calibration accuracy change and model-token saving as the near-optimality band is widened, with the 55\% saving floor and $-1$ pp LCB constraint fixed. (b) Mean calibration behavior over fallback method--cell pairs as the fallback target changes, with the 2-pp band fixed. Dotted lines mark the paper settings.}
\label{fig:appendix_selector_sensitivity}
\end{figure}

%% file: generated/appendix_operating_points_table.tex
\begin{table}[!t]
\centering
\tablestandard
\caption{\textbf{Frozen Video-MME calibration operating points.} Superscripts S/F mark the strict and fallback branches. Calibration columns report accuracy change, one-sided accuracy LCB, and Model Token saving.}
\label{tab:appendix_operating_points}
\textbf{(a) VideoSeek}\\[-2pt]
\par\noindent\hspace*{-\tabcolsep}\begin{tabularx}{\dimexpr\textwidth+2\tabcolsep\relax}{llYYYY}
\tableheadtoprule
\tableheadrow
\tableheadbottompad \tableheadtoppad \textbf{Model} & \textbf{Method} & \textbf{Parameter} & \textbf{Cal. M} & \textbf{Cal. $\Delta$} & \textbf{LCB} \\
\tableheadmidrule
Qwen3.5 & Tuned checkpoint & $k=2$$^{F}$ & \underline{46.15} & +0.00 & $-$3.31 \\
Qwen3.5 & Tuned budget & $B=34.2\mathrm{k}$$^{F}$ & 38.06 & +0.33 & $-$2.83 \\
Qwen3.5 & Confidence & $\tau=+0.525$$^{F}$ & 39.32 & \underline{+2.33} & $-$1.42 \\
Qwen3.5 & Entropy & $\tau=+0.8$$^{F}$ & 34.50 & +1.33 & $-$2.23 \\
Qwen3.5 & TASR-Video & $m=0.05$$^{F}$ & 40.31 & $-$0.33 & $-$4.19 \\
Qwen3.5 & Forward-Q & $\tau=-0.135$$^{F}$ & 39.86 & \underline{+2.33} & \underline{$-$1.09} \\
\oursrow Qwen3.5 & CASE & $\lambda=0.2,\ b=+0.195$$^{S}$ & \textbf{58.37} & \textbf{+3.67} & \textbf{$-$0.87} \\
\addlinespace[1pt]
GLM-4.6V & Tuned checkpoint & $k=9$$^{F}$ & 35.89 & $-$1.00 & $-$2.82 \\
GLM-4.6V & Tuned budget & $B=118.0\mathrm{k}$$^{F}$ & 41.11 & $-$0.67 & $-$2.22 \\
GLM-4.6V & Confidence & $\tau=+0.425$$^{F}$ & 39.82 & +0.00 & $-$3.12 \\
GLM-4.6V & Entropy & $\tau=+0.9$$^{F}$ & 31.01 & +0.67 & $-$2.54 \\
GLM-4.6V & TASR-Video & $m=0.1$$^{F}$ & 38.60 & $-$2.67 & $-$5.65 \\
GLM-4.6V & Forward-Q & $\tau=-0.035$$^{S}$ & \underline{58.19} & \underline{+2.00} & \textbf{$-$0.68} \\
\oursrow GLM-4.6V & CASE & $\lambda=0.2,\ b=-0.010$$^{S}$ & \textbf{63.06} & \textbf{+2.33} & \underline{$-$0.90} \\
\addlinespace[1pt]
InternVL3.5 & Tuned checkpoint & $k=2$$^{S}$ & 57.66 & +2.33 & \underline{$-$0.39} \\
InternVL3.5 & Tuned budget & $B=67.3\mathrm{k}$$^{F}$ & 41.26 & +1.33 & $-$0.40 \\
InternVL3.5 & Confidence & $\tau=+0.65$$^{F}$ & 40.84 & +1.67 & $-$1.39 \\
InternVL3.5 & Entropy & $\tau=+0.65$$^{F}$ & 40.95 & +1.00 & $-$1.96 \\
InternVL3.5 & TASR-Video & $m=0$$^{S}$ & 60.13 & \textbf{+5.00} & \textbf{+1.23} \\
InternVL3.5 & Forward-Q & $\tau=-0.14$$^{S}$ & \underline{63.89} & +3.33 & $-$0.60 \\
\oursrow InternVL3.5 & CASE & $\lambda=0.8,\ b=-0.175$$^{S}$ & \textbf{82.84} & \underline{+3.67} & $-$0.87 \\
\bottomrule
\end{tabularx}\hspace*{-\tabcolsep}\par
\vspace{4pt}
\textbf{(b) AVP}\\[-2pt]
\par\noindent\hspace*{-\tabcolsep}\begin{tabularx}{\dimexpr\textwidth+2\tabcolsep\relax}{llYYYY}
\tableheadtoprule
\tableheadrow
\tableheadbottompad \tableheadtoppad \textbf{Model} & \textbf{Method} & \textbf{Parameter} & \textbf{Cal. M} & \textbf{Cal. $\Delta$} & \textbf{LCB} \\
\tableheadmidrule
Qwen3.5 & Tuned checkpoint & $k=3$$^{F}$ & 31.79 & \textbf{+4.33} & \textbf{+1.45} \\
Qwen3.5 & Tuned budget & $B=4.9\mathrm{k}$$^{S}$ & \textbf{78.22} & +3.00 & $-$0.90 \\
Qwen3.5 & Confidence & $\tau=+0.675$$^{F}$ & 38.97 & $-$3.67 & $-$6.24 \\
Qwen3.5 & Entropy & $\tau=+0.7$$^{F}$ & 40.23 & $-$3.67 & $-$6.24 \\
Qwen3.5 & TASR-Video & $m=0$$^{S}$ & 57.33 & \underline{+4.00} & \underline{+0.67} \\
Qwen3.5 & Forward-Q & $\tau=-0.195$$^{S}$ & 67.39 & +2.67 & $-$0.61 \\
\oursrow Qwen3.5 & CASE & $\lambda=0.4,\ b=+0.070$$^{S}$ & \underline{69.48} & +3.00 & $-$0.58 \\
\addlinespace[1pt]
GLM-4.6V & Tuned checkpoint & $k=3$$^{F}$ & 31.68 & $-$2.33 & $-$5.75 \\
GLM-4.6V & Tuned budget & $B=28.9\mathrm{k}$$^{F}$ & 39.78 & $-$2.00 & $-$5.47 \\
GLM-4.6V & Confidence & $\tau=+0.55$$^{F}$ & \underline{40.44} & \underline{+0.33} & \underline{$-$2.19} \\
GLM-4.6V & Entropy & $\tau=+0.8$$^{F}$ & \textbf{41.35} & \textbf{+1.33} & \textbf{$-$1.36} \\
GLM-4.6V & TASR-Video & $m=0.2$$^{F}$ & 39.62 & \underline{+0.33} & $-$2.42 \\
GLM-4.6V & Forward-Q & $\tau=-0.09$$^{F}$ & 40.29 & \underline{+0.33} & $-$2.31 \\
\oursrow GLM-4.6V & CASE & $\lambda=0.05,\ b=+0.070$$^{F}$ & 39.95 & +0.00 & $-$2.58 \\
\addlinespace[1pt]
InternVL3.5 & Tuned checkpoint & $k=1$$^{S}$ & 70.02 & \textbf{+4.00} & \underline{+0.24} \\
InternVL3.5 & Tuned budget & $B=3.3\mathrm{k}$$^{S}$ & \textbf{85.75} & \underline{+3.67} & $-$0.30 \\
InternVL3.5 & Confidence & $\tau=+0.55$$^{F}$ & 40.65 & +1.00 & $-$1.13 \\
InternVL3.5 & Entropy & $\tau=+0.8$$^{F}$ & 36.14 & +1.00 & $-$0.82 \\
InternVL3.5 & TASR-Video & $m=0$$^{S}$ & 59.16 & \textbf{+4.00} & \textbf{+0.49} \\
InternVL3.5 & Forward-Q & $\tau=-0.19$$^{S}$ & 76.90 & +3.33 & $-$0.44 \\
\oursrow InternVL3.5 & CASE & $\lambda=0.8,\ b=-0.190$$^{S}$ & \underline{80.94} & \underline{+3.67} & $-$0.30 \\
\bottomrule
\end{tabularx}\hspace*{-\tabcolsep}\par
\end{table}

%% file: generated/appendix_frontier_figure.tex
\begin{figure}[!t]
\centering
\includegraphics[width=0.96\textwidth]{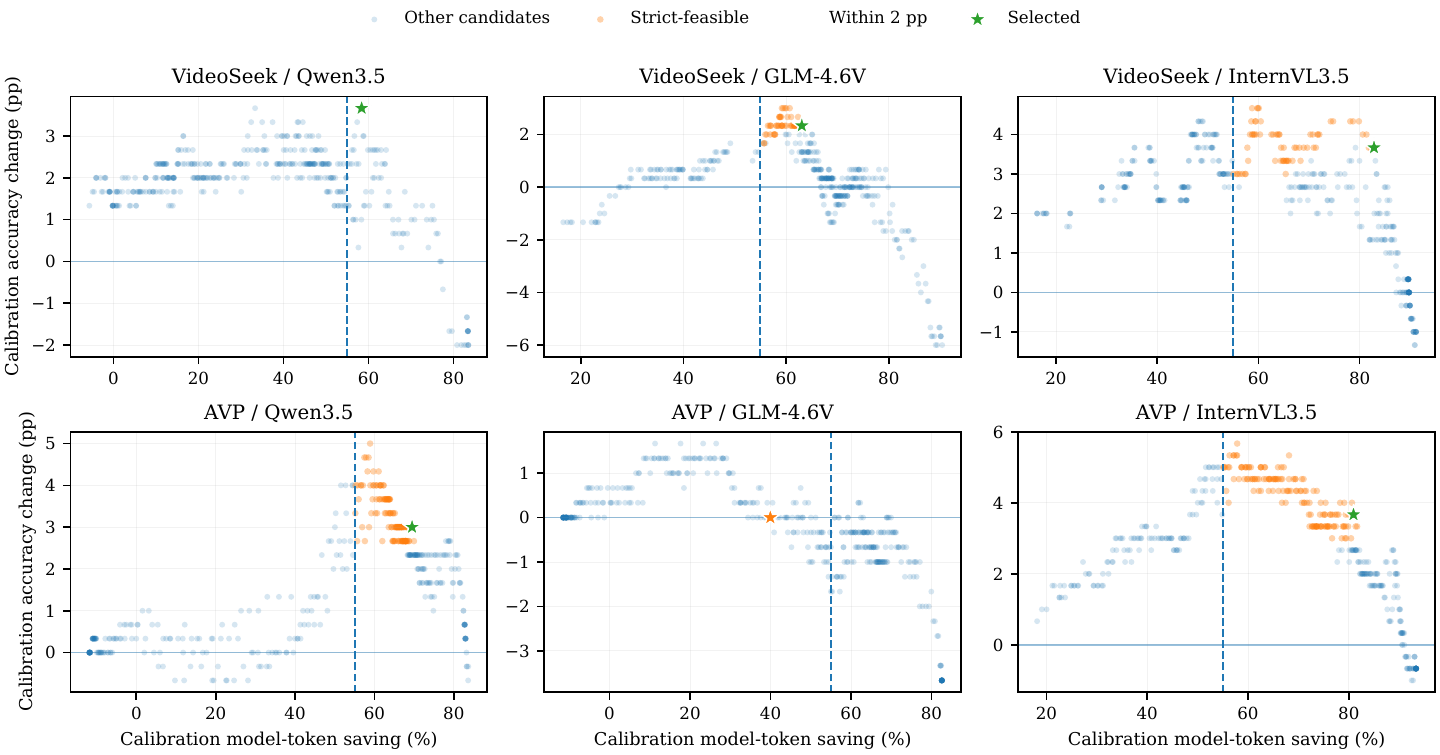}
\caption{\textbf{CASE calibration candidate clouds.} Color separates strict-feasible candidates ($\ge55\%$ saving and LCB $\ge-1$ pp) from the remaining grid; open markers show strict candidates within 2 pp of the best strict calibration accuracy, and stars mark the selected operating points. AVP--GLM uses the 40\% fallback because its strict set is empty.}
\label{fig:appendix_calibration_frontiers}
\end{figure}

%% file: generated/appendix_end_to_end_ablation_metrics_table.tex
\begin{table}[!t]
\centering
\tablestandard
\caption{\textbf{Complete end-to-end ablation metrics.} These are the same five independently selected variants as Table~\ref{tab:ablation}. $\Delta$ is accuracy change in percentage points; M and V are model- and visual-token savings; Stop is the early-stop rate. Strict counts agent--model cells selected through the primary calibration branch. Bold marks the best value and underlining the second-best.}
\label{tab:appendix_end_to_end_ablation_metrics}
\textbf{(a) Video-MME}\\[-2pt]
\par\noindent\hspace*{-\tabcolsep}\begin{tabularx}{\dimexpr\textwidth+2\tabcolsep\relax}{lYYYY}
\tableheadtoprule
\tableheadrow
\tableheadbottompad \tableheadtoppad \textbf{Variant} & \textbf{M} & \textbf{V} & \textbf{$\Delta$} & \textbf{Stop} \\
\tableheadmidrule
\oursrow Full & \underline{53.63} & \textbf{60.01} & \underline{+0.67} & \textbf{49.09} \\
Belief-only & 36.90 & 44.20 & +0.38 & 36.80 \\
Exec-state-only & \textbf{55.21} & \underline{58.05} & +0.09 & \underline{47.10} \\
Current-corr. & 35.99 & 42.21 & $-$0.19 & 35.51 \\
w/o cost correction & 46.90 & 52.69 & \textbf{+1.01} & 44.35 \\
\bottomrule
\end{tabularx}
\vspace{3pt}
\textbf{(b) LongVideoBench}\\[-2pt]
\par\noindent\hspace*{-\tabcolsep}\begin{tabularx}{\dimexpr\textwidth+2\tabcolsep\relax}{lYYYY}
\tableheadtoprule
\tableheadrow
\tableheadbottompad \tableheadtoppad \textbf{Variant} & \textbf{M} & \textbf{V} & \textbf{$\Delta$} & \textbf{Stop} \\
\tableheadmidrule
\oursrow Full & \underline{58.78} & \underline{66.66} & \underline{+3.38} & \textbf{56.53} \\
Belief-only & 33.46 & 41.55 & +2.49 & 36.37 \\
Exec-state-only & \textbf{64.48} & \textbf{68.15} & +2.26 & \underline{56.20} \\
Current-corr. & 31.55 & 39.19 & +2.02 & 35.13 \\
w/o cost correction & 50.07 & 57.32 & \textbf{+3.66} & 49.98 \\
\bottomrule
\end{tabularx}
\vspace{3pt}
\textbf{(c) MLVU}\\[-2pt]
\par\noindent\hspace*{-\tabcolsep}\begin{tabularx}{\dimexpr\textwidth+2\tabcolsep\relax}{lYYYY}
\tableheadtoprule
\tableheadrow
\tableheadbottompad \tableheadtoppad \textbf{Variant} & \textbf{M} & \textbf{V} & \textbf{$\Delta$} & \textbf{Stop} \\
\tableheadmidrule
\oursrow Full & \textbf{51.28} & \textbf{59.48} & \textbf{+4.58} & \textbf{60.40} \\
Belief-only & \underline{49.28} & \underline{57.19} & +2.70 & \underline{54.88} \\
Exec-state-only & 46.20 & 51.36 & +4.27 & 53.83 \\
Current-corr. & 40.57 & 47.96 & +1.34 & 46.93 \\
w/o cost correction & 41.02 & 49.17 & \underline{+4.33} & 53.79 \\
\bottomrule
\end{tabularx}
\vspace{4pt}
\textbf{(d) Overall summary}\\[-2pt]
\par\noindent\hspace*{-\tabcolsep}\begin{tabularx}{\dimexpr\textwidth+2\tabcolsep\relax}{lYYYYY}
\tableheadtoprule
\tableheadrow
\tableheadbottompad \tableheadtoppad \textbf{Variant} & \textbf{M} & \textbf{V} & \textbf{$\Delta$} & \textbf{Stop} & \textbf{Strict /6} \\
\tableheadmidrule
\oursrow Full & \underline{54.56} & \textbf{62.05} & \underline{+2.88} & \textbf{55.34} & \textbf{5/6} \\
Belief-only & 39.88 & 47.65 & +1.86 & 42.68 & 2/6 \\
Exec-state-only & \textbf{55.30} & \underline{59.19} & +2.21 & \underline{52.37} & \underline{4/6} \\
Current-corr. & 36.04 & 43.12 & +1.06 & 39.19 & 1/6 \\
w/o cost correction & 46.00 & 53.06 & \textbf{+3.00} & 49.37 & \underline{4/6} \\
\bottomrule
\end{tabularx}
\end{table}

%% file: generated/appendix_cost_components_table.tex
\begin{table}[H]
\centering
\tablestandard
\caption{\textbf{Quality and continuation-cost components under independent selection.} Each variant selects its own Video-MME operating point with the shared rule and transfers the frozen policy to LongVideoBench and MLVU. Entries are Model Token saving (\%) / end-to-end accuracy change (pp). The factorized Full predictor agrees with the original direct Gap-Ridge to a maximum absolute calibration-score difference of $5.6\times10^{-15}$.}
\label{tab:appendix_cost_components}
\par\noindent\hspace*{-\tabcolsep}\begin{tabularx}{\dimexpr\textwidth+2\tabcolsep\relax}{lYYYY}
\tableheadtoprule
\tableheadrow
\tableheadbottompad \tableheadtoppad \textbf{Variant} & \textbf{Video-MME} & \textbf{LongVideoBench} & \textbf{MLVU} & \textbf{Overall} \\
\tableheadmidrule
\oursrow Quality + heterogeneous-cost correction & \textbf{53.63} / +0.67 & \textbf{58.78} / +3.38 & \textbf{51.28} / \textbf{+4.58} & \textbf{54.56} / \underline{+2.88} \\
Quality only (w/o cost correction) & 46.90 / \textbf{+1.01} & 50.07 / \textbf{+3.66} & 41.02 / \underline{+4.33} & 46.00 / \textbf{+3.00} \\
Quality + constant-step cost & \underline{49.17} / \underline{+0.84} & \underline{52.35} / \underline{+3.40} & \underline{45.57} / +4.20 & \underline{49.03} / +2.82 \\
\bottomrule
\end{tabularx}\hspace*{-\tabcolsep}\par
\end{table}

%% file: generated/appendix_core_ablations_table.tex
\begin{table}[!t]
\centering
\tablestandard
\fontsize{7.0pt}{8.1pt}\selectfont
\caption{\textbf{Core saving-matched diagnostics.} Alternatives are selected at the calibration Model Token saving closest to the corresponding Full policy, with no additional saving floor. This controls stopping aggressiveness and isolates accuracy differences at similar saving. Entries report accuracy change (pp), Model Token saving (M), and stop rate (S).}
\label{tab:appendix_core_ablations}
\textbf{(a) VideoSeek}\\[-2pt]
\par\noindent\hspace*{-\tabcolsep}\begin{tabularx}{\dimexpr\textwidth+2\tabcolsep\relax}{llYYYYYYYYY}
\tableheadtoprule
\tableheadrowspan
\tableheadtoppad \textbf{Model} & \textbf{Variant} & \multicolumn{3}{c}{\textbf{Video-MME}} & \multicolumn{3}{c}{\textbf{LongVideoBench}} & \multicolumn{3}{c}{\textbf{MLVU}} \\
\tableheadrow
\tableheadbottompad  &  & \textbf{M} & \textbf{$\Delta$} & \textbf{S} & \textbf{M} & \textbf{$\Delta$} & \textbf{S} & \textbf{M} & \textbf{$\Delta$} & \textbf{S} \\
\tableheadmidrule
\oursrow Qwen3.5 & Full & \textbf{52.43} & \underline{+1.91} & \underline{42.18} & 64.24 & \textbf{+3.22} & 48.02 & 59.46 & \underline{+2.30} & 47.84 \\
Qwen3.5 & Belief-only & 50.16 & \textbf{+2.00} & 42.09 & 64.06 & +1.87 & \textbf{49.96} & \underline{68.95} & +2.12 & \textbf{55.29} \\
Qwen3.5 & Execution-state-only & \underline{51.83} & +0.98 & 37.82 & \underline{64.72} & \underline{+2.62} & 46.00 & 68.23 & +1.10 & 52.44 \\
Qwen3.5 & Current-correctness & 50.81 & +1.64 & \textbf{42.84} & 60.74 & +2.09 & 46.45 & 51.02 & \textbf{+2.94} & 45.26 \\
Qwen3.5 & One-step target & 51.31 & +0.93 & 40.49 & \textbf{65.43} & \textbf{+3.22} & \underline{49.29} & \textbf{68.98} & +1.70 & \underline{54.83} \\
\addlinespace[0.5pt]
\oursrow GLM-4.6V & Full & 49.77 & $-$0.71 & 23.42 & 54.87 & \underline{$-$0.15} & 29.54 & 45.52 & \underline{$-$0.28} & 26.03 \\
GLM-4.6V & Belief-only & 51.94 & $-$2.09 & \underline{31.47} & 52.23 & $-$1.72 & \textbf{33.21} & \underline{71.01} & $-$5.84 & \underline{58.92} \\
GLM-4.6V & Execution-state-only & 48.79 & \underline{$-$0.67} & 17.60 & \textbf{58.98} & \textbf{+0.37} & 26.03 & 27.01 & \textbf{+0.28} & 9.29 \\
GLM-4.6V & Current-correctness & \textbf{52.38} & $-$2.89 & \textbf{34.67} & 41.89 & $-$1.42 & 29.17 & \textbf{76.67} & $-$9.66 & \textbf{69.00} \\
GLM-4.6V & One-step target & \underline{52.20} & \textbf{$-$0.53} & 24.76 & \underline{55.71} & \underline{$-$0.15} & \underline{29.77} & 60.57 & $-$1.61 & 39.10 \\
\addlinespace[0.5pt]
\oursrow InternVL3.5 & Full & 69.85 & \underline{$-$0.18} & 57.47 & 74.08 & +3.89 & \underline{66.57} & 54.20 & \underline{+5.89} & 63.57 \\
InternVL3.5 & Belief-only & 70.08 & $-$0.89 & \underline{58.13} & 64.45 & \textbf{+4.26} & 58.86 & \underline{86.01} & +3.86 & \underline{90.89} \\
InternVL3.5 & Execution-state-only & \underline{70.97} & \textbf{+0.31} & 57.11 & \textbf{76.79} & \underline{+4.11} & \textbf{66.94} & 65.23 & +5.43 & 75.34 \\
InternVL3.5 & Current-correctness & \textbf{74.98} & $-$1.73 & \textbf{62.62} & \underline{74.41} & +4.04 & \textbf{66.94} & \textbf{87.00} & +3.27 & \textbf{91.49} \\
InternVL3.5 & One-step target & 69.56 & $-$0.36 & 57.38 & 73.51 & \underline{+4.11} & 66.27 & 58.86 & \textbf{+6.16} & 68.49 \\
\bottomrule
\end{tabularx}
\vspace{4pt}
\textbf{(b) AVP}\\[-2pt]
\par\noindent\hspace*{-\tabcolsep}\begin{tabularx}{\dimexpr\textwidth+2\tabcolsep\relax}{llYYYYYYYYY}
\tableheadtoprule
\tableheadrowspan
\tableheadtoppad \textbf{Model} & \textbf{Variant} & \multicolumn{3}{c}{\textbf{Video-MME}} & \multicolumn{3}{c}{\textbf{LongVideoBench}} & \multicolumn{3}{c}{\textbf{MLVU}} \\
\tableheadrow
\tableheadbottompad  &  & \textbf{M} & \textbf{$\Delta$} & \textbf{S} & \textbf{M} & \textbf{$\Delta$} & \textbf{S} & \textbf{M} & \textbf{$\Delta$} & \textbf{S} \\
\tableheadmidrule
\oursrow Qwen3.5 & Full & \textbf{56.67} & +0.27 & \underline{56.80} & \underline{55.83} & +2.69 & \textbf{61.11} & \underline{65.70} & +4.46 & \underline{75.76} \\
Qwen3.5 & Belief-only & 54.74 & $-$2.40 & 55.02 & 50.90 & +1.65 & 58.56 & 58.79 & +1.29 & 68.40 \\
Qwen3.5 & Execution-state-only & 56.22 & \textbf{+1.02} & 55.20 & \textbf{57.04} & \textbf{+3.29} & 57.52 & 52.71 & \textbf{+5.01} & 71.07 \\
Qwen3.5 & Current-correctness & 55.97 & $-$3.47 & \textbf{57.11} & 50.21 & $-$0.30 & 57.37 & \textbf{71.67} & +0.23 & \textbf{83.07} \\
Qwen3.5 & One-step target & \underline{56.49} & \underline{+0.36} & 56.62 & 55.39 & \underline{+2.77} & \underline{60.28} & 64.20 & \underline{+4.74} & 75.57 \\
\addlinespace[0.5pt]
\oursrow GLM-4.6V & Full & \underline{30.72} & $-$0.40 & \textbf{55.69} & \underline{35.43} & \underline{+3.07} & \underline{65.52} & 31.19 & +0.41 & \textbf{66.24} \\
GLM-4.6V & Belief-only & 27.00 & $-$0.62 & 42.62 & 17.92 & +0.97 & 36.13 & \textbf{45.53} & $-$0.46 & \underline{63.57} \\
GLM-4.6V & Execution-state-only & \textbf{31.97} & \textbf{+0.13} & 53.02 & \textbf{52.16} & \underline{+3.07} & \textbf{71.95} & 7.42 & \textbf{+0.78} & 29.48 \\
GLM-4.6V & Current-correctness & 26.77 & $-$0.53 & 40.76 & 21.38 & +0.97 & 39.64 & \underline{38.16} & $-$0.05 & 56.58 \\
GLM-4.6V & One-step target & 30.65 & \underline{$-$0.22} & \underline{54.76} & 33.86 & \textbf{+3.14} & 61.78 & 13.63 & \underline{+0.60} & 49.68 \\
\addlinespace[0.5pt]
\oursrow InternVL3.5 & Full & 62.35 & \textbf{+3.16} & 58.98 & 68.23 & \textbf{+7.55} & 68.44 & 51.58 & \underline{+14.67} & \underline{82.98} \\
InternVL3.5 & Belief-only & 63.32 & +0.04 & \textbf{62.09} & 67.58 & \underline{+7.18} & 68.51 & \textbf{78.29} & +10.49 & \textbf{92.50} \\
InternVL3.5 & Execution-state-only & \underline{63.57} & \underline{+2.84} & 57.91 & \underline{72.42} & \textbf{+7.55} & 68.14 & 52.11 & \textbf{+14.95} & 82.70 \\
InternVL3.5 & Current-correctness & \textbf{65.39} & +0.00 & \underline{61.16} & \textbf{76.21} & +6.43 & \textbf{72.85} & \underline{70.85} & +10.44 & 81.42 \\
InternVL3.5 & One-step target & 63.03 & \textbf{+3.16} & 59.16 & 70.98 & \underline{+7.18} & \underline{69.93} & 52.64 & +14.44 & 82.61 \\
\bottomrule
\end{tabularx}
\end{table}

%% file: generated/appendix_target_family_table.tex
\begin{table}[H]
\centering
\tablestandard
\caption{\textbf{Continuation-horizon diagnostics.} Alternatives are saving-matched to Full when feasible so that the comparison isolates the supervision horizon at similar stopping aggressiveness. Means are over the feasible agent--model cells; evaluation always uses the real complete cost ledger.}
\label{tab:appendix_target_family}
\par\noindent\hspace*{-\tabcolsep}\begin{tabularx}{\dimexpr\textwidth+2\tabcolsep\relax}{lYYYYY}
\tableheadtoprule
\tableheadrow
\tableheadbottompad \tableheadtoppad \textbf{Target} & \textbf{Feasible} & \textbf{VMME M} & \textbf{VMME $\Delta$} & \textbf{LVB M} & \textbf{LVB $\Delta$} \\
\tableheadmidrule
\oursrow Full suffix & 6/6 & 53.63 & +0.67 & 58.78 & \underline{+3.38} \\
Native-terminal only & 6/6 & \textbf{54.33} & +0.12 & \textbf{62.90} & +2.61 \\
Suffix horizon 1 & 6/6 & \underline{53.87} & +0.56 & \underline{59.15} & \underline{+3.38} \\
Suffix horizon 2 & 6/6 & 53.60 & \textbf{+0.71} & 58.44 & \textbf{+3.43} \\
Suffix horizon 4 & 6/6 & 53.49 & \underline{+0.70} & 58.85 & +3.34 \\
\bottomrule
\end{tabularx}
\end{table}

%% file: generated/appendix_fine_controls_table.tex
\begin{table}[!t]
\centering
\tablestandard
\caption{\textbf{Fine-grained state and fitting controls.} Panel (a) removes individual state channels or changes checkpoint weighting at saving-matched calibration points. Panel (b) changes only the Ridge regularization strength and uses the same calibration rule as Full.}
\label{tab:appendix_fine_controls}
\textbf{(a) State channels and weighting}\\[-2pt]
\par\noindent\hspace*{-\tabcolsep}\begin{tabularx}{\dimexpr\textwidth+2\tabcolsep\relax}{lYYYY}
\tableheadtoprule
\tableheadrow
\tableheadbottompad \tableheadtoppad \textbf{Variant} & \textbf{VMME M} & \textbf{VMME $\Delta$} & \textbf{LVB M} & \textbf{LVB $\Delta$} \\
\tableheadmidrule
\oursrow Full & \underline{53.63} & \textbf{+0.67} & 58.78 & \underline{+3.38} \\
w/o belief change & 52.88 & \underline{+0.67} & 57.73 & +3.32 \\
w/o consumed cost & 53.00 & +0.62 & 59.41 & +3.10 \\
w/o tool type & \textbf{53.68} & +0.60 & \underline{59.70} & \textbf{+3.52} \\
w/o duration + subtitle & 53.59 & +0.66 & 58.37 & +3.24 \\
Checkpoint-uniform weight & 53.53 & +0.59 & \textbf{61.55} & +3.20 \\
\bottomrule
\end{tabularx}
\vspace{4pt}
\textbf{(b) Ridge regularization}\\[-2pt]
\par\noindent\hspace*{-\tabcolsep}\begin{tabularx}{\dimexpr\textwidth+2\tabcolsep\relax}{YYYYY}
\tableheadtoprule
\tableheadrow
\tableheadbottompad \tableheadtoppad \textbf{$\alpha$} & \textbf{VMME M} & \textbf{VMME $\Delta$} & \textbf{LVB M} & \textbf{LVB $\Delta$} \\
\tableheadmidrule
0.1 & 51.05 & +0.43 & 53.81 & +2.87 \\
1.0 & 51.29 & +0.48 & 55.58 & +3.12 \\
\oursrow 10.0 & \textbf{53.63} & \underline{+0.67} & \textbf{58.78} & +3.38 \\
100.0 & \underline{51.52} & +0.57 & \underline{55.94} & \textbf{+3.58} \\
1000.0 & 48.61 & \textbf{+0.89} & 52.05 & \underline{+3.55} \\
\bottomrule
\end{tabularx}
\end{table}

%% file: generated/appendix_training_scaling_figure.tex
\begin{figure}[H]
\centering
\includegraphics[width=0.82\textwidth]{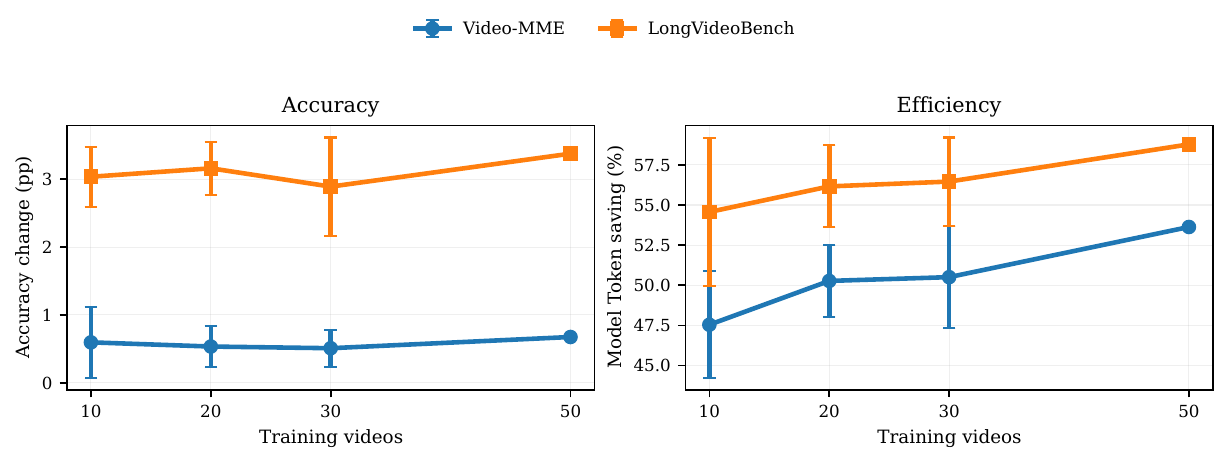}
\caption{\textbf{Training-video scaling.} For 10, 20, and 30 videos, points show the mean over three fixed nested training orders and error bars show one standard deviation; the 50-video point is the canonical training role. Probe temperatures and Gap-Ridge are refit for every subset, while the 100-video calibration role remains fixed.}
\label{fig:appendix_training_scaling}
\end{figure}

%% file: generated/appendix_probe_controls_table.tex
\begin{table}[!t]
\centering
\tablestandard
\caption{\textbf{Probe cadence and probability calibration.} Values are means over all six agent--model cells under the shared calibration rule.}
\label{tab:appendix_probe_controls}
\textbf{(a) Decision-check cadence}\\[-2pt]
\par\noindent\hspace*{-\tabcolsep}\begin{tabularx}{\dimexpr\textwidth+2\tabcolsep\relax}{lYYYY}
\tableheadtoprule
\tableheadrow
\tableheadbottompad \tableheadtoppad \textbf{Cadence} & \textbf{VMME M} & \textbf{VMME $\Delta$} & \textbf{LVB M} & \textbf{LVB $\Delta$} \\
\tableheadmidrule
\oursrow Every checkpoint & \textbf{53.63} & \underline{+0.67} & \textbf{58.78} & \textbf{+3.38} \\
Every 2 checkpoints & 44.64 & \textbf{+0.90} & 51.02 & \underline{+2.98} \\
Every 3 checkpoints & \underline{46.10} & \underline{+0.67} & \underline{51.73} & +2.31 \\
\bottomrule
\end{tabularx}
\vspace{4pt}
\textbf{(b) Option-probability calibration}\\[-2pt]
\par\noindent\hspace*{-\tabcolsep}\begin{tabularx}{\dimexpr\textwidth+2\tabcolsep\relax}{lYYYY}
\tableheadtoprule
\tableheadrow
\tableheadbottompad \tableheadtoppad \textbf{Calibration} & \textbf{VMME M} & \textbf{VMME $\Delta$} & \textbf{LVB M} & \textbf{LVB $\Delta$} \\
\tableheadmidrule
Raw ($T=1$) & \underline{52.44} & \underline{+0.51} & \textbf{59.02} & \underline{+2.94} \\
Single global $T$ & 51.45 & +0.34 & 57.42 & +2.84 \\
\oursrow Depth-group $T$ & \textbf{53.63} & \textbf{+0.67} & \underline{58.78} & \textbf{+3.38} \\
\bottomrule
\end{tabularx}
\end{table}

%% file: generated/appendix_probe_calibration_quality_table.tex
\begin{table}[!t]
\centering
\tablestandard
\caption{\textbf{Held-out option-probability calibration.} Temperatures are fit only on the Video-MME training role. NLL and multiclass Brier score are pooled over eligible checkpoints of the six agent--model cells; lower is better.}
\label{tab:appendix_probe_calibration_quality}
\begin{tabularx}{\textwidth}{lYYYYY}
\tableheadtoprule
\tableheadrow
\tableheadbottompad \tableheadtoppad \textbf{Evaluation set} & \textbf{Checkpoints} & \textbf{Raw NLL} & \textbf{Depth-$T$ NLL} & \textbf{Raw Brier} & \textbf{Depth-$T$ Brier} \\
\tableheadmidrule
Video-MME & 45,620 & 1.945 & \textbf{1.216} & 0.793 & \textbf{0.653} \\
LongVideoBench & 30,460 & 1.879 & \textbf{1.313} & 0.767 & \textbf{0.661} \\
MLVU & 48,013 & 1.799 & \textbf{1.197} & 0.759 & \textbf{0.640} \\
\bottomrule
\end{tabularx}
\end{table}

%% file: generated/appendix_answer_transitions_table.tex
\begin{table}[!t]
\centering
\tablestandard
\caption{\textbf{Per-question answer transitions.} $W\!\rightarrow\!C$ and $C\!\rightarrow\!W$ count correctness changes from Native to CASE; their difference divided by the full benchmark size gives the net accuracy change.}
\label{tab:appendix_answer_transitions}
\textbf{(a) Video-MME}\\[-2pt]
\par\noindent\hspace*{-\tabcolsep}\begin{tabularx}{\dimexpr\textwidth+2\tabcolsep\relax}{llYYYY}
\tableheadtoprule
\tableheadrow
\tableheadbottompad \tableheadtoppad \textbf{Agent} & \textbf{Model} & \textbf{$W\!\rightarrow\!C$} & \textbf{$C\!\rightarrow\!W$} & \textbf{Both correct} & \textbf{Net $\Delta$ (pp)} \\
\tableheadmidrule
VideoSeek & Qwen3.5 & 197 & 154 & 1012 & +1.91 \\
VideoSeek & GLM-4.6V & 62 & 78 & 1191 & -0.71 \\
VideoSeek & InternVL3.5 & 167 & 171 & 1018 & -0.18 \\
AVP & Qwen3.5 & 105 & 99 & 1364 & +0.27 \\
AVP & GLM-4.6V & 72 & 81 & 1222 & -0.40 \\
AVP & InternVL3.5 & 189 & 118 & 1032 & +3.16 \\
\bottomrule
\end{tabularx}\hspace*{-\tabcolsep}\par
\vspace{3pt}
\textbf{(b) LongVideoBench}\\[-2pt]
\par\noindent\hspace*{-\tabcolsep}\begin{tabularx}{\dimexpr\textwidth+2\tabcolsep\relax}{llYYYY}
\tableheadtoprule
\tableheadrow
\tableheadbottompad \tableheadtoppad \textbf{Agent} & \textbf{Model} & \textbf{$W\!\rightarrow\!C$} & \textbf{$C\!\rightarrow\!W$} & \textbf{Both correct} & \textbf{Net $\Delta$ (pp)} \\
\tableheadmidrule
VideoSeek & Qwen3.5 & 154 & 111 & 436 & +3.22 \\
VideoSeek & GLM-4.6V & 51 & 53 & 582 & -0.15 \\
VideoSeek & InternVL3.5 & 162 & 110 & 453 & +3.89 \\
AVP & Qwen3.5 & 110 & 74 & 715 & +2.69 \\
AVP & GLM-4.6V & 86 & 45 & 641 & +3.07 \\
AVP & InternVL3.5 & 217 & 116 & 395 & +7.55 \\
\bottomrule
\end{tabularx}\hspace*{-\tabcolsep}\par
\vspace{3pt}
\textbf{(c) MLVU}\\[-2pt]
\par\noindent\hspace*{-\tabcolsep}\begin{tabularx}{\dimexpr\textwidth+2\tabcolsep\relax}{llYYYY}
\tableheadtoprule
\tableheadrow
\tableheadbottompad \tableheadtoppad \textbf{Agent} & \textbf{Model} & \textbf{$W\!\rightarrow\!C$} & \textbf{$C\!\rightarrow\!W$} & \textbf{Both correct} & \textbf{Net $\Delta$ (pp)} \\
\tableheadmidrule
VideoSeek & Qwen3.5 & 229 & 179 & 812 & +2.30 \\
VideoSeek & GLM-4.6V & 53 & 59 & 1033 & -0.28 \\
VideoSeek & InternVL3.5 & 233 & 105 & 873 & +5.89 \\
AVP & Qwen3.5 & 199 & 102 & 1103 & +4.46 \\
AVP & GLM-4.6V & 55 & 46 & 1109 & +0.41 \\
AVP & InternVL3.5 & 399 & 80 & 661 & +14.67 \\
\bottomrule
\end{tabularx}\hspace*{-\tabcolsep}\par
\end{table}

%% file: generated/appendix_cost_decile_figure.tex
\begin{figure}[!t]
\centering
\includegraphics[width=0.94\textwidth]{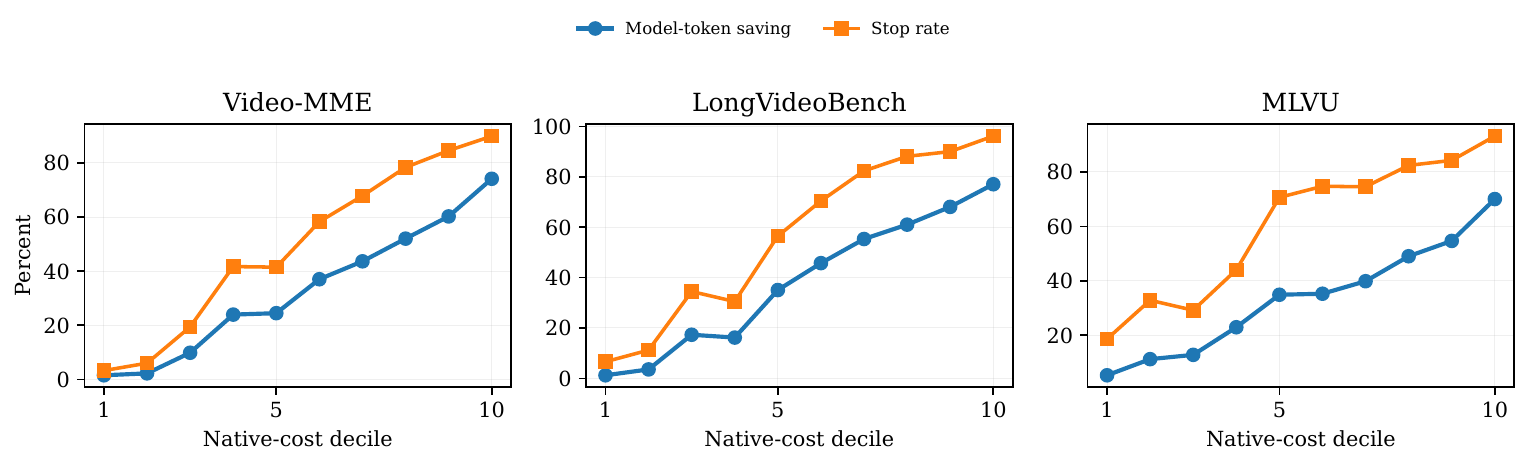}
\caption{\textbf{Cost concentration of CASE stopping.} Questions are partitioned by Native Model Token cost within each agent--model--benchmark cell and then averaged over the six cells.}
\label{fig:appendix_cost_deciles}
\end{figure}

%% file: generated/appendix_ridge_coefficients_figure.tex
\begin{figure}[!t]
\centering
\includegraphics[width=0.96\textwidth]{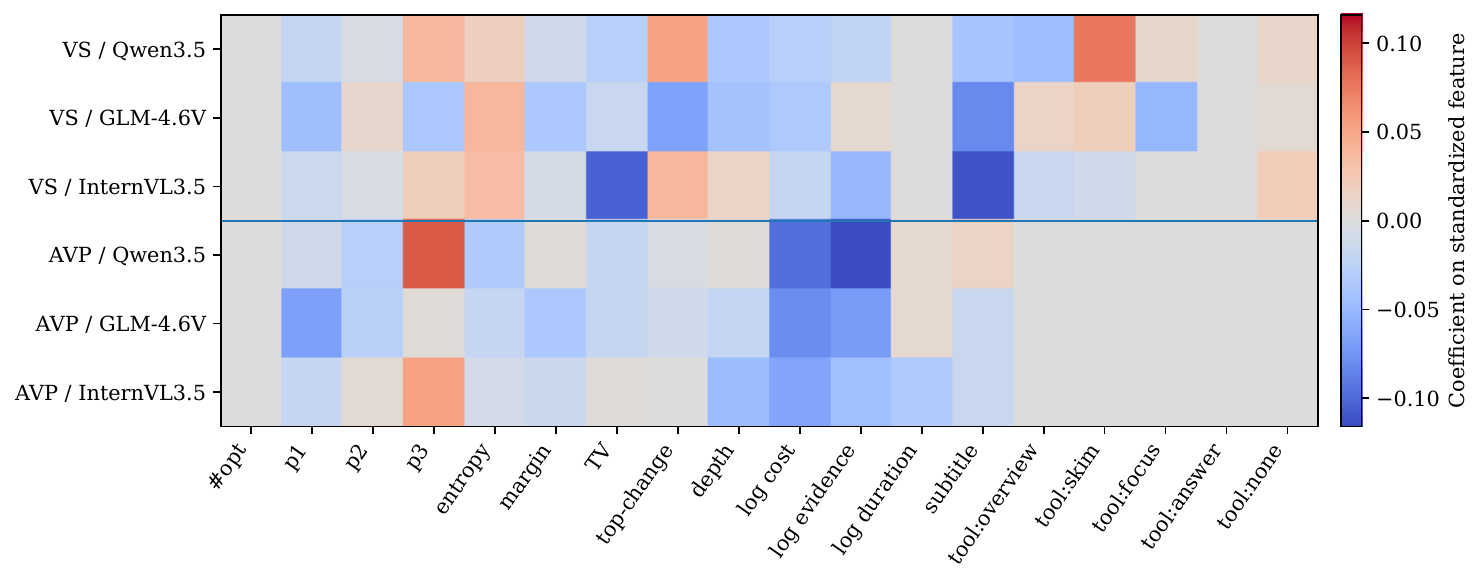}
\caption{\textbf{Standardized Ridge coefficients of the six selected CASE policies.} Features are standardized within each fitted policy; the plot summarizes the learned linear stopping signals across hosts and models.}
\label{fig:appendix_ridge_coefficients}
\end{figure}

%% file: generated/appendix_dev_resampling_table.tex
\begin{table}[!t]
\centering
\tablestandard
\caption{\textbf{Balanced development-role resampling.} Mean $\pm$ sample standard deviation over nine alternative 50-train/100-calibration assignments of the same 150 Video-MME development videos. Every split is refitted and selected with the same calibration rule; LongVideoBench uses the frozen Video-MME policy.}
\label{tab:appendix_dev_resampling}
\par\noindent\hspace*{-\tabcolsep}\begin{tabularx}{\dimexpr\textwidth+2\tabcolsep\relax}{llYYYYY}
\tableheadtoprule
\tableheadrow
\tableheadbottompad \tableheadtoppad \textbf{Agent} & \textbf{Model} & \textbf{VMM M} & \textbf{VMM $\Delta$} & \textbf{LVB M} & \textbf{LVB $\Delta$} & \textbf{Strict splits} \\
\tableheadmidrule
VideoSeek & Qwen3.5 & 51.82 $\pm$ 5.00 & +1.39 $\pm$ 0.64 & 68.82 $\pm$ 7.03 & +1.60 $\pm$ 1.88 & 2/9 \\
VideoSeek & GLM-4.6V & 45.22 $\pm$ 8.13 & -1.30 $\pm$ 0.90 & 46.62 $\pm$ 23.52 & -1.60 $\pm$ 1.69 & 2/9 \\
VideoSeek & InternVL3.5 & 61.03 $\pm$ 16.43 & +0.07 $\pm$ 0.91 & 59.28 $\pm$ 24.51 & +3.56 $\pm$ 1.00 & 7/9 \\
AVP & Qwen3.5 & 61.75 $\pm$ 8.31 & -1.06 $\pm$ 1.22 & 61.07 $\pm$ 14.60 & -0.79 $\pm$ 2.92 & 8/9 \\
AVP & GLM-4.6V & 40.75 $\pm$ 1.83 & -0.49 $\pm$ 0.18 & 41.05 $\pm$ 7.78 & +2.88 $\pm$ 0.59 & 0/9 \\
AVP & InternVL3.5 & 65.07 $\pm$ 3.03 & +1.93 $\pm$ 1.25 & 74.12 $\pm$ 2.81 & +7.01 $\pm$ 0.48 & 9/9 \\
\bottomrule
\end{tabularx}\hspace*{-\tabcolsep}\par
\end{table}

%% file: generated/appendix_multiseed_table.tex
\begin{table}[H]
\centering
\tablestandard
\caption{\textbf{Trajectory-seed robustness.} Mean $\pm$ sample standard deviation over native-trajectory seeds 42, 43, and 3407. The complete CASE controller learned and selected with seed 42 is held fixed for all three seeds; only the native host trajectories change. $\Delta$ is accuracy change in percentage points and M is Model Token saving (\%).}
\label{tab:appendix_multiseed}
\par\noindent\hspace*{-\tabcolsep}\begin{tabularx}{\dimexpr\textwidth+2\tabcolsep\relax}{llYYYYYY}
\tableheadtoprule
\tableheadrowspan
\tableheadtoppad \textbf{Agent} & \textbf{Model} & \multicolumn{2}{c}{\textbf{Video-MME}} & \multicolumn{2}{c}{\textbf{LongVideoBench}} & \multicolumn{2}{c}{\textbf{MLVU}} \\
\tableheadrow
\tableheadbottompad  &  & \textbf{M} & \textbf{$\Delta$} & \textbf{M} & \textbf{$\Delta$} & \textbf{M} & \textbf{$\Delta$} \\
\tableheadmidrule
VideoSeek & Qwen3.5 & 51.89 $\pm$ 3.25 & +1.59 $\pm$ 2.33 & 61.42 $\pm$ 4.13 & +2.37 $\pm$ 2.01 & 56.92 $\pm$ 2.30 & +2.30 $\pm$ 0.83 \\
VideoSeek & GLM-4.6V & 42.17 $\pm$ 8.59 & -0.56 $\pm$ 1.34 & 49.35 $\pm$ 6.82 & -0.20 $\pm$ 1.12 & 41.97 $\pm$ 5.94 & -0.78 $\pm$ 0.51 \\
VideoSeek & InternVL3.5 & 59.50 $\pm$ 10.00 & -0.21 $\pm$ 0.18 & 71.01 $\pm$ 4.10 & +5.68 $\pm$ 1.80 & 56.47 $\pm$ 6.52 & +7.41 $\pm$ 1.43 \\
\midrule
AVP & Qwen3.5 & 56.36 $\pm$ 0.34 & +0.06 $\pm$ 0.22 & 55.94 $\pm$ 1.01 & +1.84 $\pm$ 1.22 & 67.14 $\pm$ 1.56 & +3.99 $\pm$ 0.48 \\
AVP & GLM-4.6V & 30.13 $\pm$ 0.70 & +0.10 $\pm$ 0.51 & 34.76 $\pm$ 0.90 & +3.86 $\pm$ 0.87 & 30.50 $\pm$ 0.62 & +0.44 $\pm$ 0.46 \\
AVP & InternVL3.5 & 61.55 $\pm$ 0.87 & +1.69 $\pm$ 1.28 & 66.41 $\pm$ 2.31 & +4.06 $\pm$ 4.05 & 48.46 $\pm$ 4.58 & +8.95 $\pm$ 5.00 \\
\bottomrule
\end{tabularx}\hspace*{-\tabcolsep}\par
\end{table}